\documentclass{article}

\usepackage{microtype}
\usepackage{graphicx}
\usepackage{subfigure}
\usepackage{booktabs} 

\usepackage{hyperref}

\usepackage[accepted]{icml2024}

\renewcommand{\algorithmiccomment}[1]{// #1}
\renewcommand\algorithmicthen{}
\renewcommand\algorithmicdo{}

\usepackage{amsmath}
\usepackage{amssymb}
\usepackage{mathtools}
\usepackage{amsthm}

\usepackage{microtype}
\usepackage{graphicx}
\usepackage{booktabs} 
\usepackage{multirow}
\usepackage{array}
\usepackage{bbm}
\usepackage{caption}
\usepackage{setspace}
\usepackage{dblfloatfix}
\newcolumntype{P}[1]{>{\centering\arraybackslash}p{#1}}

\usepackage{bm}
\usepackage[capitalize,noabbrev]{cleveref}

\newcommand{\EXD}[2]{\mathbb{E}_{#2}\Big[ #1 \Big]}
\newcommand{\weight}{\bm{\pi}}
\newcommand{\state}{\bm{\theta}}
\newcommand{\observation}{\bm{z}}
\newcommand{\Observation}{Z}
\newcommand{\loss}[1]{\ell_{\state}(#1)}
\newcommand{\Loss}[2]{\ell_{#1}(#2)}
\newcommand{\elbo}[1]{\mathcal{L}(#1)}
\newcommand{\beststate}{\state^{\star}}
\newcommand{\truedistr}{p}
\newcommand{\argmin}[2]{\underset{#1}{\arg\min}\, #2}
\newcommand{\argmax}[2]{\underset{#1}{\arg\max}\, #2}

\newcommand{\mysum}[2]{\sum\limits_{#1}^{#2}}
\newcommand{\one}{\bm{1}}

\newcommand{\aw}{\langle \weight \rangle}
\newcommand{\naw}{1 - \langle \weight \rangle}
\newcommand{\lossi}{\ell_i}

\theoremstyle{plain}

\theoremstyle{definition}

\theoremstyle{remark}

\newcommand{\coseniorauthors}{\textsuperscript{*}Co-senior authors}

\usepackage[textsize=tiny]{todonotes}

\icmltitlerunning{Adaptive Robust Learning using Latent Bernoulli Variables}

\begin{document}

\twocolumn[
\icmltitle{Adaptive Robust Learning using Latent Bernoulli Variables}




\begin{icmlauthorlist}
\icmlauthor{Aleksandr Karakulev}{it}
\icmlauthor{Dave Zachariah$^*$}{it}
\icmlauthor{Prashant Singh$^*$}{it,life}
\end{icmlauthorlist}

\icmlaffiliation{it}{Uppsala University, Sweden}
\icmlaffiliation{life}{Science for Life Laboratory, Sweden}

\icmlcorrespondingauthor{Aleksandr Karakulev}{aleksandr.karakulev@it.uu.se}

\icmlkeywords{Machine Learning, ICML}

\vskip 0.3in
]



\printAffiliationsAndNotice{\coseniorauthors}

\begin{abstract}
We present an adaptive approach for robust learning from corrupted training sets. 
We identify corrupted and non-corrupted samples with latent Bernoulli variables and thus formulate the learning problem as maximization of the likelihood where latent variables are marginalized. The resulting problem is solved via variational inference, using an efficient Expectation-Maximization based method. The proposed approach improves over the state-of-the-art by automatically inferring the corruption level, while adding minimal computational overhead. We demonstrate our robust learning method and its parameter-free nature on a wide variety of machine learning tasks including online learning and deep learning where it adapts to different levels of noise and maintains high prediction accuracy.
\end{abstract}    
\section{Introduction}
\label{introduction}
Several statistical learning problems are formulated as estimation of parameters $\state \in \Theta$ of a probabilistic model by maximizing its likelihood function $\prod_{i=1}^n p(\observation_i | \state)$ given $n$ independent observations $\Observation = \{\observation_i\}_{i=1}^n$, $\observation_i \sim \truedistr(\observation)$.
By defining the loss function as the negative log-likelihood $\loss{\observation} = -\ln p(\observation | \state)$, one can equivalently solve
\begin{align}
\state_\text{ML} = \argmax{\state \in \Theta}{\prod_{i=1}^n p(\observation_i | \state)} = \argmin{\state \in \Theta} \mysum{i=1}{n} \loss{\observation_i}.
    \label{eq:ml}
\end{align}
However, real-world data is often corrupted: samples arise from the true distribution $\truedistr(\observation)$ and a corrupting source $q(\observation)$,
\begin{equation}
    \observation_i \sim 
    (1 - \varepsilon)\truedistr(\observation) + \varepsilon q(\observation),
\label{eq:corruption}
\end{equation}
leading to a suboptimal solution $\state_{\text{ML}}$. In this contaminated mixture, corrupted samples {$\observation \sim q(\observation)$} may result from inaccurate measurements, errors or oversight in data acquisition or labeling, or even malicious attacks.
Further, the corrupting distribution $q(\observation)$ is typically unknown, and the level of corruption $\varepsilon$, difficult to determine. 
Equation \eqref{eq:corruption} used herein follows the well-known Huber contamination model from the robust statistics literature \cite{huber_robust_2011,maronna2019robust}.

\begin{figure}[h]
\includegraphics[width=0.65\columnwidth]{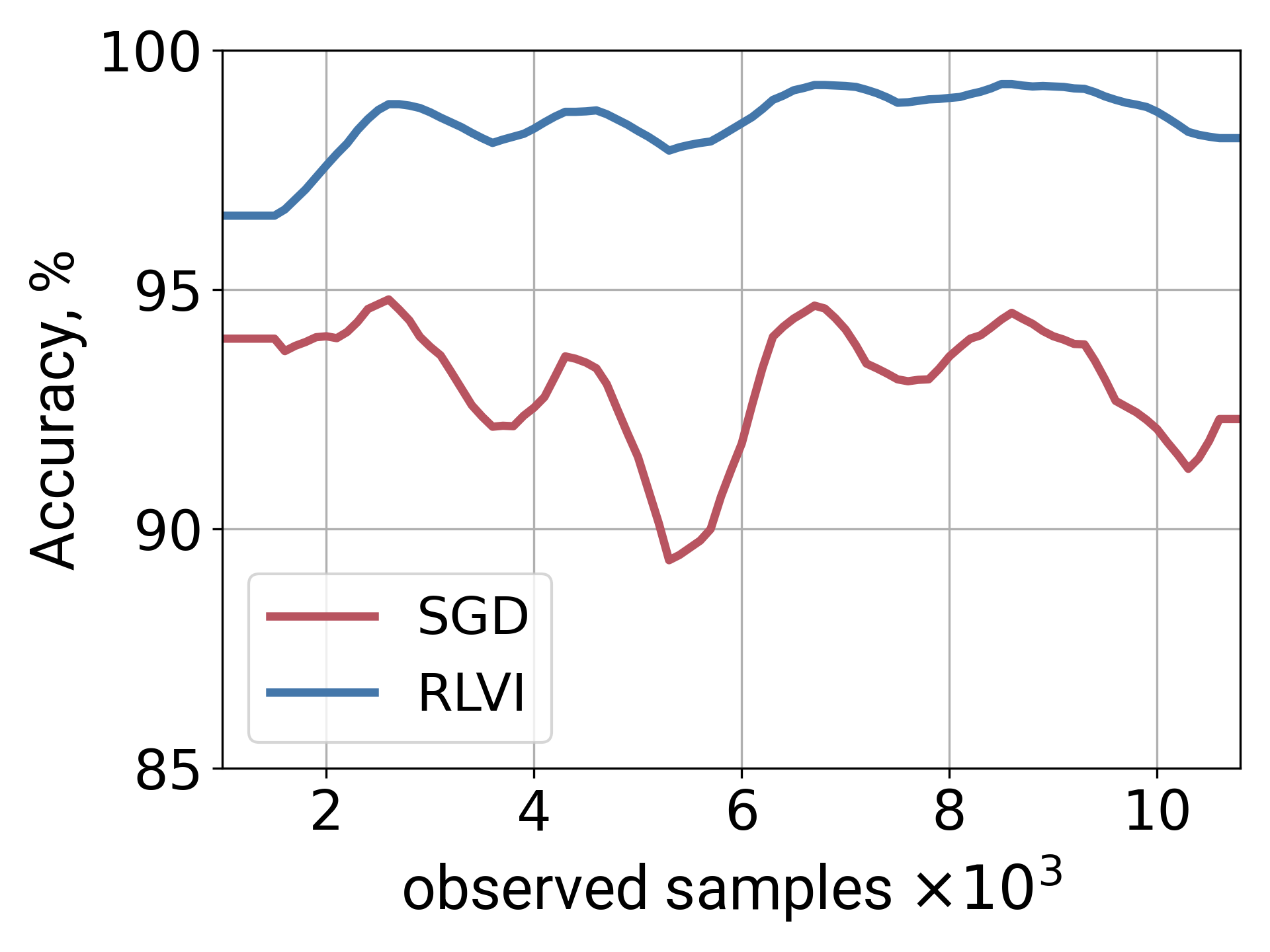}
  \centering
  \caption{Classification of online streaming data with varying number of corrupted labels. Adaptive nature of our approach (RLVI) allows for automatic identification of outliers when learning from batches of data with different $\varepsilon$. Our method is robust and thus has higher accuracy than the standard stochastic optimization of the likelihood (SGD).}
  \label{fig:example}
\end{figure}

Existing approaches require an estimation of the noise structure or the corruption level $\varepsilon$. This imposes restrictions on using these methods in settings where such pre-processing becomes impractical, e.g., in online learning, wherein the assumption about $\varepsilon$ needs to be optimized in continuously arriving data.
In \cref{fig:example}, we show an example: the accuracy of binary classification trained on data that is continuously collected from the HAR dataset \cite{hard}, and subsequently corrupted with varying number of randomly flipped labels in each batch (details in \cref{subsec:online}).

\textbf{Contribution}. 
This paper presents a principled approach for robust learning from corrupted data that is
\begin{itemize}
    \item widely applicable given any likelihood function,
    \item robust against a wide class of contamination sources,
    \item adaptable and tuning-parameter free,
    \item scalable for large data sets and deep learning models.
\end{itemize}

\textbf{Robust Learning via Variational Inference}. 
The proposed approach, denoted as \textsc{RLVI}, employs latent Bernoulli variables to identify corrupted and non-corrupted training samples. The distribution of the latent variables is characterized from training data by maximizing the variational lower bound of marginal likelihood (variational inference) \cite{murphy2023probabilistic}. This allows detection of corrupted samples and computation of the optimal corruption level $\varepsilon$ automatically, subsequently 
learning the model $\state$ in a robust way. The problem formulation entailing \textsc{RLVI} is theoretically simple and also leads to an efficient and straightforward implementation.

\textbf{Related work}. While some approaches address learning from corrupted data with a special loss function, e.g., the Huber loss \cite{huber1992robust}, others construct certain criteria to separate samples from the true distribution and the corrupted ones \cite{bhatia2015robust,bhatia2017consistent}. Recently, new general approaches, such as \textsc{Sever} \cite{diakonikolas2019sever} and Robust Risk Minimization (\textsc{RRM}) \cite{osama2020robust}, were introduced. Both methods are not model-specific. \textsc{Sever} is a general gradient-based learning method that penalizes the gradient components of the loss function corresponding to outliers. \textsc{RRM} involves obtaining sample weights from the constraint on the entropy of a weighted empirical distribution and minimizing the modified empirical risk. Robust Risk Minimization does not require calibrating multiple hyperparameters, unlike \textsc{Sever}.
Despite not requiring the exact value of $\varepsilon$, both approaches are reliant on its upper-bound estimate $\widetilde{\varepsilon} \geq \varepsilon$. 
A very relevant work is \cite{wang2017robust}, where the authors introduce latent variables $\bm{w}$ and raise likelihood terms to these variables to account for the departure from model's assumptions. The paper suggests to impose a prior distribution for $\bm{w}$ that suits the problem at hand and infer $\bm{w}$ together with the model parameters. Thus, being quite general, this framework provides much freedom to users while leaving inference of $\bm{w}$ as a computationally challenging task solved with probabilistic programming. 
In this paper, we rather consider a specific case and define Bernoulli latent variables based on the Huber model \eqref{eq:corruption}, which enables efficient computation and does not require to specify any hyperparameters for a prior distribution.

In the recent past, learning from corrupted data has also received attention in the field of deep learning, and several approaches have been proposed to train a neural network in a robust way. For instance, one can use the correction of the regular loss function \cite{zhang2018generalized,patrini2017making}. 
Alternatively, some approaches directly estimate the noise transition matrix \cite{goldberger2016training,patrini2017making}. Others utilize an additional neural network to identify and prevent overfitting of the initial model, as in \cite{jiang2018mentornet}, and to obtain a better accuracy by exchanging the information between two simultaneously trained networks, as in \cite{han2018co} and \cite{wei2020combating}. 
Approach from \cite{ren2018learning} is based on re-weighting the loss terms for each observation, wherein sample weights are treated as hyperparameters optimized with additional gradient computations.
Another option is to combine techniques that are specific to deep learning: early-stopping \cite{xia2020robust} or dropout \cite{xu2023usdnl}, with a criterion that aims to eliminate corrupted samples. However, training a neural network in the regular context where the contamination issue is not addressed, already requires certain hyperparameters to improve generalization, e.g., learning rate, batch size, number of layers, etc. The aforementioned approaches introduce additional parameters to combat the problem of corrupted labels, including $\varepsilon$, which makes their performance dependent on the efficacy of these parameters and the underlying assumptions.

\section{Problem Formulation}
\label{problem}
Central to our approach is the introduction of a latent variable $t_i$ for each observation $\observation_i \in \Observation$ such that
\begin{align}
    t_i = 
    \begin{cases}
        1,\, \observation_i \sim p(\observation), \\
        0,\, \observation_i \sim q(\observation).
    \end{cases}
\end{align}
Knowing the values of these variables will allow avoiding minimization of the losses on corrupted samples in the dataset in \eqref{eq:ml} by simply dropping the corresponding terms from the sum. 
In other words, latent variables enable us to define a likelihood function with respect to the non-corrupted data, such that
\begin{align}
    p(\Observation | \bm{t}, \state) = \prod_{i=1}^{n} p(\observation_i | \state)^{t_i},
\label{eq:new-ml}
\end{align}
where $\bm{t} = (t_1, t_2, \dots, t_n)$. Optimization with respect to this likelihood function implies a combinatorial search over the latent variables $\bm{t}$, which is computationally infeasible even for moderate $n$.

We observe, however, that the contamination model \eqref{eq:corruption} implies that each sample has a probability $\varepsilon$ of being corrupted. Thus we have the following distribution over $\bm{t}$,
\begin{equation}
    {p(\bm{t} | \varepsilon) = \prod_{i=1}^{n} (1 - \varepsilon)^{t_i} \varepsilon^{1 - t_i}}.
\end{equation}
This prior information enables us to marginalize out the latent variables and obtain the \emph{marginalized} likelihood,
\begin{align}
    p(\Observation | \state, \varepsilon) = \sum_{\bm{t}} p(\Observation | \bm{t}, \state) p(\bm{t} | \varepsilon).
\label{eq:marginal}
\end{align}
Then the maximum marginal likelihood solution,
\begin{align}
    \widehat{\state} = \argmax{\state}{\max_{\varepsilon} p(\Observation | \state, \varepsilon)},
\label{eq:mml}
\end{align}
addresses the problem of robust learning of  $\state$ while obviating the need for specifying $\varepsilon$ or a combinatorial search over $\bm{t}$. We now turn to developing a method to approximately solve \eqref{eq:mml} in an efficient manner.
\section{Method}
We begin by introducing a variational bound on the marginal likelihood in \eqref{eq:mml} and then turn to implementation-specific aspects of the derived method.

\label{method}
\textbf{Variational inference}. To solve the formulated optimization problem, we need a tractable way to compute the sum over $\bm{t}$ in \eqref{eq:marginal}. Using Bayes' rule, we express the marginal likelihood as
\begin{align}
    p(\Observation | \state, \varepsilon) = \dfrac{p(\Observation | \bm{t}, \state) p(\bm{t} | \varepsilon)}{p(\bm{t} | \Observation, \state, \varepsilon)},
\end{align}
where the denominator is the posterior distribution of latent variables $\bm{t}$ given the data $\Observation$, specified model $\state$, and the corruption level $\varepsilon$. As this posterior is intractable, we consider its variational approximation denoted $r(\bm{t} | \weight)$. Specifically, we use a distribution of $n$ independent Bernoulli variables with probabilities $\weight = (\pi_1, \dots, \pi_n)$,
\begin{align}
    r(\bm{t} | \weight) = \prod_{i=1}^{n} \pi_i^{t_i} (1 - \pi_i)^{1 - t_i}.
\label{eq:variationalapproximation}
\end{align}
Therefore, instead of maximizing the marginal likelihood function $p(\Observation | \state, \varepsilon)$ directly, we apply the variational inference framework and optimize the so-called evidence lower-bound (ELBO) \cite{murphy2023probabilistic}. This lower bound has the following general form,
\begin{equation}
    \ln{p(\Observation | \state, \varepsilon)}
    \geq 
    \underbrace{
    \EXD{\ln{p(\Observation | \state, \bm{t})}}{r(\bm{t} | \weight)}
    - \text{KL} \Big[ r(\bm{t} | \weight)\big|\big|p(\bm{t} | \varepsilon) \Big]
    }_{\text{ELBO}(\state, \,\weight, \,\varepsilon)},
\label{eq:elbo}
\end{equation}
where the first term is the expected value of the log-likelihood over latent variables and the second term is the Kullback--Leibler  divergence between the variational approximation and the prior.
The first ELBO term can be rewritten using the loss function $\loss{\observation_i} = -\ln p (\observation_i | \state)$, 
\begin{align}
    \EXD{-\sum_{i=1}^{n} t_i \loss{\observation_i}}{r(\bm{t} | \weight)} = -\mysum{i=1}{n}\pi_i \loss{\observation_i},
\end{align}
where the equality follows from \eqref{eq:variationalapproximation}. This term corresponds to an average loss with non-uniform sample weights, with $\pi_i$ representing the probability of sample $i$ being non-corrupted. 

The second ELBO term consists of the closed form,
\begin{align}
    \text{KL} \Big[ r\,\big|\big|\,p \Big] = \mysum{i=1}{n}\pi_i \ln \frac{\pi_i}{1 - \varepsilon} + (1 - \pi_i) \ln \frac{1 - \pi_i}{\varepsilon}.
\end{align}
Note that corruption level does not appear in the first term of the ELBO and we can optimize 
$\varepsilon$ directly and independently of the model $\state$:
\begin{gather}
    \varepsilon = \arg\max \text{ELBO}(\state, \weight, \varepsilon) 
     = 1 - \frac{1}{n}\sum\limits_{i=1}^n\pi_i.
\end{gather}
That is, variational inference framework resolves an unknown hyperparameter by explicitly optimizing the corruption level. Moreover, we get an intuitively satisfying result that parameters $\pi_i$ defining the probability of each sample being non-corrupted, should sum up to the number of non-corrupted samples:
\begin{equation}
    n (1 - \varepsilon) = \mysum{i=1}{n} \pi_i.
\end{equation}
Hence we arrive at the following objective -- the negative ELBO, expressed as
\begin{equation}
    \elbo{\state, \weight} = \mysum{i=1}{n}\pi_i \loss{\observation_i}
    + \pi_i \ln \frac{\pi_i}{\langle \weight \rangle}
    + (1 - \pi_i) \ln \frac{1 - \pi_i}{1 - \langle \weight \rangle},
\label{objective}
\end{equation}
where $\langle \weight \rangle := \sum_{i=1}^n \pi_i \,/\, n$ is the average of Bernoulli probabilities.

Consequently, the objective in \eqref{eq:mml} is replaced by the optimum of its variational bound, i.e., 
\begin{equation}
\boxed{
    \state_{\text{RLVI}} = \argmin{\state \in \Theta}{\min_{\weight \in (0;1)^n}{\elbo{\state, \weight}}}.
}
\label{eq:optimization}
\end{equation}

\textbf{Numerical optimization}. The resulting optimization problem is solved using the block-wise \cref{alg:em}, which follows the general Expectation-Maximization (\textsc{EM}) scheme \cite{bishop2006pattern}.  The E-step consists of minimizing $\mathcal{L}(\state, \weight)$ in $\weight$ for fixed parameters $\state$. In the M-step, the inferred probabilities are used to maximize the re-scaled log-likelihood, $-\sum_{i=1}^n \pi_i \loss{\observation_i}$. As an example, in case of linear regression the latter step solves a weighted least-squares problem. For a classification task, we minimize the cross-entropy loss corrected with sample weights.

Note that the E-step can be performed efficiently, as the objective $\mathcal{L}(\state, \weight)$ is convex in $\weight$. Therefore, to find the optimal parameters $\weight$ for a fixed model, the derivative of $\elbo{\state, \weight}$ is equated to zero w.r.t. $\pi_j$ for all $j = 1, \dots, n$,
\begin{align}
    \dfrac{\partial \mathcal{L}}{\partial \pi_j} = 0 \iff 
    \pi_j = \left( 1 + \dfrac{1 - \langle \weight \rangle}{\langle \weight \rangle}e^{ \loss{\observation_j} } \right)^{-1}.
\label{eq:stationary}
\end{align}
This is a system of nonlinear equations for $\weight$, which we solve with the fixed-point iterations,
\begin{align}
    \pi_j^{\text{new}} = \left( 1 + \dfrac{1 - \langle \weight \rangle^\text{old}}{\langle \weight \rangle^\text{old}}e^{\loss{\observation_j}} \right)^{-1}.
\label{eq:fixed-point}
\end{align}
In \eqref{eq:fixed-point}, for the mean weight, we use the value from the previous fixed-point iteration and compute $\pi_j^{\text{new}}$ independently with simple vector operations that scale well for large data. A proof for $\elbo{\state, \weight}$ being convex in $\weight$ and iterations \eqref{eq:fixed-point} converging to a stationary point is provided in the Appendix.

\begin{algorithm}[h]
    \setstretch{1.35}
    \caption{\textsc{RLVI}: robust learning from corrupted data}
    \label{alg:em}
 \begin{algorithmic}[1]
    \STATE {\bfseries Input:} data $\Observation = \{\observation_i\}_{i=1}^n$
    \STATE $\state^{0}  \leftarrow \argmin{\state \in \Theta}{\sum_{i=1}^{n} \loss{\observation_i}}$ 
    \hfill\COMMENT{$\pi^0_i = 1$}
    \FOR{$k = 1, 2, \dots$}
        \STATE Evaluate $\loss{\observation_i}$ for each $\observation_i \in \Observation$ using $\state^{k-1}$
        \STATE $\weight^{k} \leftarrow$ fixed-point iterations \eqref{eq:fixed-point} 
        \hfill\COMMENT{\textit{E-step}}
        \STATE $\state^{k}  \leftarrow \argmin{\state \in \Theta}{\sum_{i=1}^{n} \pi^{k}_i \loss{\observation_i}}$ \hfill\COMMENT{\textit{M-step}}
        \IF{$\lVert \state^{k} - \state^{k-1} \rVert\leq tolerance$} 
            \STATE {\bfseries return} $\state^{k}$
        \ENDIF
    \ENDFOR
 \end{algorithmic}
 \end{algorithm}

\textbf{Stochastic approximation}. The formulated objective $\mathcal{L}(\state, \weight)$ alleviates the need to know or assess $\varepsilon$, which is particularly useful when $\varepsilon$ is not fixed, such as in the online learning setting. Another useful property of this variational bound is its structure: with respect to $\state$, the function consists of $n$ independent terms, just as the standard likelihood function, which can be employed by stochastic optimization of $\mathcal{L}(\state, \weight)$. If one does not have access to the full dataset but only to its batches, the M-step in \cref{alg:em} can be replaced with a stochastic gradient descent (SGD) update
\begin{equation}
{\state^{k}=\state^{k-1} -\alpha \sum_{i=1}^{b} \pi_i^{k} \nabla\Loss{\state}{\observation_i}},
\label{eq:sgd}
\end{equation}
where the loss function and, therefore, the gradient components for each batch of $b$ samples $\observation_i$ are re-weighted with $\pi^{k}_i, \, i = 1, \dots, b$. The latter are computed using the corresponding loss values $\loss{\observation_i}$ at the preceding E-step with \eqref{eq:fixed-point}. 
Batch minimization of $\mathcal{L}(\state, \weight)$, in effect, corresponds to a stochastic variant of the EM algorithm -- see, e.g., stepwise EM in \cite{murphy2023probabilistic}.

\textbf{Truncation as a form of regularization}. Since the objective in RLVI is suitable for stochastic optimization, our robust learning approach can also be used in deep learning, where SGD-like approaches are dominant. However, neural networks are overparameterized models and thus
are prone to overfitting, which in theory (and in practice: see \cref{fig:regularization}) hinders the performance of RLVI.
Indeed, 
if we substitute zero loss for all training samples into the stationary point condition \eqref{eq:stationary},
we find the corresponding minimum for all $i = 1, \dots, n$,
\begin{align}
&\pi^{\star}_i = \left( 1 + \frac{\varepsilon}{1 - \varepsilon}e^{ \loss{\observation_i} } \right)^{-1} \Bigg|_{\loss{\observation_i} = 0} = 1 - \varepsilon.
\end{align}
That is, when overfitting commences, the marginal likelihood approach treats all samples as non-corrupted since overparameterized model is capable of minimizing loss to zero on both `clean' and corrupted samples. 
Nevertheless, as is generally the case, to prevent overfitting, one can use regularization of the loss function. 
In this work, we introduce regularization to the RLVI algorithm,
making the algorithm effective in the overparameterized regime as well. 
Namely, samples that have a low probability of being non-corrupt are eliminated from SGD updates:
\begin{equation}
    \pi_i < \tau \implies \pi_i \leftarrow 0.
\label{eq:truncation}
\end{equation}
Furthermore, we define the threshold $\tau$ based on the following criterion: maximize the number of samples to be used for learning (maximize $\tau$) subject to a bounded type II error (number of corrupted samples treated as `clean'). 
Hence, 
$\tau = \max \, \{\pi_1, \pi_2, \dots, \pi_n\}$, such that
\begin{equation}
    \frac{\mathbb{E}_r \Big[\# \text{False Clean}\Big]}{\mathbb{E}_r \Big[\# \text{Corrupted \Big] }}
    = \frac{\mysum{i=1}{n} (1 - \pi_i) \mathbbm{1}[\pi_i \geq \tau]}{\mysum{i=1}{n} (1 - \pi_i)} \leq 0.05.
\label{eq:error-ineq}
\end{equation}
In this criterion, the admissable type II error is common for statistical hypothesis testing and equals 5\%. Also note how the obtained posterior approximation is employed: the numerator is the expected number of corrupted samples considered as `clean', and the denominator is the expected total number of corrupted samples, based on $r(\bm{t} | \weight)$.

The resulting variant of RLVI, to be used for overparameterized models, is listed as \cref{alg:mml-sgd}.
It implements RLVI as stochastic gradient optimization of neural network parameters $\state$. Parameters $\pi_i$ are updated at the end of each epoch using efficient iterations~\eqref{eq:fixed-point}. To prevent model's overfitting to corrupted samples, the algorithm eliminates gradient terms corresponding to low $\pi_i$. Threshold for truncation is re-computed across epochs as a non-decreasing value from the type II error criterion~\eqref{eq:error-ineq}. Note that Bernoulli probabilites are updated once at each epoch, which makes solution for $\weight$ less dependent on the batch size in this case.
 
\section{Experiments}
\label{experiments}
The proposed RLVI method is compared to existing approaches in three problem settings: standard parameter estimation, online learning, and deep learning\footnote{Implementation of RLVI and our experiments are available at \url{https://github.com/akarakulev/rlvi}.}. 

\begin{figure*}[ht]
    \centering
  \begin{minipage}[t]{.25\linewidth}
    \includegraphics[width=\linewidth]{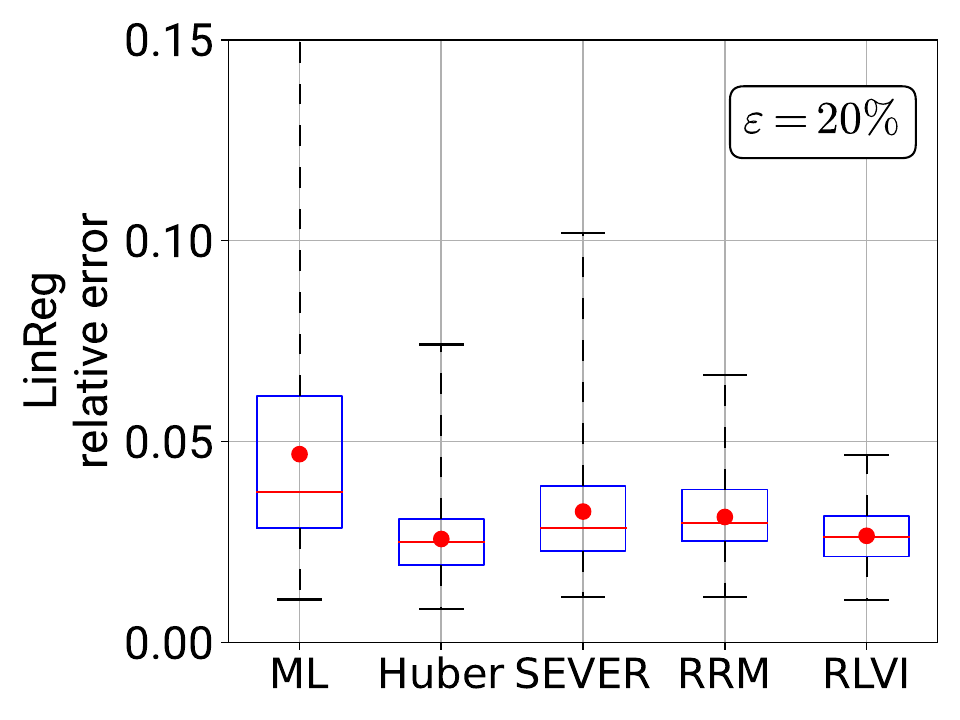}
  \end{minipage} \quad
  \begin{minipage}[t]{.25\linewidth}
    \includegraphics[width=\linewidth]{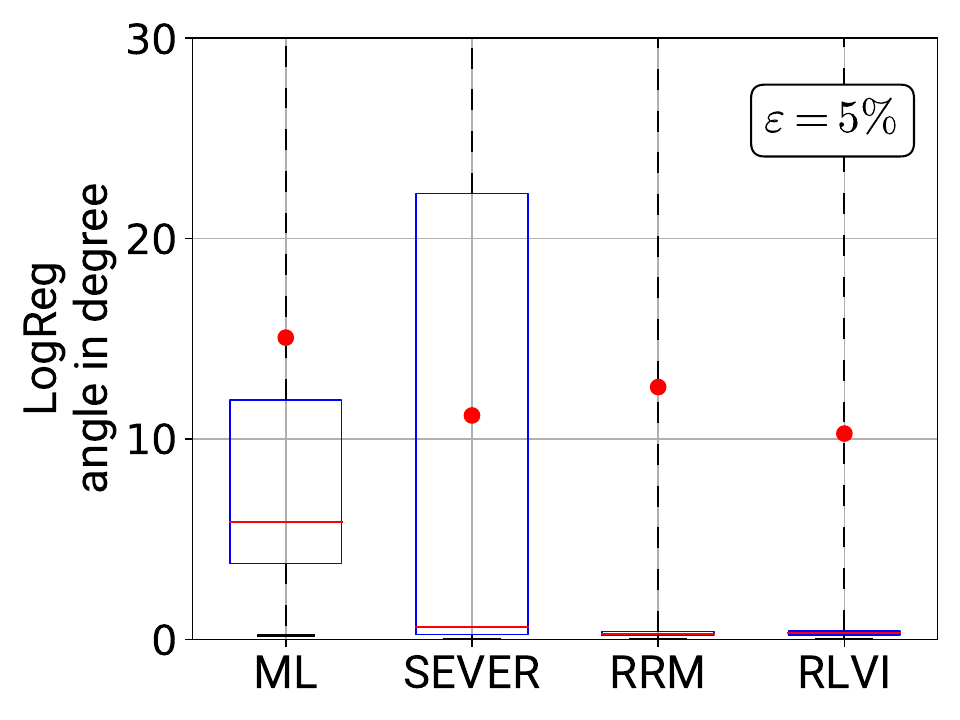}
  \end{minipage} \quad
  \begin{minipage}[t]{.25\linewidth}
    \includegraphics[width=\linewidth]{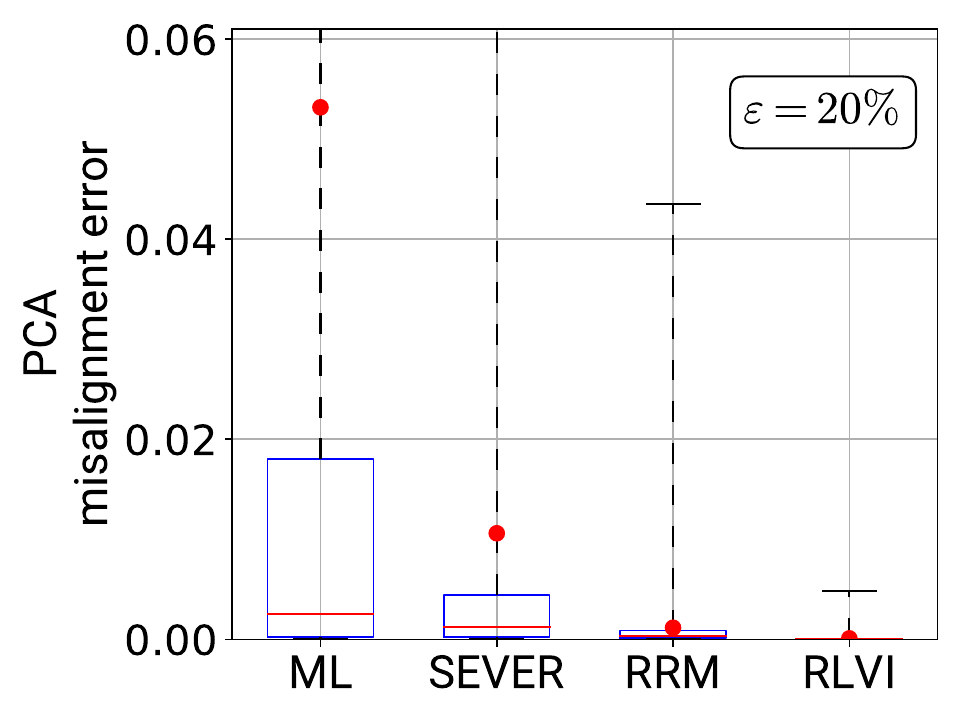}
  \end{minipage}  
  \caption{
    Box plots of relative errors for a fixed value of corruption level $\varepsilon$.
    \textit{Left}. Linear regression: relative errors ${\| \widehat{\state}-\beststate \|_2 / \| \beststate \|_2}$.
    \textit{Middle}. Logistic regression: angle in degrees between the true separating hyperplane $\beststate$ and estimates $\widehat{\state}$.
    \textit{Right}. PCA: misalignment errors $1 - |\cos{( \widehat{\state}\,^\top \beststate )}|$ for the subspace spanned by the first principal component.
    Each box spans the 25th to 75th quantiles; red dots depict the means. For all plots, 100 Monte Carlo runs are used.
    }
  \label{fig:mc-experiments}
\end{figure*}

\subsection{Benchmark on standard learning problems}
\label{subsec:standard}

\begin{figure}[H]
    \centering
    \begin{minipage}[t]{0.65\columnwidth}
      \centering
      \includegraphics[width=\linewidth]{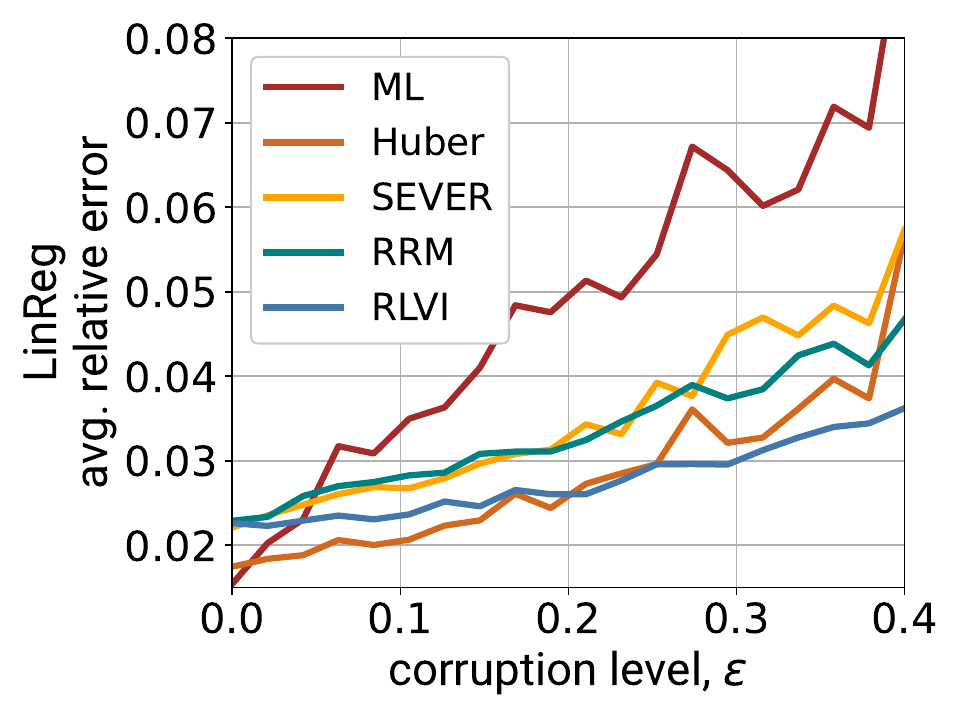}
    \end{minipage}
  \caption{Linear regression. Average relative error versus varying corruption level $\varepsilon$; 100 Monte Carlo runs are used.}
    \label{fig:linear-regression}
\end{figure}

First, we demonstrate that our method is applicable to different maximum likelihood problems and achieves higher model accuracy over contemporary alternatives. Thereto, we reproduce the experiments from \cite{osama2020robust} comparing various algorithms for robust learning on three problems:
linear regression, logistic regression, and dimensionality reduction using principal component analysis (PCA).
Since each problem can be formulated as likelihood maximization, we define the corresponding loss functions as the negative log-likelihood.
In the experiments, synthetic data ($n$ samples) is generated from the mixture of $\truedistr(\observation)$ and $q(\observation)$ with a fixed ratio of corrupted samples $\varepsilon$ using the specified model $\beststate$. The optimal model $\widehat{\state}$ is estimated using different algorithms: standard ML, \textsc{Huber} \cite{zoubir2018robust}, \textsc{Sever} \cite{diakonikolas2019sever}, \textsc{RRM} \cite{osama2020robust}, and RLVI. 
The reader is referred to \cite{osama2020robust} for more details on the three test problems, including specifics of distributions used as true and corrupted.
Subsequently, we perform 100 Monte Carlo runs and plot the statistics for corresponding errors with boxplots in \cref{fig:mc-experiments}. For linear regression the results, shown in \cref{fig:linear-regression}, also include average relative error for $\varepsilon$ varying in $[0; 0.4]$. Again, 100 Monte Carlo runs are used for each fixed $\varepsilon$.

\begin{table}[h]
\centering
\renewcommand{\arraystretch}{1.1}
\resizebox{0.9\columnwidth}{!}{
\begin{tabular}{l|P{30mm}|c|c|c}
    Problem & dimension & $n$ & $\varepsilon$ & $\widetilde{\varepsilon}$ \\ \hline
    LinReg  & $\state \in \mathbb{R}^{10}$ & 40 & 0.2 & 0.4 \\ \hline
    LogReg & $\state \in \mathbb{R}^3$ & 100 & 0.05 & 0.3 \\ \hline
    PCA & $\state \in \{\mathbb{R}^2: \lVert \state \rVert = 1\}$ & 40 & 0.2 & 0.4 \\ 
\end{tabular}
}
\caption{Experiments reproduced from \cite{osama2020robust}:
$\state$ is a parametric model learned from synthetic data ($n$ samples in total, $\varepsilon n$ are corrupted); $\widetilde{\varepsilon} \geq {\varepsilon}$ is the upper-bound used for \textsc{Huber}, \textsc{Sever}, and \textsc{RRM}.}
\label{tab:benchmark}
\end{table}
\raggedbottom

Note that using the unbounded likelihood for defining $\loss{\observation}$ within RLVI can lead to a degenerate solution and thus might require additional regularization -- see an example with a covariance estimation problem in the Appendix.

From \cref{fig:mc-experiments,fig:linear-regression}, one can see that RLVI achieves better average accuracy and tighter confidence intervals than the competing methods. 

\subsection{Online learning}
\label{subsec:online}
To further evaluate RLVI's adaptivity, we apply it to online learning -- the setting where $\varepsilon$ changes dynamically.
Online learning is used when dealing with, for example, signals from remote sensors, user clicks on a website, daily weather conditions, etc. In such cases, data is being collected sequentially and the model is incrementally learned from each new set of observations. This allows continuous refinement of the model and out-of-core inference when the dataset is too large to be stored and handled entirely. 

As described in Section \ref{method}, the objective $\mathcal{L}(\state, \weight)$ in RLVI allows for an incremental learning scheme by replacing the M-step of \textsc{EM} algorithm with an update of stochastic gradient descent: ${\state^{k}=\state^{k-1} -\alpha \sum_{i=1}^{b} \pi_i^{k} \nabla\Loss{\state}{\observation_i}}$. Here, parameters $\weight^{k}$ are computed at the E-step with~\eqref{eq:fixed-point}. The computational overhead over the standard stochastic likelihood maximization is not significant: the fixed-point algorithm only involves $O(b)$ vectorized operations at each SGD step.

\begin{figure}[H]
    \centering
    \begin{minipage}[t]{0.5\columnwidth}
      \includegraphics[width=\linewidth]{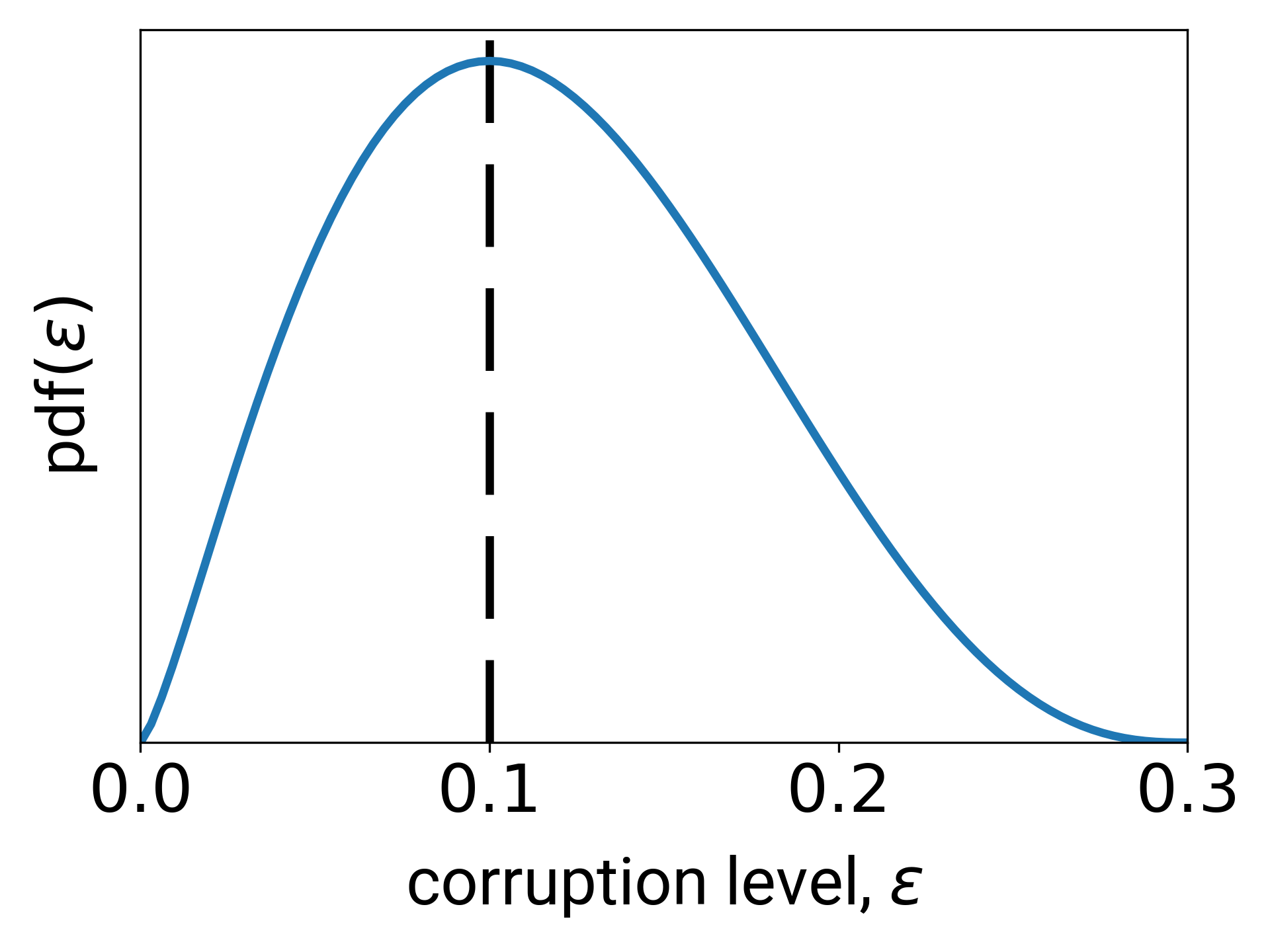}
    \end{minipage}\hfill
    \begin{minipage}[t]{0.5\columnwidth}
      \includegraphics[width=\linewidth]{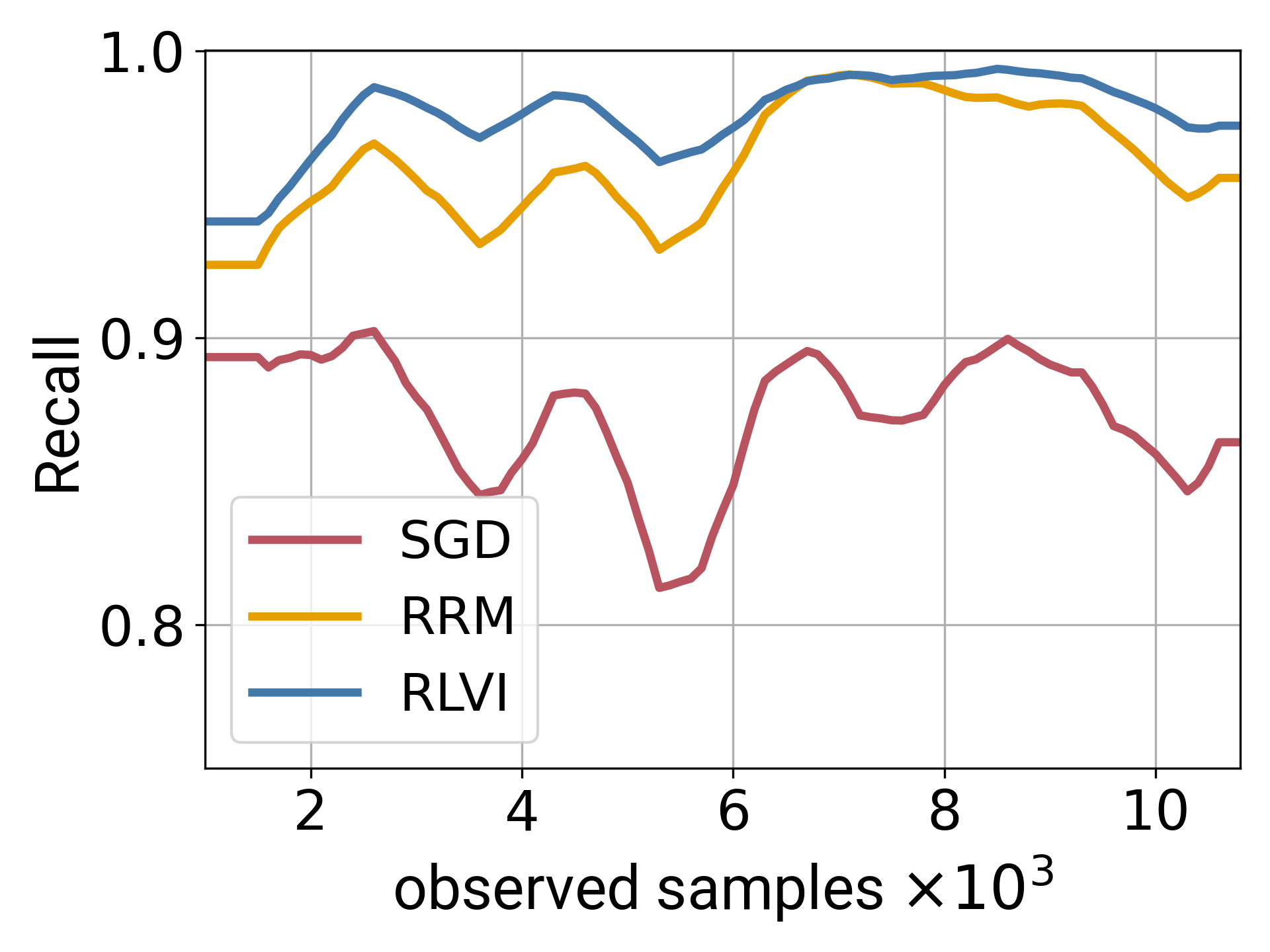}
    \end{minipage}\hfill
  \caption{Online classification.
    \textit{Left}. Distribution of corruption level across batches of streaming data.
    \textit{Right}. Recall (true positive rate) on left-out data versus total number of observed samples (smoothed with a moving average filter).
    }
    \label{fig:incr-learn}
\end{figure}

We consider the Human Activity Recognition dataset from \cite{hard} containing $24,075$ measurements from smartphone sensors. Each measurement $\observation_i = (\bm{x}_i, y_i)$ consists of $\bm{x}_i \in \mathbb{R}^{60}$ features extracted from accelerometers during different human activities: Sitting, Standing, Walking, Running, and Dancing. We perform binary classification and towards that end, partition the five initial labels into two:
resting state ($y = 0$) and active state ($y = 1$). To simulate a data stream, at each iteration we retrieve data in batches of size $2b$, where $b = 100$ samples are used for training and another $b = 100$ samples serve performance evaluation.
To introduce noise, we corrupt the data in each training batch by randomly flipping $\varepsilon$ percent of positive labels. Moreover, the corruption level $\varepsilon$ changes in each iteration. Since we are testing the robustness of RLVI against corruption, we focus specifically on variation of noise -- not on the gradual change of $\state$ (concept drift) or other possible challenges related more to the method to learn $\state$ incrementally.
For each batch, $\varepsilon$ is sampled from a linearly transformed Beta distribution (called PERT \cite{johnson1995continuous}) that is simple to parameterize with three values, so that the samples are within the interval $[\varepsilon_{\text{min}}, \varepsilon_{\text{max}}]$ with the mode $\varepsilon_{\text{mode}}$. To simulate a typical case, we set ${\varepsilon_{\text{min}} = 0}$, ${\varepsilon_{\text{max}} = 0.3}$, and ${\varepsilon_{\text{mode}} = 0.1}$. Thus, the ratio of corrupted labels in each batch varies according to the distribution in \cref{fig:incr-learn}.

Since \textsc{RRM} is also based on block-wise optimization, it can similarly be implemented as incremental learning, where inference for $\state$ consists of anti-gradient steps minimizing the modified empirical risk for each batch. Hence, the experiments compare three versions of online classification: the standard stochastic maximization of the likelihood (SGD), incremental minimization of the risk function defined in \textsc{RRM} \cite{osama2020robust} with a threshold for $\varepsilon$ set to ${\varepsilon_\text{max} = 0.3}$, and the proposed approach, RLVI.

Classification performance is evaluated with recall (true positive rate) computed for 100 test samples in each iteration. \cref{fig:incr-learn} shows the evolution of recall smoothed with a moving average filter with a window of 10 batches (also, see the smoothed accuracy curve in \cref{fig:example}). It can be seen that standard SGD clearly suffers from label corruption, with \textsc{RRM} being affected to a lesser extent. RLVI attains consistently higher accuracy and recall values as it does not depend on a global estimate of the noise magnitude, and robustly evaluates it from incoming data.

\subsection{Overparameterized model}
\label{subsec:overparameterized}
We now extend RLVI to learning an overparameterized model and consider image classification using a convolutional neural network when training labels are corrupted.

\textbf{Existing approaches}. State-of-the-art performance is achieved in this setting by methods that identify and distill corrupted samples based on some criterion. The algorithm Co-teaching \cite{han2018co} is based on training two neural networks in parallel and learning only on the samples that attain a small loss for both models. The~{JoCoR} approach \cite{wei2020combating} also trains two models but aims to reduce the diversity between their predictions. Due to simultaneous training of two models, Co-teaching and {JoCoR} effectively double the computational time. CDR \cite{xia2020robust} trains one model -- it employs weight decay to diminish the impact of network parameters that overfit to corrupted samples, and early-stopping using a validation set. The recent algorithm USDNL \cite{xu2023usdnl} estimates prediction uncertainty for training samples using dropout, thus identifying the samples with corrupted labels. 
Each of the above methods combines its own criterion for corrupted samples with a schedule to gradually consider fewer samples and thus account for model overfitting occurring in later epochs. The schedule is a non-decreasing function defined with the corruption level $\varepsilon$ (assumed to be known in advance) so that, by the end of training, only $(1 - \varepsilon)$ ratio of the training set is being used. The algorithm~{BARE} from \cite{patel2023adaptive} aims to robustly train neural networks for classification independently of $\varepsilon$. It removes samples from the loss function using batch statistics, assuming that the class conditional noise has a special structure. 

\textbf{Regularization}. As discussed in Section \ref{method}, highly overparameterized models overfit to all samples, including the corrupted ones. To prevent overfitting within RLVI, we use hard truncation $\eqref{eq:truncation}$, where truncation boundary $\tau$ is defined by the type II error criterion \eqref{eq:error-ineq}.
Furthermore, as studied empirically \cite{jiang2018mentornet,nguyen2019self,xia2020robust}, neural networks overfit in later epochs. 
Thus, regularization $\eqref{eq:truncation}$ is applied after the model starts to overfit, which can be identified by a decrease in prediction accuracy on a contaminated validation set. 

\begin{figure}[h]
    \centering
    \begin{minipage}[t]{.65\linewidth}
      \includegraphics[width=\linewidth]{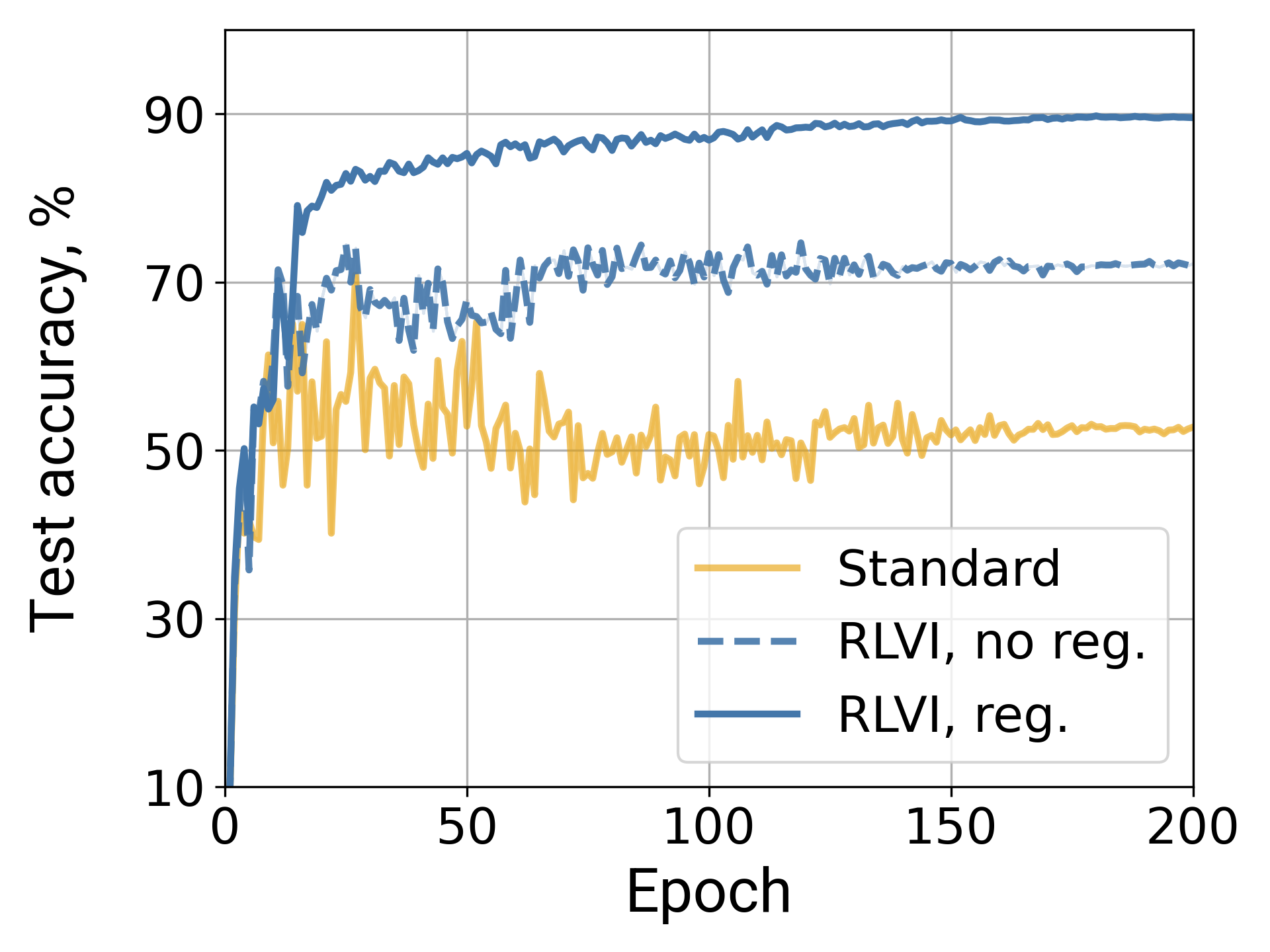}
    \end{minipage}\hfill\vfill
    \begin{minipage}[t]{.65\linewidth}
      \includegraphics[width=\linewidth]{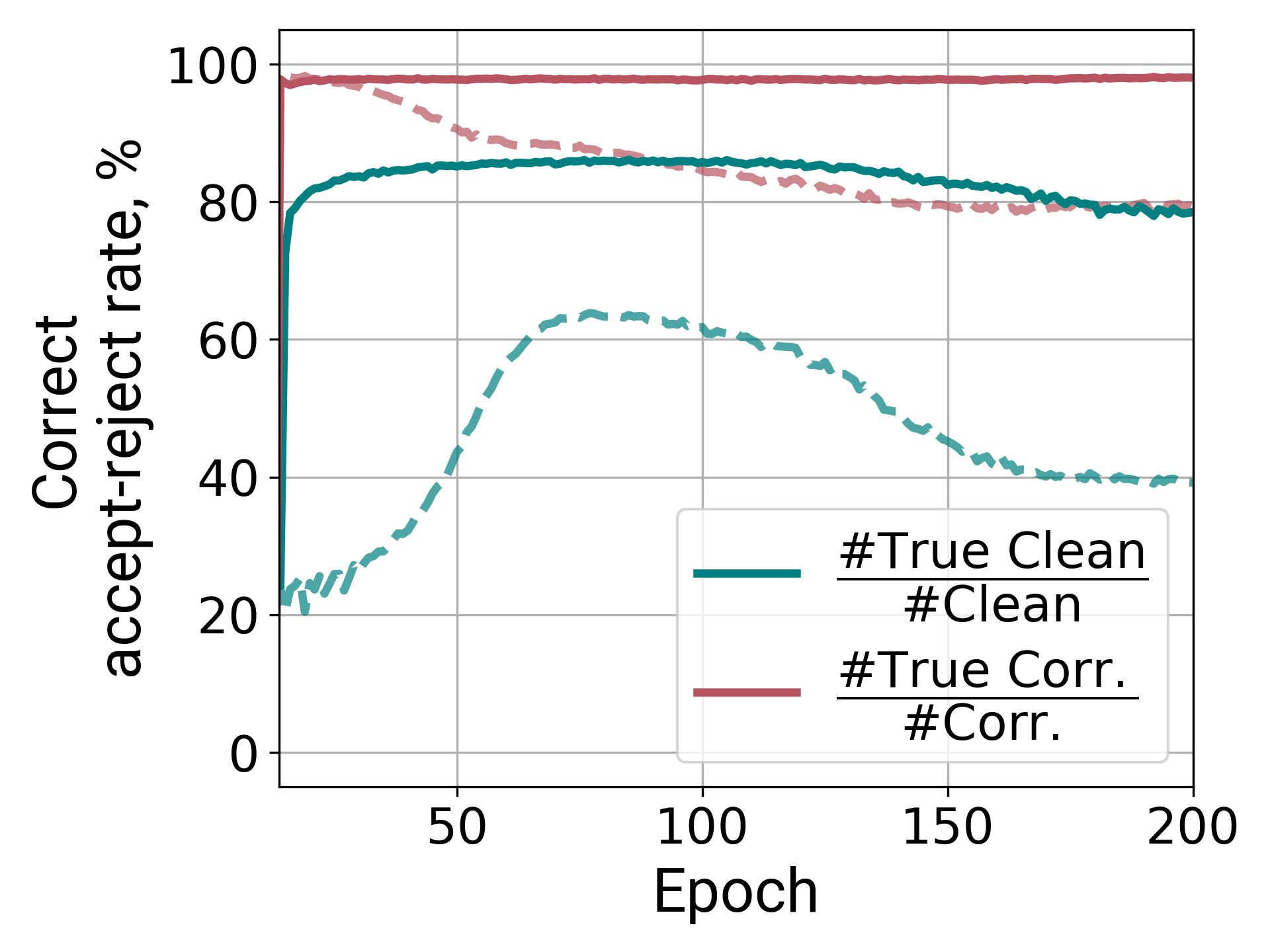}
    \end{minipage}\hfill
  \caption{
    Image classification on CIFAR10 corrupted with synthetic noise (pairflip, $\varepsilon = 45\%$).
    \textit{Top}. Accuracy on the clean testing set for standard SGD and RLVI.
    \textit{Bottom}. Percentage of corrupted and non-corrupted samples correctly identified with the decision boundary ${\pi_i < \tau}$.
    In both plots, the dashed line corresponds to RLVI with no regularization -- solid line indicates that truncation (${\pi_i < \tau \implies \pi_i \leftarrow 0}$) is used for the terms in anti-gradient updates.
    Threshold $\tau$ is computed from \eqref{eq:error-ineq}. Regularization based on the bounded type II error makes differentiation of corrupted samples more effective, ultimately improving test accuracy for the overparameterized setting.
    }
    \label{fig:regularization}
\end{figure}

\cref{fig:regularization} presents an example of image classification on CIFAR10 with randomly flipped labels.
It illustrates how the introduced regularization functions in practice: if no regularization is used, the proportion of identified corrupted observations decreases while model gradually fits $q(\observation)$ despite marginal likelihood formulation, thus attaining lower test accuracy. But with regularization,
differentiation of corrupted and non-corrupted data points according to $\pi_i$ is more effective during all iterations, which leads to better generalization. 
In the Appendix, we provide similar plots for various types and levels of noise.

\begin{algorithm}[h]
    \setstretch{1.25}
    \caption{RLVI: robust training of neural networks}
    \label{alg:mml-sgd}
    \begin{algorithmic}[1]
        \STATE {\bfseries Input:} training set $\Observation_{tr} = \{\observation_i\}_{i=1}^n$, noisy validation set $\Observation_{val}$, learning rate $\alpha$, batch size $b$
        \STATE $\state \leftarrow$ initialize neural network parameters 
        \STATE $\pi_i \leftarrow 1, \; i = 1, \dots, n$
        \STATE $L \leftarrow$ empty array of size $n$ 
        \hfill\COMMENT{to store $\loss{\observation_i}, \observation_i \in \Observation_{tr}$}
        \STATE $overfit$ $\leftarrow$ False
        \STATE $\tau \leftarrow 0$
        \FOR{$epoch = 1, 2, \dots, n_{\text{epochs}}$}
            \STATE {\bfseries for $\mathcal{I}_b \sim U\{1, \dots, n\}$}
            \hfill\COMMENT{for a batch of $b$ indices}
                \STATE \quad $L_i \leftarrow \loss{\observation_i}, \;i \in \mathcal{I}_b$ 
                \hfill\COMMENT{store loss value}
                \STATE \quad $\state \leftarrow \state - \alpha \sum_{i \in \mathcal{I}_b} \pi_i \nabla\Loss{\state}{\observation_i}$
            \STATE $\weight \leftarrow$ fixed-point \eqref{eq:fixed-point} using loss values in $L$
            \STATE {\bfseries if $overfit$} 
                \STATE \quad $\tau^{\star} \leftarrow \max \{\pi_1, \dots, \pi_n\}$, s.t. error bound \eqref{eq:error-ineq}
                \STATE \quad $\tau \leftarrow \max(\tau, \tau^{\star})$ 
                \hfill\COMMENT{can only increase}
                \STATE \quad $\pi_i \leftarrow \pi_i$ \textbf{if} $\pi_i \geq \tau$ \textbf{else} $0$ \hfill\COMMENT{regularization}
            \STATE {\bfseries else if} accuracy on $\Observation_{val}$ dropped
                \STATE \quad $overfit \leftarrow$ True 
        \ENDFOR
        \STATE {\bfseries Output:} $\state$
    \end{algorithmic}
\end{algorithm}

\textbf{Synthetic corruption}. To test the algorithm, we conduct the experiments in image classification using convolutional neural networks and cross-entropy loss as $\ell_{\state}$. The datasets being used are MNIST \cite{lecun1998mnist}, that contains hand-written digits, and CIFAR10 and CIFAR100
\cite{krizhevsky2009learning} -- both containing the same images that are classified into ten and one hundred categories respectively. Subsequently, the training labels in these datasets are corrupted with four types of synthetic noise: symmetric, asymmetric, and pairflip (class-dependent) and instance (feature-dependent). These noise types have been commonly used in previous works \cite{han2018co,wei2020combating,xia2020robust,xu2023usdnl}. Additionally, we perform the experiment in which data contains both the corrupted labels and the out-of-distribution samples.
To this end, we consider the dataset CIFAR80N-O \cite{yao2021jo} which is obtained from CIFAR100 as follows: the last 20 classes in CIFAR100 are regarded as out-of-distribution images, and images from the remaining 80 classes are subsequently corrupted by one of the class-dependent synthetic noise types (pairflip, symmetric, and asymmetric).
Thereto, we demonstrate that, owing to the generality of Huber contamination model \eqref{eq:corruption}, RLVI can be succesfully applied to learning in the presence of noise of arbitrary structure. 
In these experiments, we employ commonly used hyperparameter settings found in literature specific to each dataset and model architecture. These settings can be found in \cref{tab:dl-configs} of the Appendix.

We compare the attained classification accuracy on the test set with the standard likelihood maximization approach and recently proposed alternative methods: Co-teaching, JoCoR, CDR, USDNL, and BARE. For CDR and RLVI, 10\% of the training data is used as a validation set:
in RLVI, we apply regularization \eqref{eq:truncation} after validation accuracy at the current epoch becomes less than the average of its two previous values.
In contrast, CDR uses a validation set for early-stopping: optimization concludes if the validation accuracy exceeds some specified threshold. The latter implies that for CDR we report the test accuracy corresponding to the lowest validation loss. 
For all the alternative methods, we use their default hyperparameters related to robust learning, including the same schedule for the ratio of considered samples during epochs, as defined in \cite{han2018co}, deduced from the true noise level employed.
Also, since USDNL is based on uncertainty estimation using dropout, for USDNL specifically, we used variants of the corresponding neural nets with dropout layers, where dropout rate was set to 0.25, as in the original paper \cite{xu2023usdnl}.

\begin{table*}[t]
    \caption{
    Test accuracy ($\%$) after training on corrupted datasets: mean $\pm$ standard deviation over five random initializations. 
    }
    \centering
    \resizebox{0.85\linewidth}{!}{\begin{tabular}{llcccccccc}
       \cmidrule[1pt]{1-10}
       \multirow{2}{*}{Dataset} & \multirow{2}{*}{Method} & \multicolumn{2}{c}{Symmetric} & \multicolumn{2}{c}{Asymmetric} & \multicolumn{2}{c}{Pairflip} & \multicolumn{2}{c}{Instance} \\
        \cmidrule{3-10}
        & & 20\% & 45\% & 20\% & 45\% & 20\% & 45\% & 20\% & 45\% \\
        \midrule
        & Standard & 95.66$\pm$0.28 & 87.47$\pm$0.82 & 98.29$\pm$0.14  & 87.73$\pm$0.65 & 97.21$\pm$0.41 & 69.26$\pm$3.65 & 95.66$\pm$0.36 & 71.98$\pm$1.43\\
        & Co-teaching & 96.62$\pm$0.08  & 96.68$\pm$0.07 & 95.94$\pm$0.06 & 93.40$\pm$0.42 & 96.19$\pm$0.10 & 92.91$\pm$0.43 & 96.49$\pm$0.23 & 95.39$\pm$0.59\\
        \multirow{2}{*}{MNIST} 
        & JoCoR & 99.07$\pm$0.05  & 98.38$\pm$0.10 & 99.02$\pm$0.05  & 95.21$\pm$3.70 & 98.96$\pm$0.09 & 91.06$\pm$5.01 & 99.03$\pm$0.05 & 97.81$\pm$0.35\\
        & CDR & 98.81$\pm$0.09 & 98.27$\pm$0.09 & 99.14$\pm$0.05  & 94.35$\pm$1.29 & 98.95$\pm$0.06 & 87.55$\pm$1.26 & 98.25$\pm$0.16 & 88.27$\pm$1.92\\
        & USDNL & 98.38$\pm$0.12  & 97.72$\pm$0.13 & 98.32$\pm$0.12  & 95.83$\pm$0.64 & 98.23$\pm$0.09 & 89.94$\pm$1.62 & 98.13$\pm$0.11 & 96.52$\pm$0.38\\
        & BARE & 99.07$\pm$0.11 & \textbf{98.78}$\bm{\pm}$\textbf{0.10} & \textbf{99.16}$\bm{\pm}$\textbf{0.06} & \textbf{98.70}$\bm{\pm}$\textbf{0.12} & 99.05$\pm$0.06 & 98.22$\pm$0.21 & 99.05$\pm$0.07 & \textbf{98.42}$\bm{\pm}$\textbf{0.45}\\
        \cmidrule{2-10}
        & RLVI & \textbf{99.10}$\bm{\pm}$\textbf{0.06} & \textbf{98.70}$\bm{\pm}$\textbf{0.18} & 98.90$\pm$0.22  & \textbf{98.69}$\bm{\pm}$\textbf{0.05} & \textbf{99.10}$\bm{\pm}$\textbf{0.07} & \textbf{98.52}$\bm{\pm}$\textbf{0.07} & \textbf{99.12}$\bm{\pm}$\textbf{0.03} & 98.38$\pm$0.08\\
        \midrule
        & Standard & 83.36$\pm$0.35 & 60.22$\pm$0.28 & 87.26$\pm$0.40  & 75.03$\pm$0.28 & 81.14$\pm$0.43 & 52.30$\pm$0.85 & 81.76$\pm$0.37 & 55.51$\pm$1.04\\
        & Co-teaching & 87.55$\pm$0.38 & 83.70$\pm$0.46 & 86.98$\pm$0.26 & 64.84$\pm$0.87 & 86.57$\pm$0.21 & 67.38$\pm$2.93 & 86.40$\pm$0.29 & 66.39$\pm$7.16\\
        \multirow{2}{*}{CIFAR10} 
        & JoCoR & 90.42$\pm$0.07 & 86.33$\pm$0.23 & 90.60$\pm$0.19 & 76.94$\pm$2.56 & 89.25$\pm$0.30 & 72.53$\pm$3.57 & 89.08$\pm$0.36 & 79.48$\pm$1.37\\
        & CDR & 85.57$\pm$0.38 & 77.83$\pm$0.30 & 87.98$\pm$0.61 & 75.99$\pm$1.95 & 87.34$\pm$0.34 & 68.13$\pm$2.03 & 85.72$\pm$0.74 & 66.82$\pm$2.48\\
        & USDNL & 88.65$\pm$0.21 & 83.33$\pm$0.23 & 88.36$\pm$0.33  & 74.62$\pm$0.70 & 87.05$\pm$0.24 & 66.49$\pm$2.45 & 86.90$\pm$0.18 & 71.04$\pm$4.17\\
        & BARE & 85.67$\pm$0.46 & 69.90$\pm$2.06 & 87.38$\pm$0.24 & \textbf{77.82}$\bm{\pm}$\textbf{1.10} & 84.84$\pm$0.77 & 57.18$\pm$2.81 & 85.08$\pm$0.73 & 69.00$\pm$2.05 \\
        \cmidrule{2-10}
        & RLVI & \textbf{92.30}$\bm{\pm}$\textbf{0.14} & \textbf{88.69}$\bm{\pm}$\textbf{0.33} & \textbf{91.73}$\bm{\pm}$\textbf{0.12} & 76.96$\pm$0.50 & \textbf{92.15}$\bm{\pm}$\textbf{0.27} & \textbf{89.13}$\bm{\pm}$\textbf{0.29} & \textbf{91.97}$\bm{\pm}$\textbf{0.33} & \textbf{85.15}$\bm{\pm}$\textbf{1.57}\\
        \midrule
        & Standard & 61.96$\pm$0.11 & 42.36$\pm$0.80 & 62.88$\pm$0.20  & 39.91$\pm$0.58 & 62.62$\pm$0.53 & 38.94$\pm$0.42 & 62.87$\pm$0.32 & 43.17$\pm$0.63\\
        & Co-teaching & 58.74$\pm$0.66 & 48.11$\pm$1.01 & 54.32$\pm$0.22  & 34.53$\pm$0.68 & 56.27$\pm$0.60 & 34.47$\pm$0.97 & 57.44$\pm$0.32 & 34.97$\pm$0.61 \\
        \multirow{2}{*}{CIFAR100}
        & JoCoR & \textbf{69.43}$\bm{\pm}$\textbf{0.25} & 62.37$\pm$0.94 & 62.93$\pm$0.60 & 38.68$\pm$0.83 & 65.90$\pm$0.35 & 40.94$\pm$0.70 & 67.98$\pm$0.32 & 52.47$\pm$0.31\\
        & CDR & 61.57$\pm$0.41 & 46.31$\pm$0.81 & 62.89$\pm$0.30  & 39.47$\pm$0.75 & 61.27$\pm$2.23 & 38.55$\pm$0.22 & 62.09$\pm$0.45 & 41.80$\pm$0.77\\
        & USDNL & 64.96$\pm$0.56 & 53.82$\pm$1.15 & 59.12$\pm$0.10  & 36.44$\pm$0.93 & 61.76$\pm$1.01 & 35.81$\pm$0.20 & 63.13$\pm$0.54 & 45.42$\pm$1.30\\
        & BARE & 59.59$\pm$1.12 & 46.56$\pm$1.10 & 52.91$\pm$1.28 & 29.48$\pm$1.12 & 53.29$\pm$1.57 & 30.27$\pm$1.14 & 56.47$\pm$0.84 & 37.24$\pm$1.27 \\
        \cmidrule{2-10}
        & RLVI & \textbf{69.64}$\bm{\pm}$\textbf{0.55} & \textbf{64.11}$\bm{\pm}$\textbf{0.79} & \textbf{69.33}$\bm{\pm}$\textbf{0.83}  & \textbf{55.76}$\bm{\pm}$\textbf{2.12} & \textbf{69.25}$\bm{\pm}$\textbf{0.77} & \textbf{55.77}$\bm{\pm}$\textbf{1.01} & \textbf{69.54}$\bm{\pm}$\textbf{0.84} & \textbf{62.00}$\bm{\pm}$\textbf{1.29}\\
        \midrule
        & Standard & 59.41$\pm$0.40	& 37.84$\pm$0.79 & 61.01$\pm$0.37 & 39.13$\pm$0.26 & 60.88$\pm$0.33 & 38.88$\pm$0.52\\
        & Co-teaching & 59.77$\pm$0.67 & 48.45$\pm$1.44 & 55.64$\pm$0.90 & 35.98$\pm$0.82 & 58.44$\pm$1.27 & 35.61$\pm$0.47\\
        \multirow{2}{*}{CIFAR80N-O}
        & JoCoR & 70.02$\pm$0.80 &	62.29$\pm$0.41 & 64.12$\pm$0.19	& 40.13$\pm$0.40 & 67.11$\pm$0.83 & 41.85$\pm$0.29\\
        & CDR & 56.08$\pm$0.98 & 44.46$\pm$1.03	& 58.16$\pm$1.47 &	36.76$\pm$1.28 & 57.96$\pm$0.57 &	36.16$\pm$1.41\\
        & USDNL & 64.07$\pm$1.63 & 52.00$\pm$2.89 & 59.43$\pm$0.69 & 37.04$\pm$0.46 & 62.20$\pm$0.52 & 37.12$\pm$1.09\\
        & BARE & 57.39$\pm$1.23 &	42.56$\pm$2.20	& 54.47$\pm$1.01 & 30.12$\pm$2.09 & 55.21$\pm$1.28 & 30.09$\pm$0.86 \\
        \cmidrule{2-8}
        & RLVI & $\textbf{71.13}\bm{\pm}\textbf{0.71}$ & $\textbf{63.18}\bm{\pm}\textbf{0.36}$ & $\textbf{71.96}\bm{\pm}\textbf{0.39}$ & $\textbf{54.49}\bm{\pm}\textbf{1.76}$	& $\textbf{71.45}\bm{\pm}\textbf{0.33}$ & $\textbf{56.12}\bm{\pm}\textbf{0.23}$\\
        \cmidrule[1pt]{1-8}
    \end{tabular}}
    \label{tab:dl-benchmark}
\end{table*}

Results are presented as mean $\pm$ standard deviation over five runs with random initialization of network's parameters. 
\cref{tab:dl-benchmark} shows that RLVI, with the described regularization, attains results competitive with alternative approaches.
Also note that additional steps in \cref{alg:mml-sgd} involving $\weight$ (fixed-point iterations and truncation) do not significantly increase the computational time compared to standard SGD.
\cref{tab:time} shows the average time per epoch during training.
\begin{table}[H]
    \caption{
    Time per one epoch in seconds when training on CIFAR100: mean $\pm$ standard deviation over 200 epochs.
    \centering (Standard: 8.82$\pm$0.08)
    }
    \centering
    \resizebox{0.6\columnwidth}{!}{
    \begin{tabular}{cccccc}
        \toprule
        Co-teaching & JoCoR & CDR\\
        \midrule
        17.91$\pm$0.35 & 18.11$\pm$0.11 & 13.88$\pm$0.35 \\
        \midrule
        \midrule
        USDNL & BARE & RLVI \\
        \midrule
        9.40$\pm$0.22 & 9.94 $\pm$ 0.10 & 10.52$\pm$0.56 \\
        \bottomrule
    \end{tabular}}
    \label{tab:time}
\end{table}

\textbf{Real corruption}. To demonstrate how RLVI performs in a naturally contaminated 
setting, we train ResNet50 -- pre-trained on ImageNet \cite{rw2019timm} -- on the challenging dataset Food101 \cite{bossard2014food} consisting of 101 food categories. For each class, 250 testing images were cleaned manually, while the remaining 750 training images per class still contain corrupted labels. We use the Adam optimizer with hyperparameters listed in \cref{tab:dl-configs}. 

\begin{figure}[H]
    \centering
    \includegraphics[width=0.9\columnwidth]{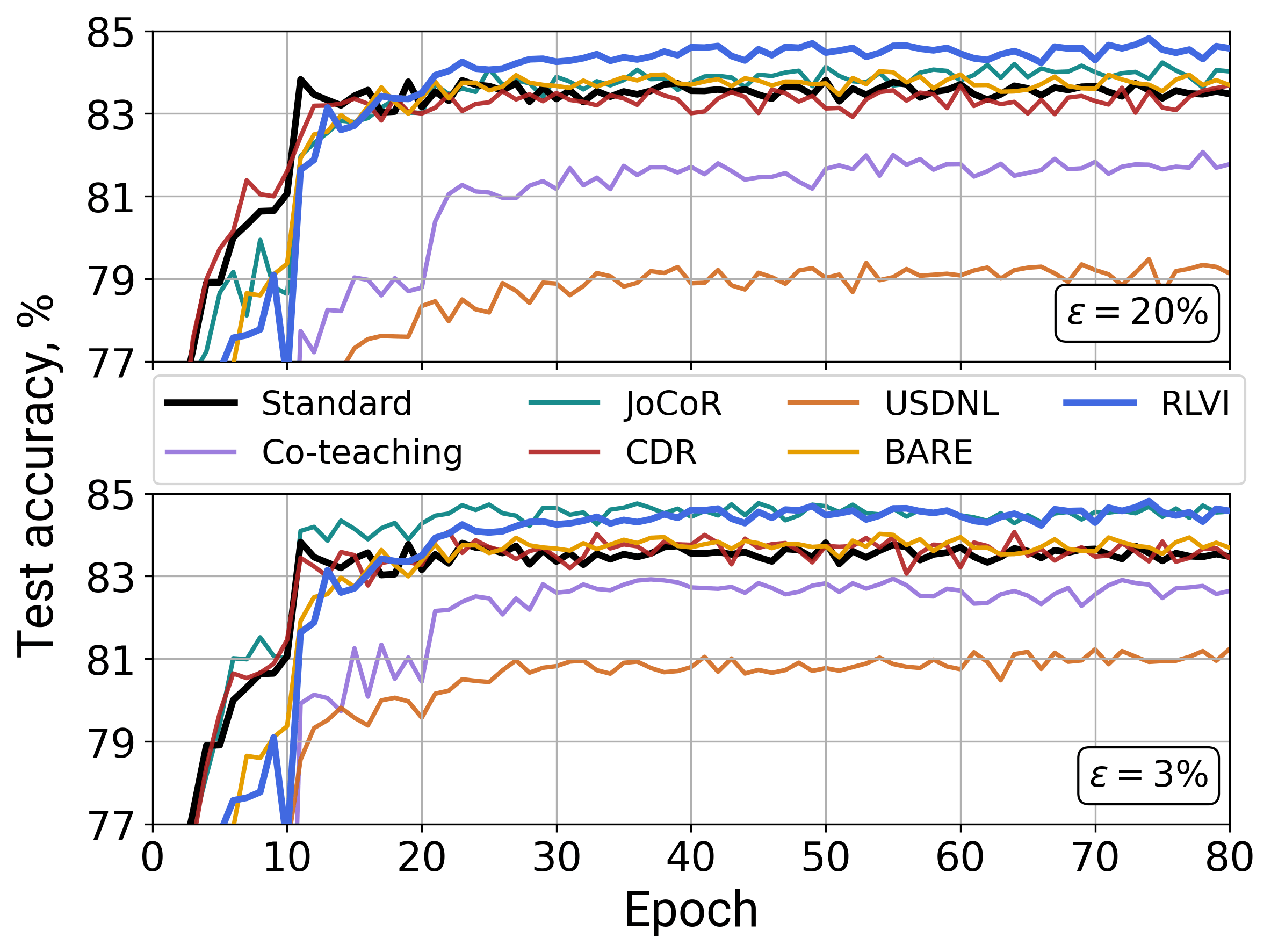}
  \caption{
    Food101. Test accuracy assuming high and low $\varepsilon$.
    }
    \label{fig:food101}
\end{figure}

In the real setting, $\varepsilon$ is unknown and has to be optimized for Co-teaching, JoCoR, CDR, and USDNL.
\cref{fig:food101} shows that by varying the estimated $\varepsilon$ for alternative methods (from a large 20\% -- to a moderate 3\% value), one can enhance performance. In contrast, RLVI improves over the standard approach without the need for additional optimization of the hyperparameter. In Appendix, we provide results for other assumptions on $\varepsilon$.

\setlength{\belowcaptionskip}{-7pt}

\section{Conclusions}
\label{conclusions}
We presented the novel robust learning algorithm \textsc{RLVI}
for likelihood maximization problems with corrupted datasets.
It leverages variational inference to identify corrupted samples under the Huber contamination model using latent Bernoulli variables. This alleviates the need for specifying hyperparameters such as the corruption level, which existing approaches rely on. RLVI can also be implemented as stochastic optimization, which makes it adaptive and applicable to learning from data with varying noise and out-of-core inference for large datasets. 
We demonstrated the effectiveness of the method on benchmark test problems in both -- traditional statistical learning, as well as online and deep learning settings. The proposed RLVI algorithm meets or exceeds performance across considered experimental settings in a parameter-free and efficient manner.

\section*{Acknowledgements}
\label{acknowledgements}
The authors are thankful to Prof. Peter Stoica for insightful discussions during the work on this article. The computations/data handling were enabled by the Berzelius resource provided by the Knut and Alice Wallenberg Foundation at the National Supercomputer Centre and by the National Academic Infrastructure for Supercomputing in Sweden (NAISS) at Chalmers e-Commons at Chalmers,
and Uppsala Multidisciplinary Center for Advanced Computational Science (UPPMAX) at Uppsala University, partially funded by the Swedish Research Council through grant agreement nos. 2022-06725 and 2018-05973. DZ and PS acknowledge support from the Swedish Research Council through grant agreement nos. 2018-05040 and 2023-05593 respectively.
\section*{Impact Statement}
\label{impact}
We introduce a novel approach for robust machine learning in likelihood maximization settings. The approach will potentially make robust machine learning easier to apply, owing to its parameter-free and general nature. There are many potential societal consequences of our work, none which we feel must be specifically highlighted here.

{
    \small
    \bibliographystyle{icml2024}
    \bibliography{main}
}

\newpage
\onecolumn
    \appendix
\section*{A. Optimization of Bernoulli probabilities}
\label{convexity}
\subsection*{A1. Convexity}
Objective $\elbo{\state, \weight}$ defined in \cref{objective} is a convex function of variables $\weight \in (0; 1)^n$.
\begin{proof}
    The objective can be viewed as a sum of three terms,
    \begin{align}
        \elbo{\state, \weight} = \mysum{i=1}{n}\pi_i \loss{\observation_i}
            + \mysum{i=1}{n} \pi_i \ln \frac{\pi_i}{\langle \weight \rangle}
            + \mysum{i=1}{n} (1 - \pi_i) \ln \frac{1 - \pi_i}{1 - \langle \weight \rangle}.
    \end{align}
    The first term is linear in $\weight$, and we only need to verify if the second and the third terms are convex.
    Therefore, we first focus on the second term and its Hessian. Using $\one$ for an $n$-dimensional vector of ones and $\textbf{diag}(\cdot)$ for a diagonal matrix, we write
    \begin{align}
        f(\weight) &:= \mysum{i=1}{n} \pi_i \ln{\dfrac{\pi_i}{\langle \weight \rangle}}, \\
        \nabla^2 f(\weight) &= \textbf{diag} \left( \dfrac{1}{\pi_i} \right) - \dfrac{\one \cdot \one^\top}{n \langle \weight \rangle}.
    \end{align}
    For $f(\weight)$ to be convex, its Hessian has to be positive semi-definite: $\nabla^2 f \succcurlyeq 0$. For clarity we multiply $\nabla^2 f$ by a positive scalar $n \langle \weight \rangle$ and consider a matrix
    \begin{align}
        n \langle \weight \rangle \nabla^2 f = \textbf{diag}\left( \dfrac{n \langle \weight \rangle}{\pi_i} \right) - \one \cdot \one^\top = \bm{D} - \one \cdot \one^\top.
    \end{align}
    Notice that $\bm{D} - \one \cdot \one^\top$ is a Schur complement of the block matrix
    \begin{align}
        \begin{pmatrix}
            \bm{D} & \one \\
            \one^\top & 1
        \end{pmatrix}.
    \end{align}
    And, since $\bm{D} \succ 0$, the following inequalities should hold simultaneously by the properties of the Schur complement:
    \begin{align}
        &\bm{D} - \one \cdot \one^\top \succcurlyeq 0& 
        \iff& 
        &\begin{pmatrix}
            \bm{D} & \one \\
            \one^\top & 1
        \end{pmatrix} \succcurlyeq 0&
        &\iff&
        &1 - \one^\top \bm{D}^{-1}\one \geq 0.
    \end{align}
    However, expression on the right holds true. Indeed,
    \begin{align}
        1 - \one^\top \bm{D}^{-1}\one = 1 - \one^\top \textbf{diag} \left( \dfrac{\pi_i}{n \langle \weight \rangle} \right) \one = 1 - \dfrac{1}{n \langle \weight \rangle} \mysum{i=1}{n} \pi_i = 1 - \dfrac{\langle \weight \rangle}{\langle \weight \rangle} = 0.
    \end{align}
    Therefore, we conclude that $\nabla^2 f \succcurlyeq 0$ and the second term in $\elbo{\state, \weight}$ is convex. To establish the convexity of the third term, it suffices to consider $1 - \pi_i$ as its variables, which makes the proof identical to the above. Hence $\elbo{\state, \weight}$ is convex in $\weight$ as a sum of convex functions.
    \end{proof}

\subsection*{A2. Convergence}
To see why iterations \eqref{eq:fixed-point} converge to a minimizer of $\elbo{\state, \weight}$ in Bernoulli probabilities, consider the following. If we over-parameterize $\elbo{\state, \weight}$ and let $\pi_\text{o} := \langle \weight \rangle$ be a free variable, then the objective becomes a function of separate variables $\weight$ and $\pi_\text{o}$. \cref{eq:stationary} defines its closed-form minimizer with respect to $\weight$ due to convexity. Whereas the stationary point condition in terms of $\pi_\text{o}$ results in $\pi_\text{o} = \sum_{i=1}^n \pi_i \,/\, n$. Thus iterations $\eqref{eq:fixed-point}$ essentially implement the coordinate descent for overparameterized objective in $\weight$ and $\pi_\text{o}$ and hence converge to a minimizer $\weight^\star$ of $\elbo{\state, \weight}$.

\section*{B. Online learning with varying level of noise}
\label{online-results}
\begin{figure}[H]
    \centering
    \begin{minipage}[t]{0.35\textwidth}
      \includegraphics[width=\linewidth]{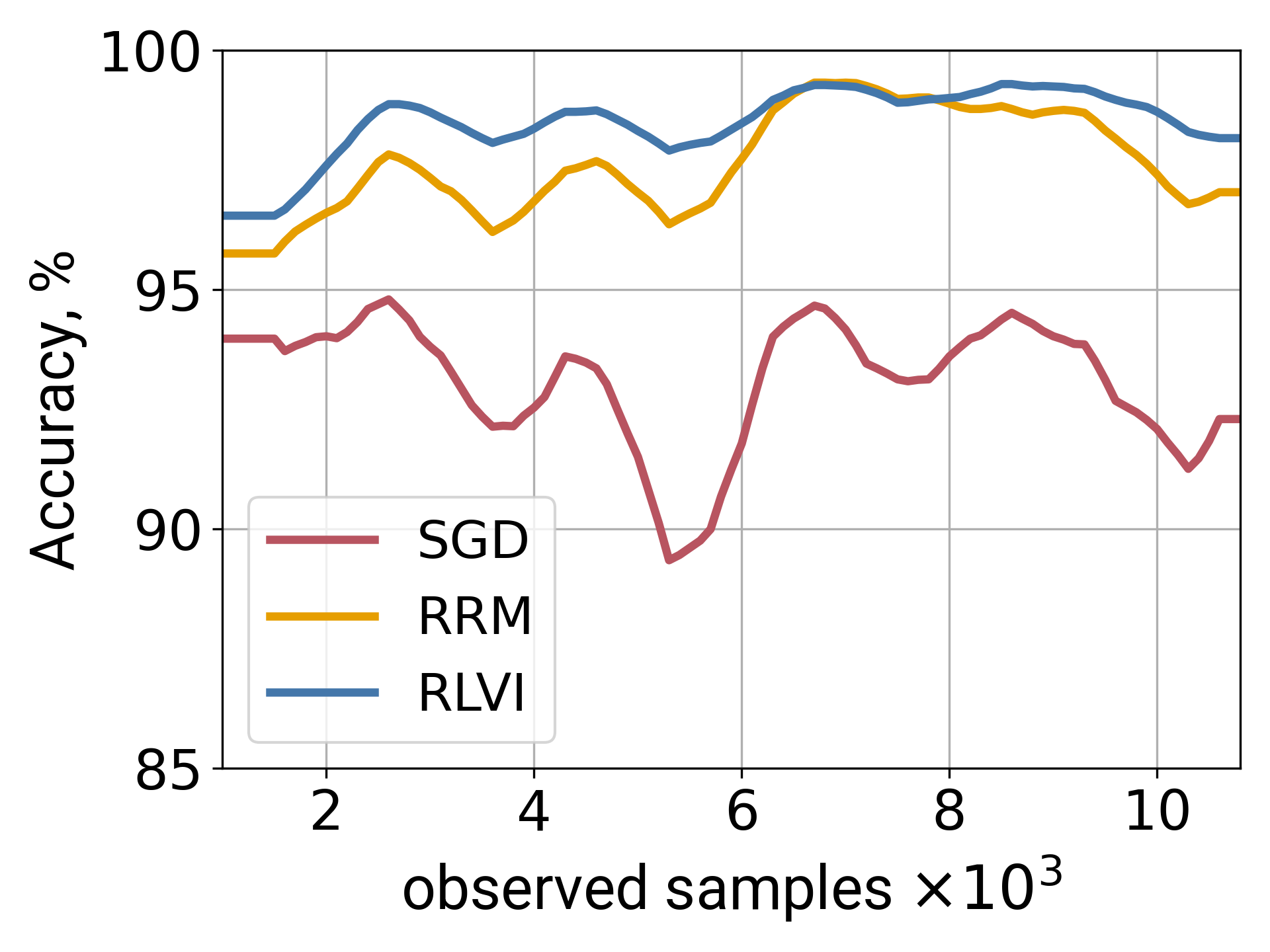}
    \end{minipage}~
    \begin{minipage}[t]{0.35\textwidth}
      \includegraphics[width=\linewidth]{hard_tp.png}
    \end{minipage}\hfill
  \caption{Online classification.
    Accuracy and recall when learning with the batch size 100.
    }
    \label{fig:online-results}
\end{figure}

In \cref{fig:online-results}, we provide the performance metrics of the standard SGD approach, RRM, and RLVI, used for binary classification of data that arrives in batches with varying number of corrupted labels. In this figure, we show both accuracy and recall values for all three methods. The metrics are smoothed with a moving average filter.

\begin{figure}[H]
    \centering
    \begin{minipage}[t]{0.35\textwidth}
      \includegraphics[width=\linewidth]{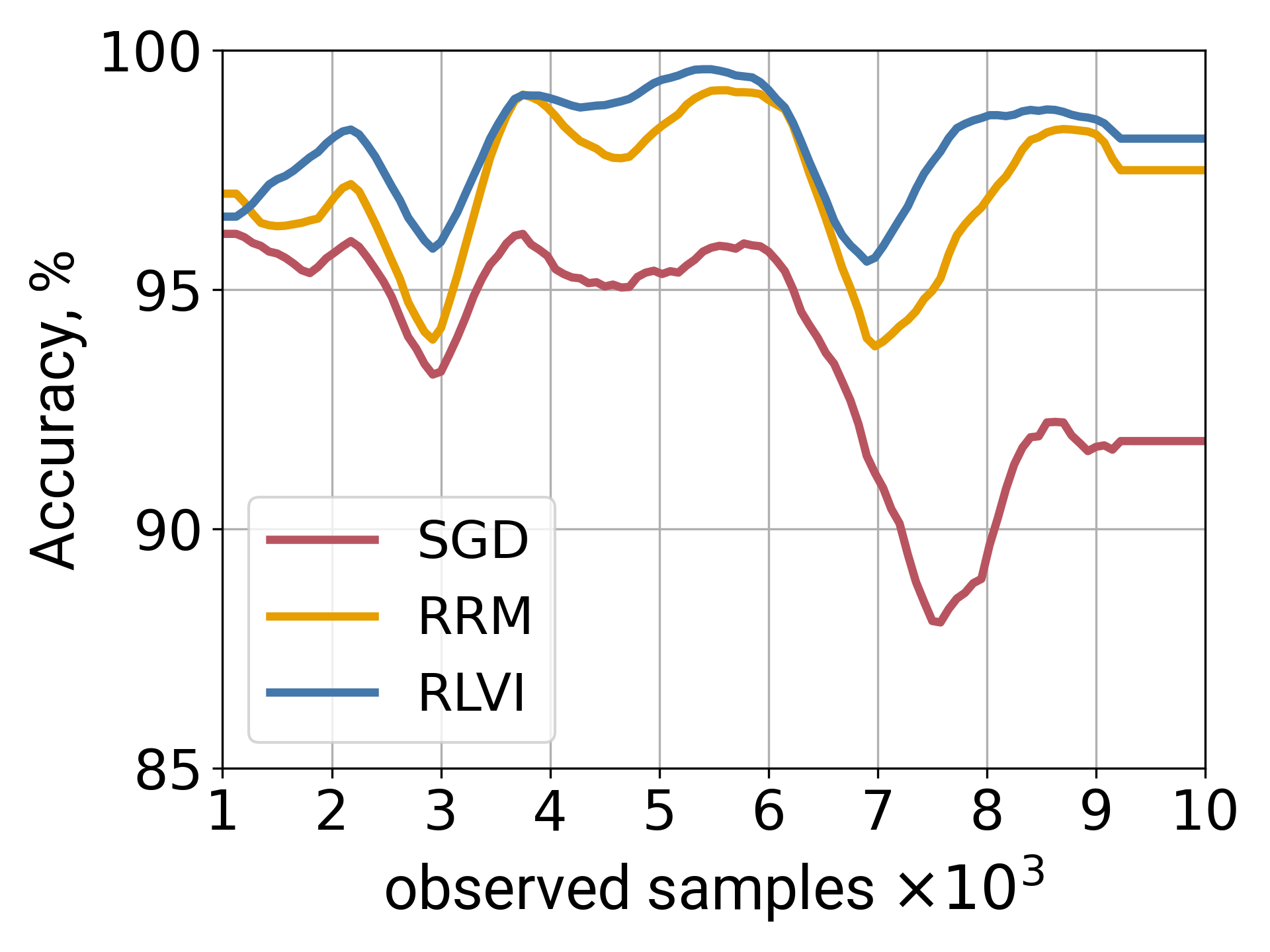}
    \end{minipage}~
    \begin{minipage}[t]{0.35\textwidth}
      \includegraphics[width=\linewidth]{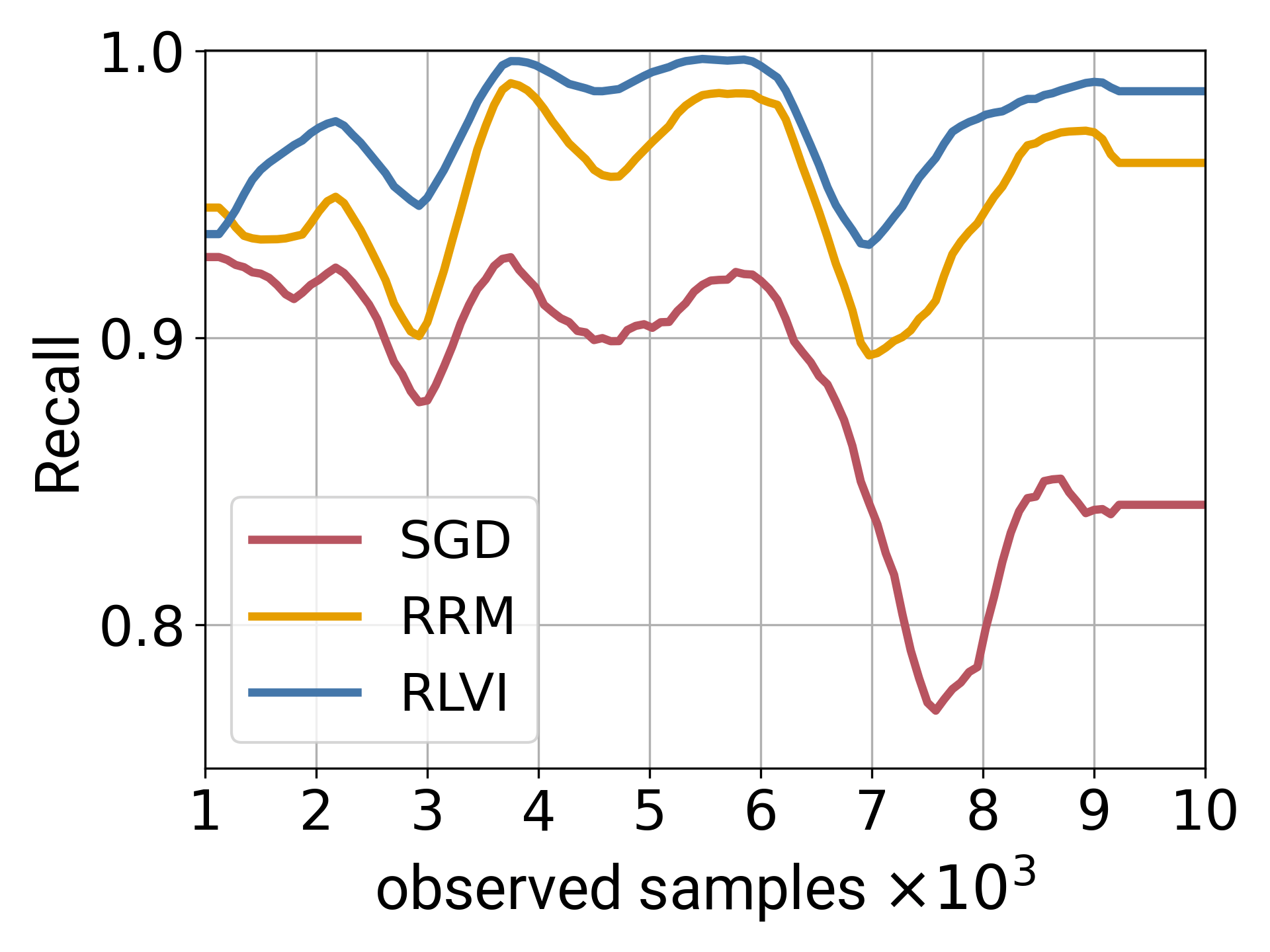}
    \end{minipage}\hfill
  \caption{Online classification.
    Accuracy and recall when learning with the batch size 75.
    }
    \label{fig:online-results-bs75}
\end{figure}

\begin{figure}[H]
    \centering
    \begin{minipage}[t]{0.35\textwidth}
      \includegraphics[width=\linewidth]{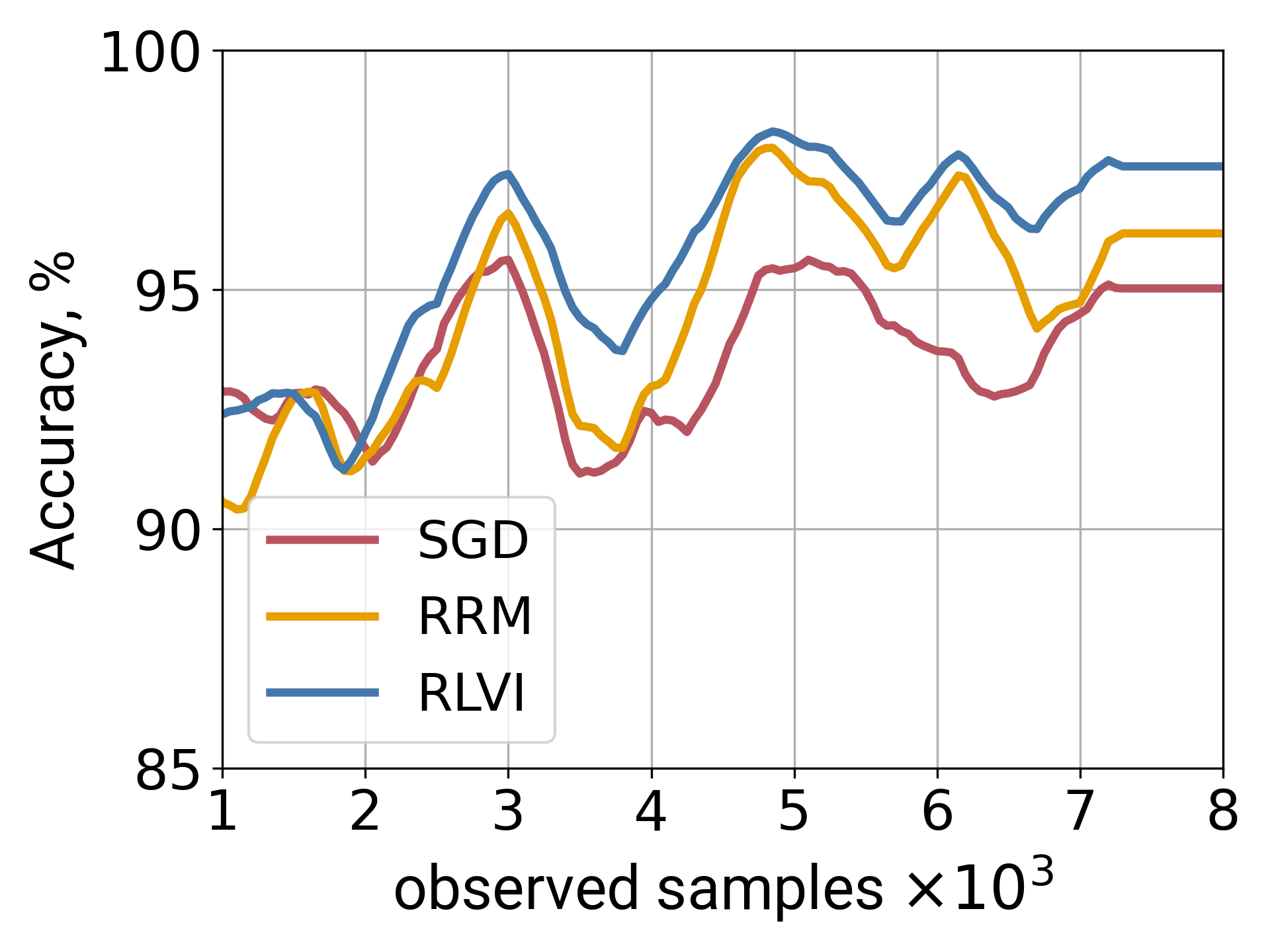}
    \end{minipage}~
    \begin{minipage}[t]{0.35\textwidth}
      \includegraphics[width=\linewidth]{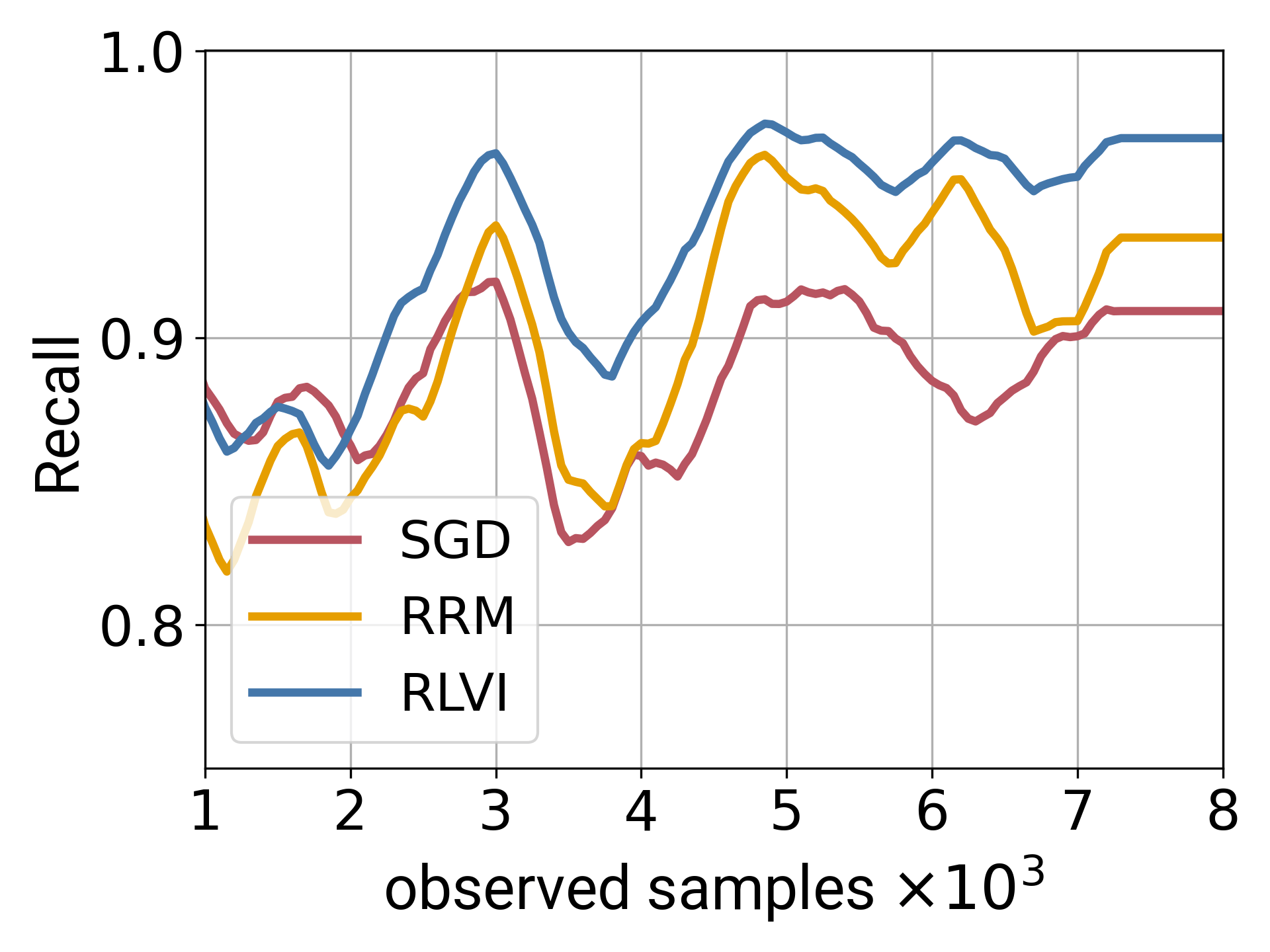}
    \end{minipage}\hfill
  \caption{Online classification.
    Accuracy and recall when learning with the batch size 50.
    }
    \label{fig:online-results-bs50}
\end{figure}

To obtain the main results in online classification, shown in \cref{fig:online-results}, we used batches of 100 observations during optimization (and reserved 100 samples for computing accuracy and recall in each iteration). To demonstrate how batch size affects the performance, in \cref{fig:online-results-bs75,fig:online-results-bs50} we additionally show the same metrics when 75 and 50 examples are used for training respectively (but 100 samples are still used for testing).


\begin{table}[ht]
    \caption{
        Hyperparameter settings for the deep learning experiments (LR = learning rate, mom. = momentum)
    }
    \resizebox{\textwidth}{!}{
    \begin{tabular}{lcccc}
    \toprule
        Dataset & MNIST & CIFAR10 & CIFAR100 \& CIFAR80N-O & Food101 \\
        \cmidrule{1-5}
        Model & LeNet & ResNet18 & ResNet34 & ResNet50 (pre-trained on Imagenet) \\
        Optimizer & SGD with mom. 0.9 & SGD with mom. 0.9 & SGD with mom. 0.9 & Adam \\
        Epochs & 100 & 200 & 200 & 80 \\
        Batch size & 32 & 128 & 128 & 32 \\
        Weight decay & $10^{-3}$ & $5 \cdot 10^{-4}$ & $5 \cdot 10^{-4}$ & $10^{-4}$ \\
        LR schedule & linear decay & cosine annealing & cosine annealing & multi-step \\
        Initial LR & $10^{-2}$ & $10^{-2}$ & $10^{-2}$ & $10^{-3}$ \\
        \bottomrule
    \end{tabular}
    }
    \label{tab:dl-configs}
\end{table}

\section*{C. Overparameterized setting: image classification using convolutional neural networks}
\subsection*{C1. Hyperparameters}
In the experiments on image classification, we use hyperparameter configurations that follow the common settings found in literature. These settings are listed in \cref{tab:dl-configs}.


\subsection*{C2. Synthetic corruption}
\label{synthetic-noise}
\begin{figure*}[h]
    \centering
    \begin{minipage}[t]{0.25\linewidth}
      \includegraphics[width=\linewidth]{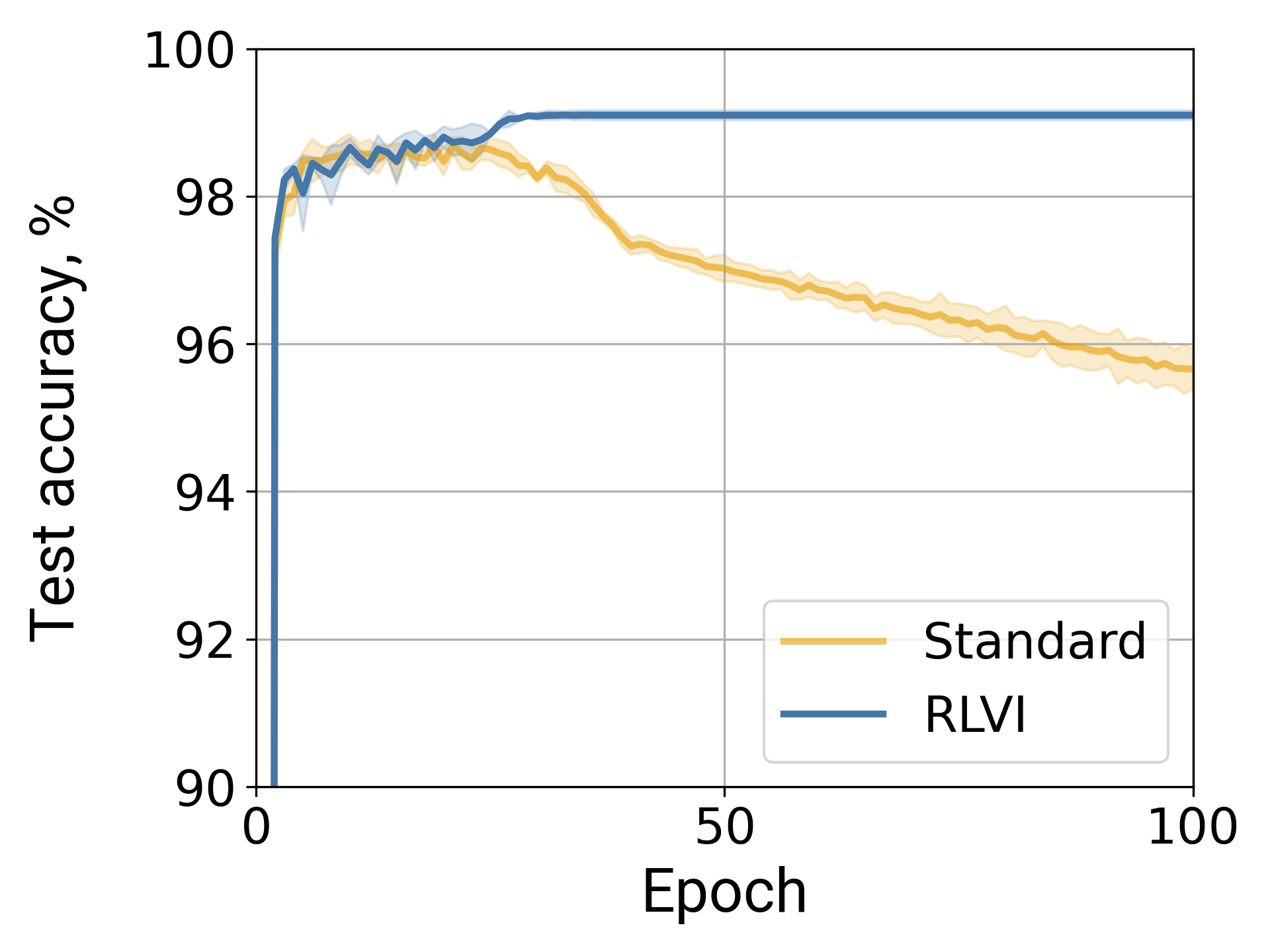}
    \end{minipage}\hfill
    \begin{minipage}[t]{0.25\linewidth}
      \includegraphics[width=\linewidth]{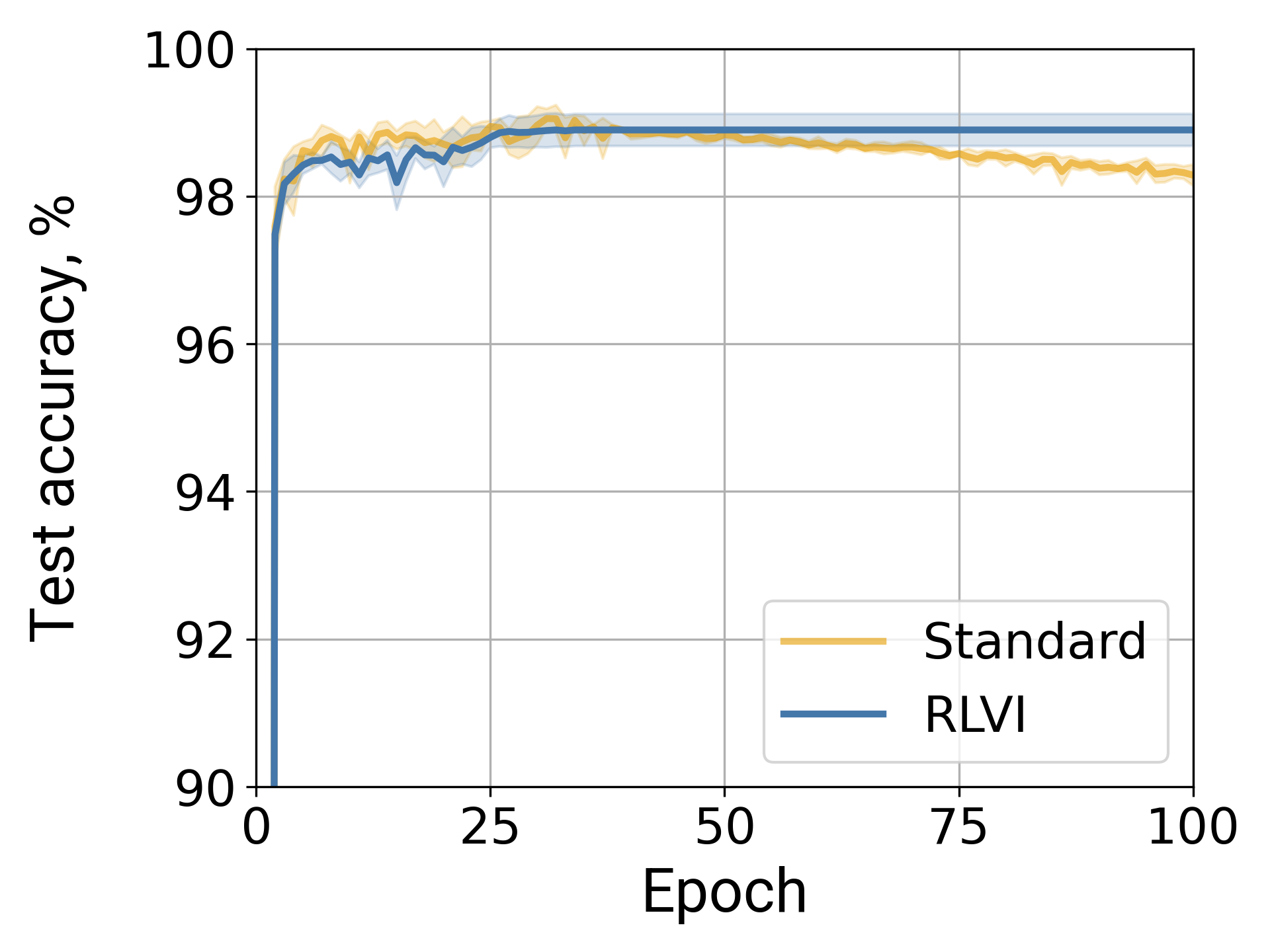}
    \end{minipage}\hfill
    \begin{minipage}[t]{0.25\linewidth}
      \includegraphics[width=\linewidth]{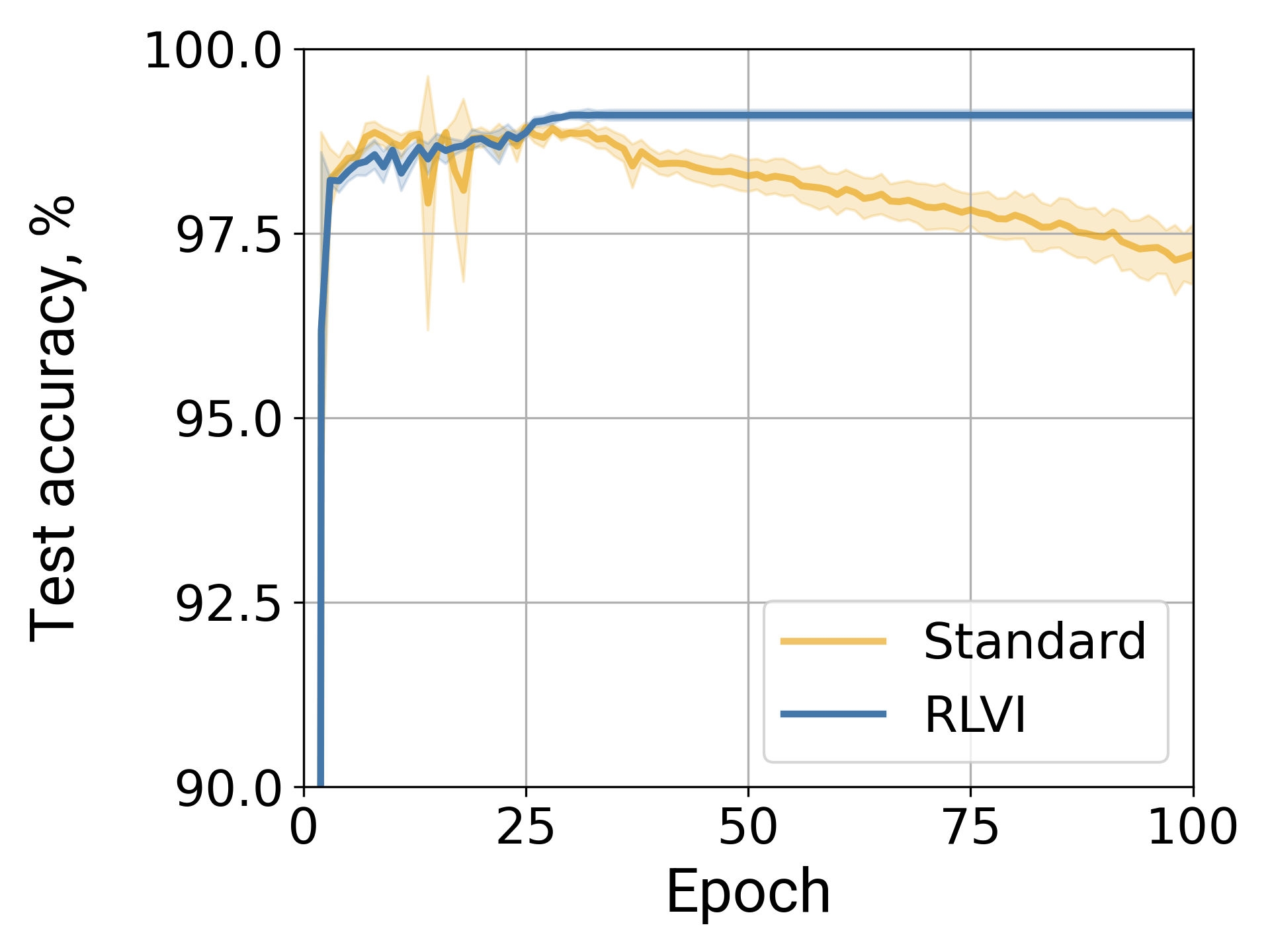}
    \end{minipage}\hfill
    \begin{minipage}[t]{0.25\linewidth}
      \includegraphics[width=\linewidth]{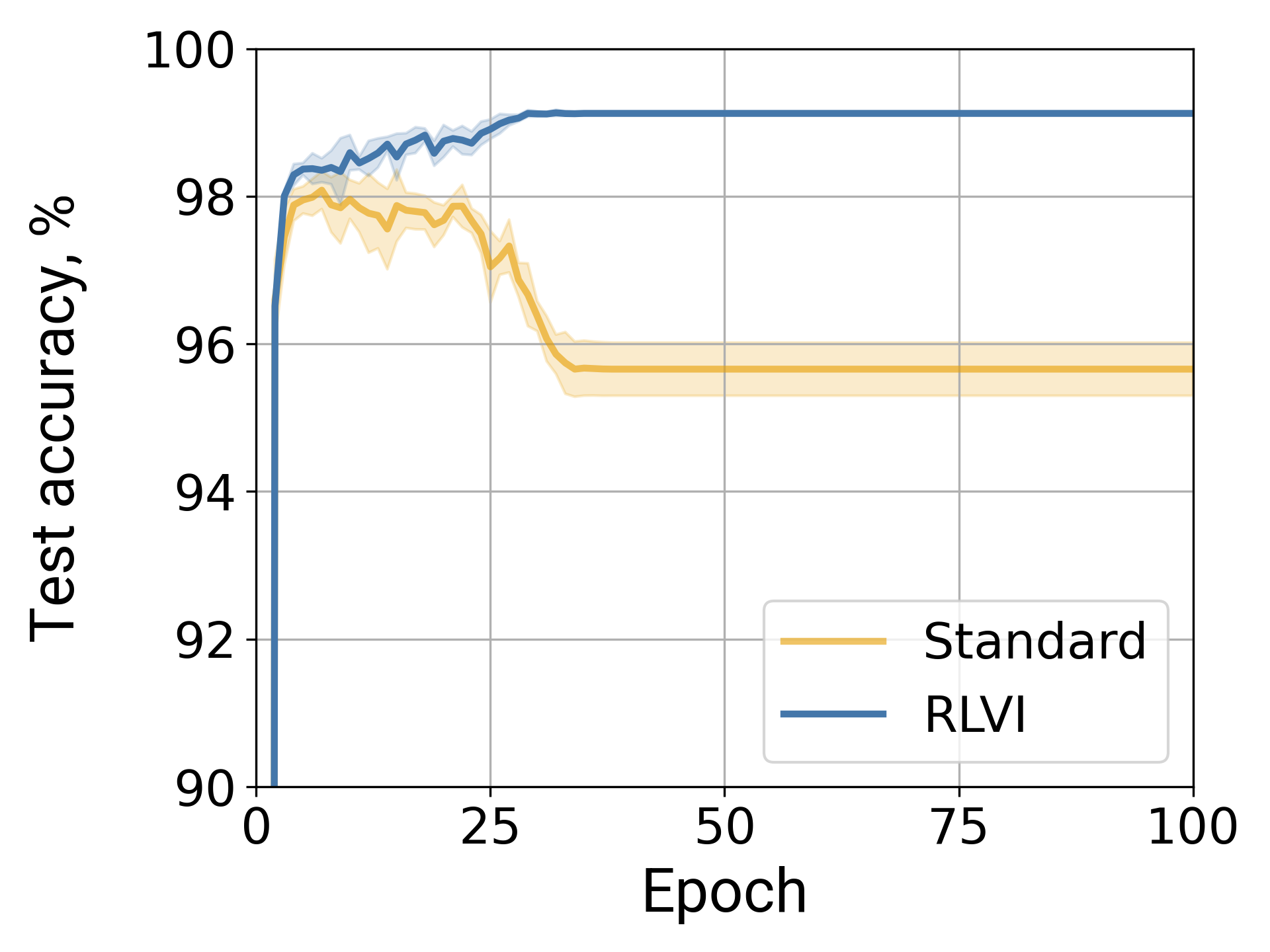}
    \end{minipage}\hfill
    \vfill
    \begin{minipage}[t]{0.25\linewidth}
      \includegraphics[width=\linewidth]{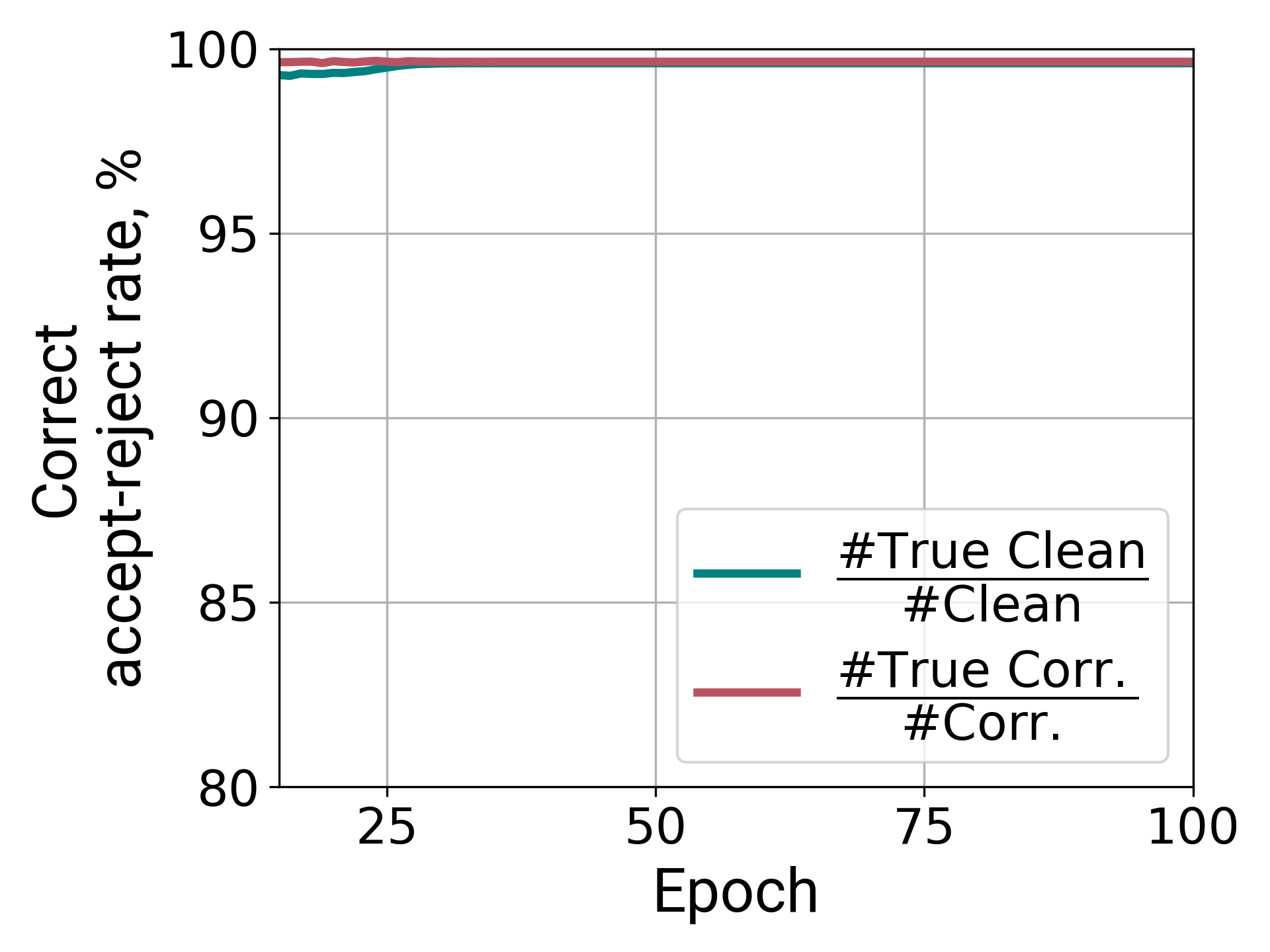}
      \captionsetup{justification=centering}
      \caption*{Symmetric}
    \end{minipage}\hfill
    \begin{minipage}[t]{0.25\linewidth}
      \includegraphics[width=\linewidth]{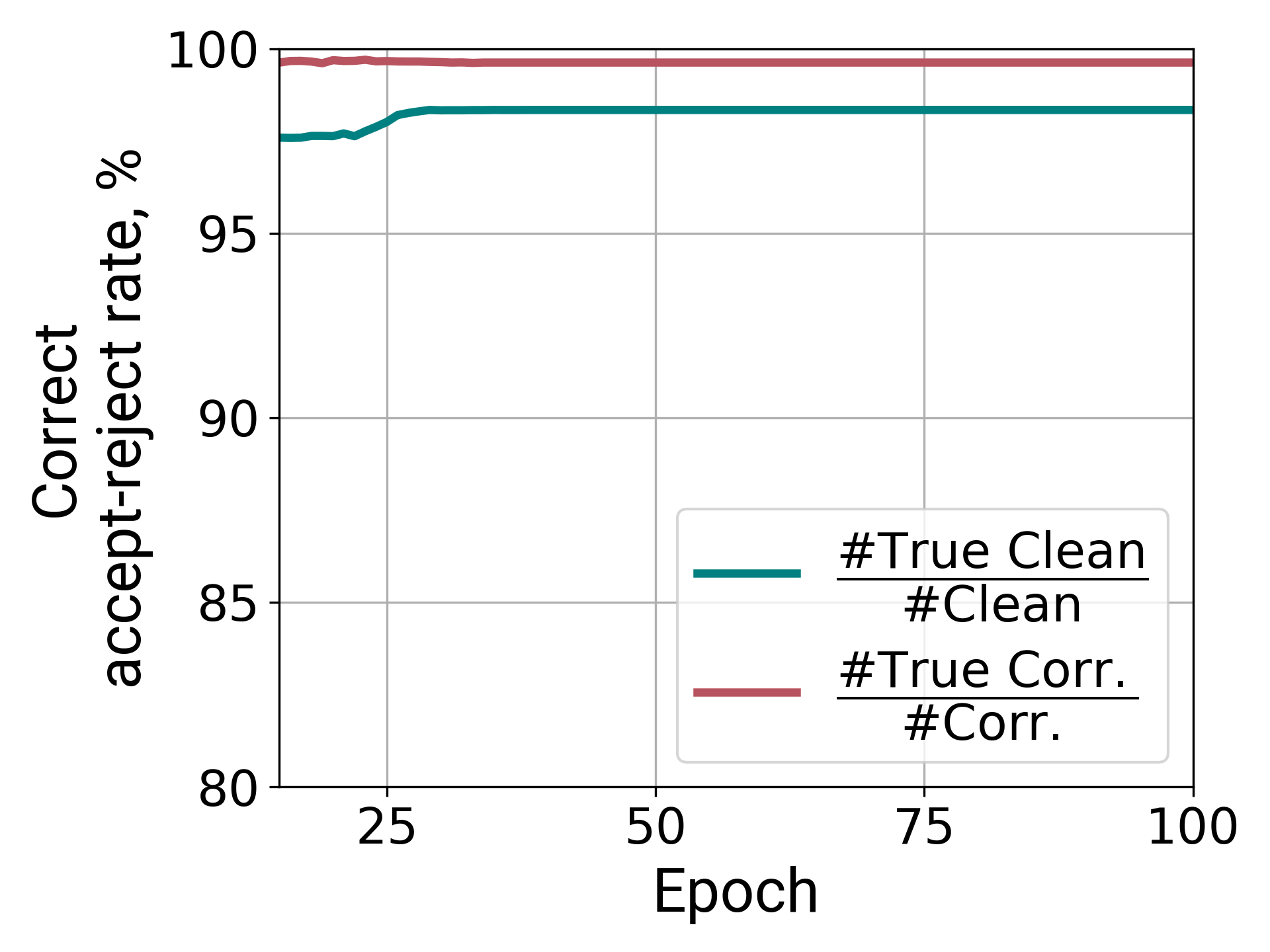}
      \caption*{Asymmetric}
    \end{minipage}\hfill
    \begin{minipage}[t]{0.25\linewidth}
      \includegraphics[width=\linewidth]{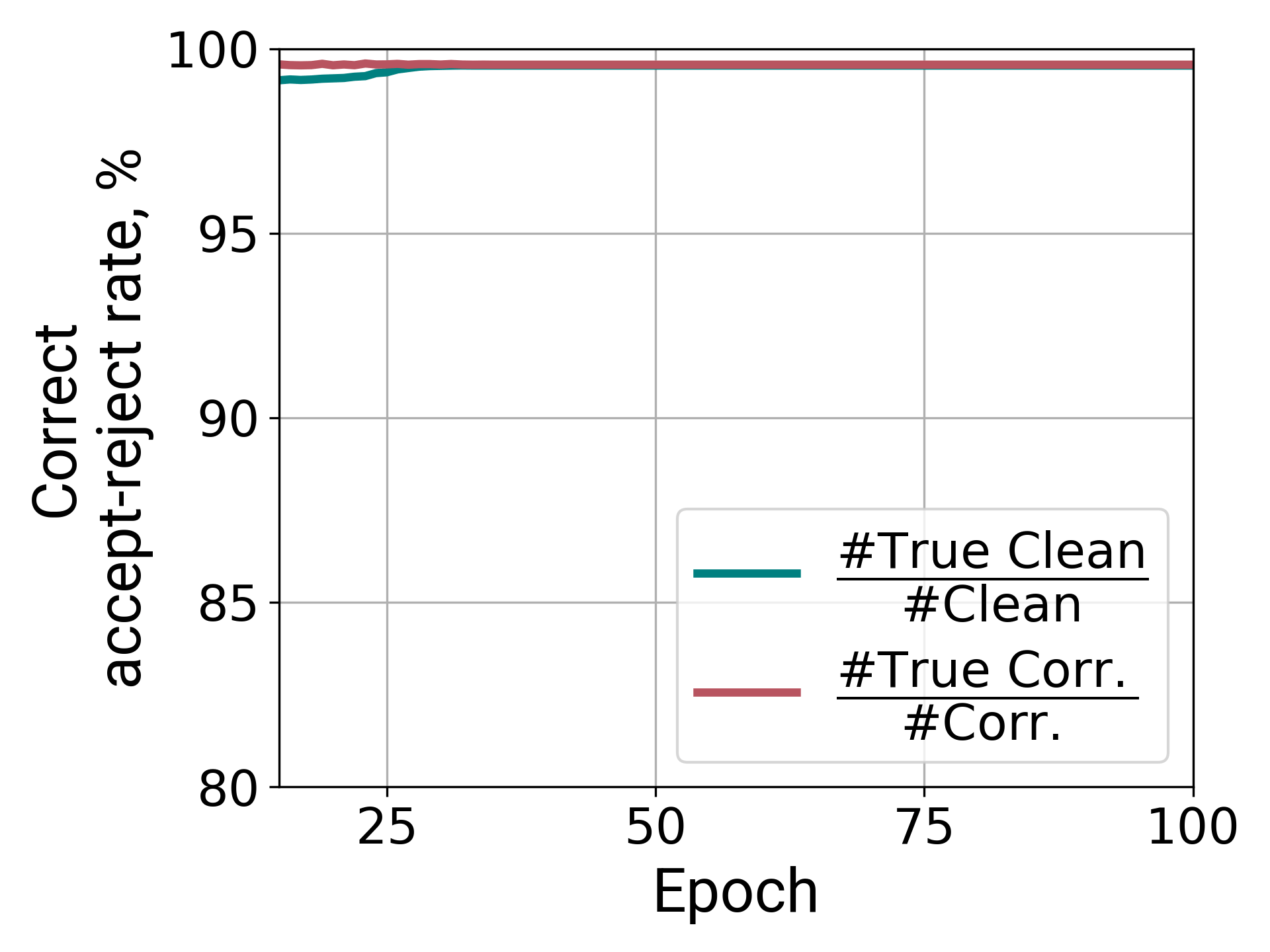}
      \caption*{Pairflip}
    \end{minipage}\hfill
    \begin{minipage}[t]{0.25\linewidth}
      \includegraphics[width=\linewidth]{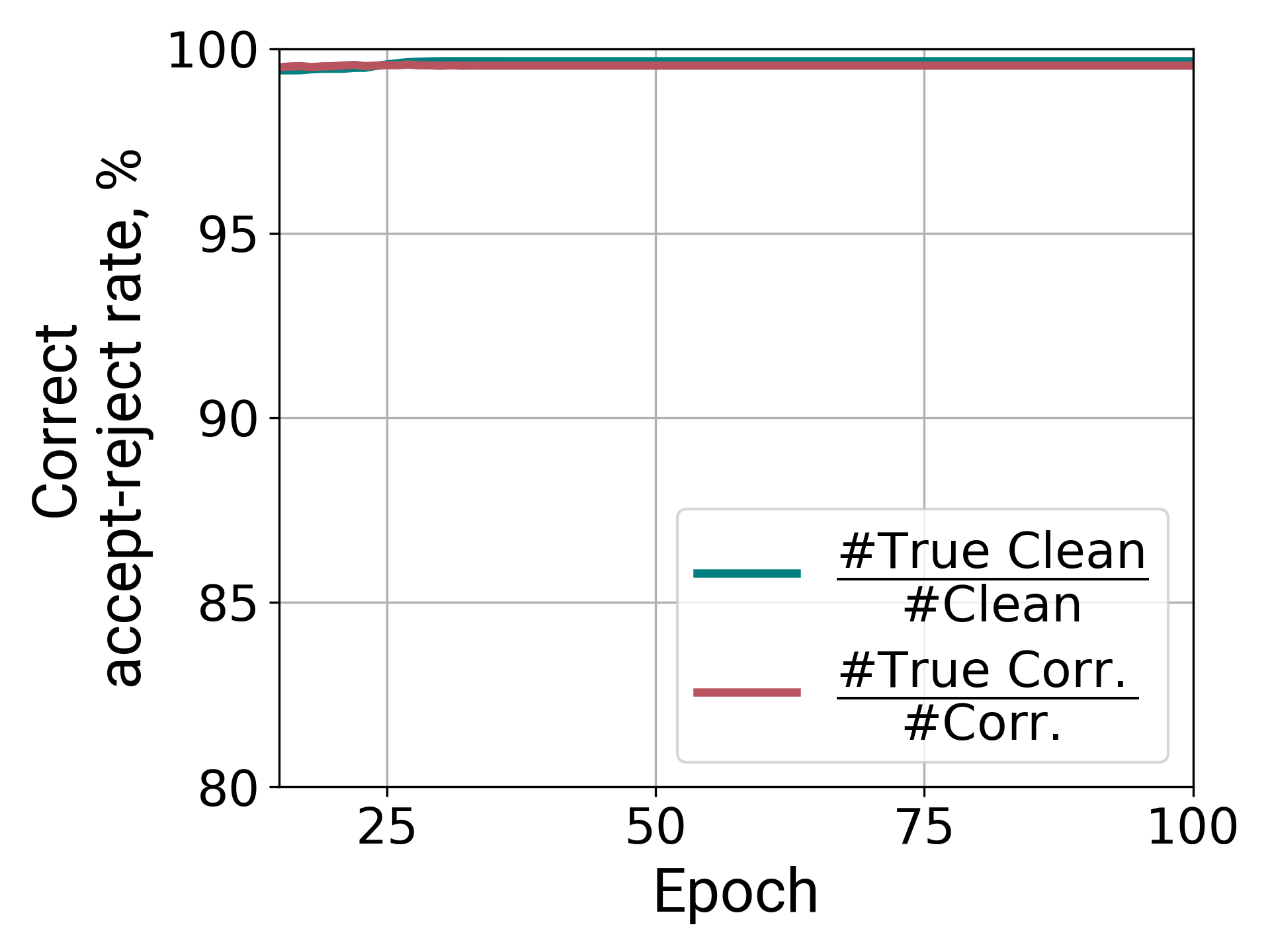}
      \caption*{Instance}
    \end{minipage}\hfill
\caption{
MNIST, 20\% noise rate. \emph{Top}: test accuracy (mean $\pm$ st. dev. over 5 runs). \emph{Bottom}: type I and type II errors (mean over 5 runs).
}
\label{fig:performance-1}
\end{figure*}

\begin{figure*}[h]
    \centering
    \begin{minipage}[t]{0.25\linewidth}
      \includegraphics[width=\linewidth]{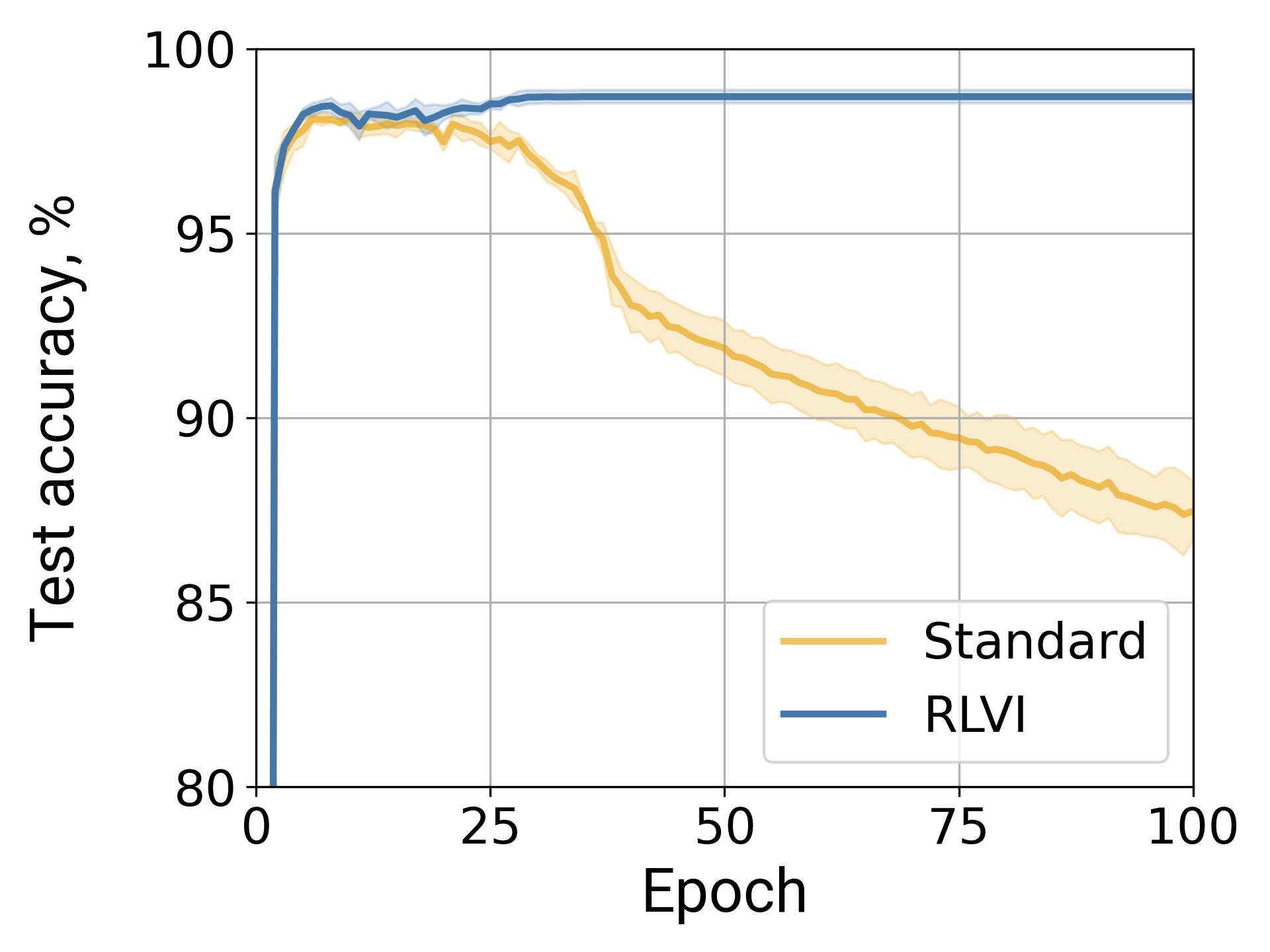}
    \end{minipage}\hfill
    \begin{minipage}[t]{0.25\linewidth}
      \includegraphics[width=\linewidth]{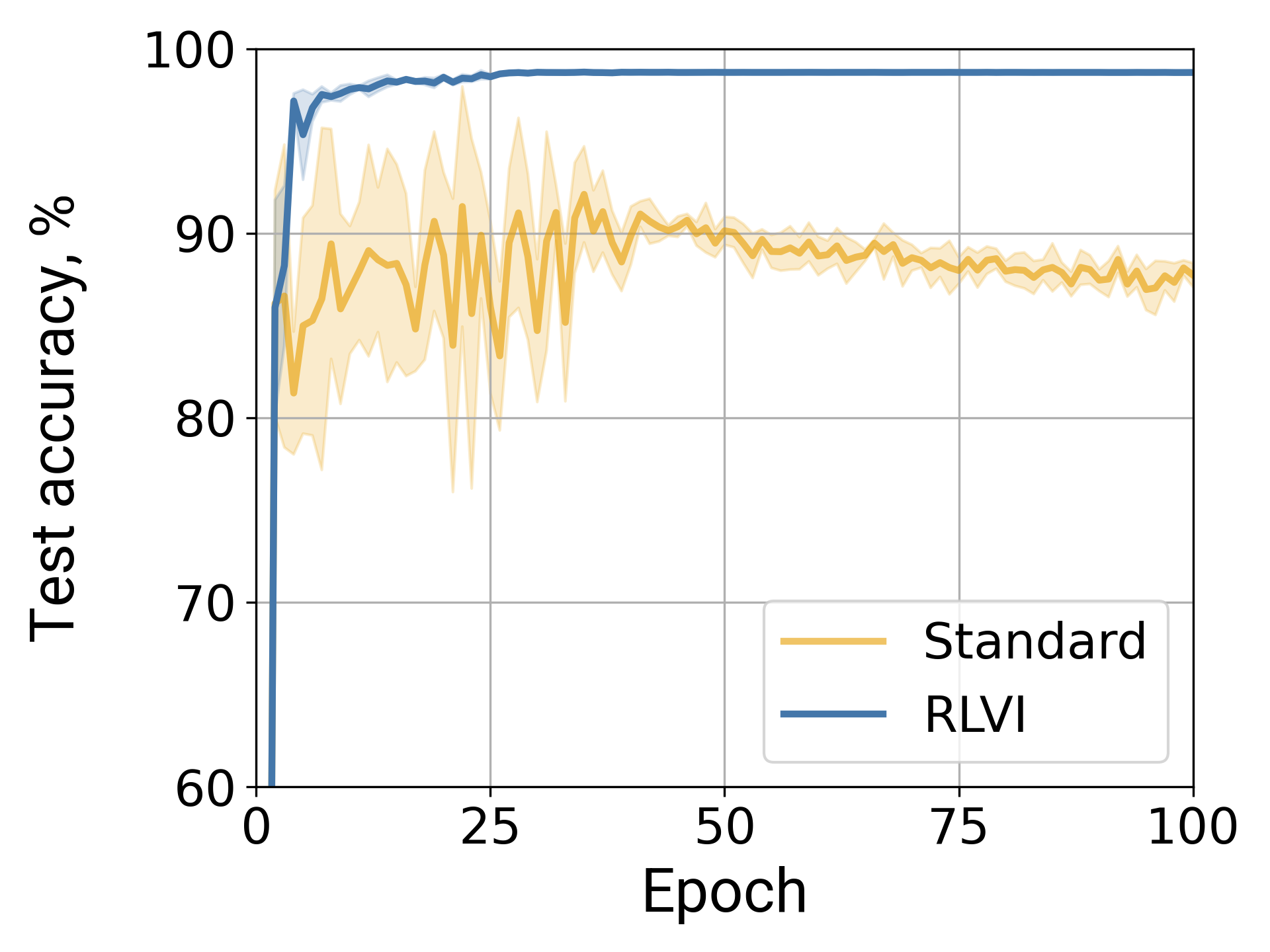}
    \end{minipage}\hfill
    \begin{minipage}[t]{0.25\linewidth}
      \includegraphics[width=\linewidth]{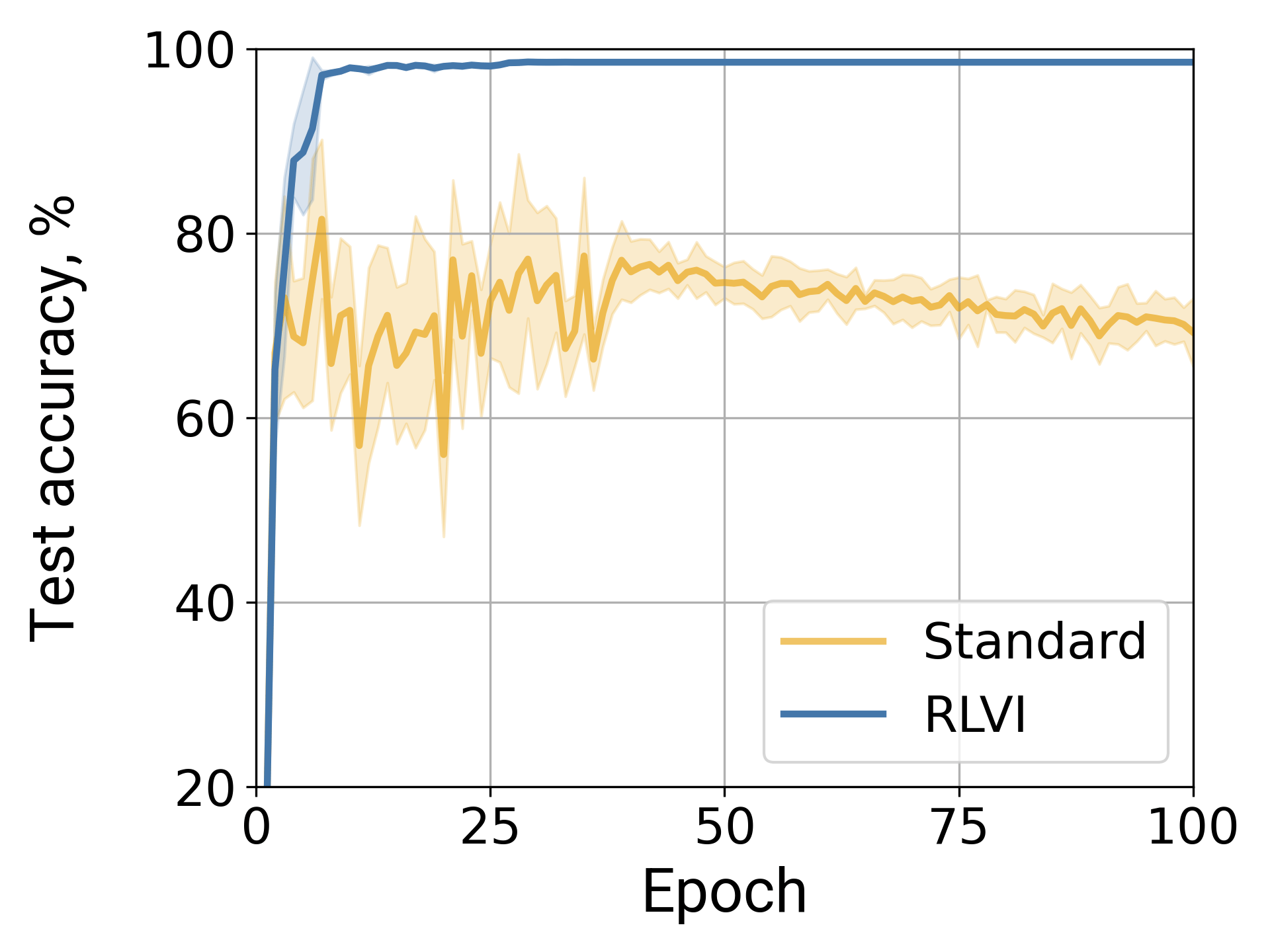}
    \end{minipage}\hfill
    \begin{minipage}[t]{0.25\linewidth}
      \includegraphics[width=\linewidth]{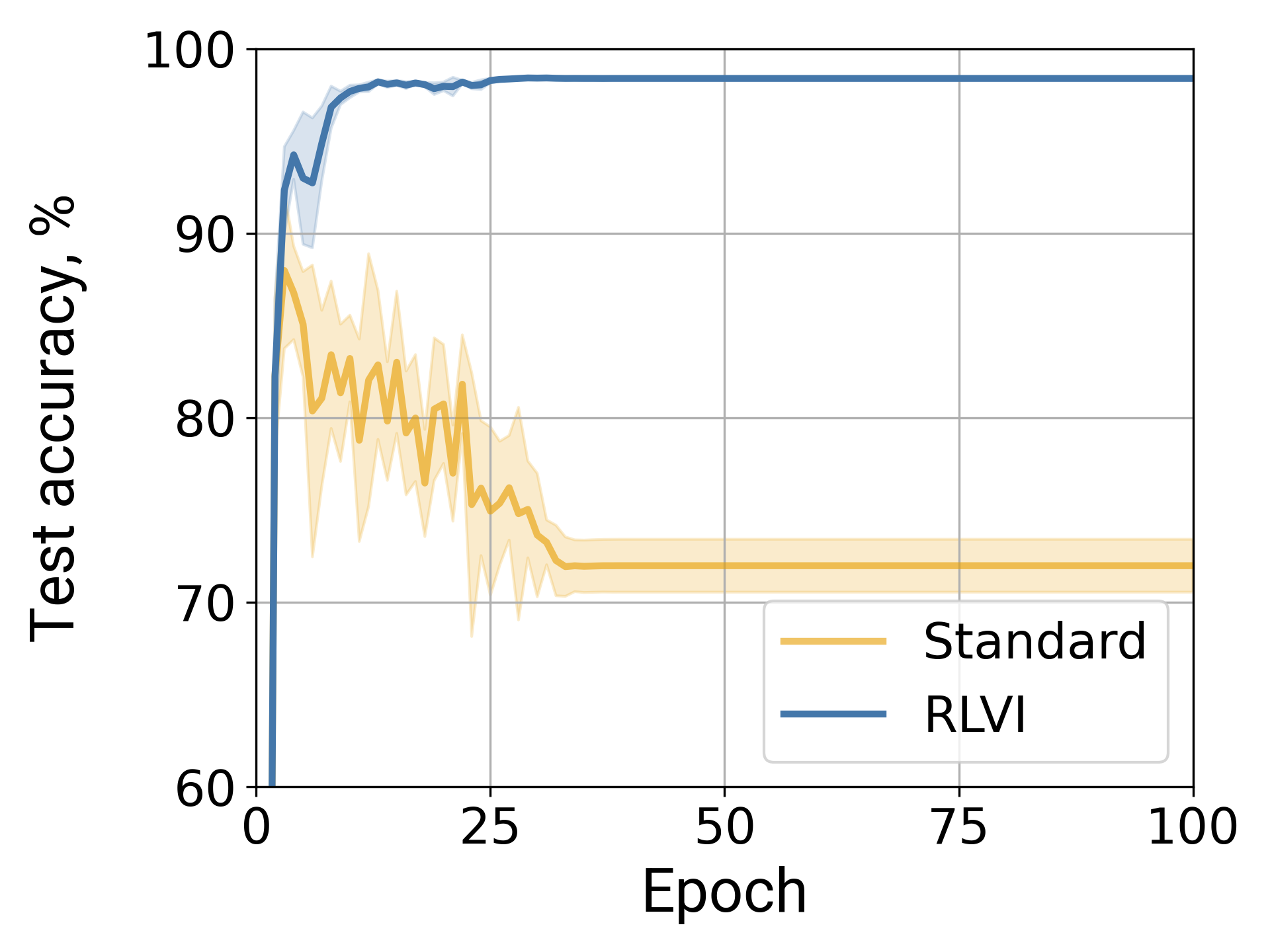}
    \end{minipage}\hfill
    \vfill
    \begin{minipage}[t]{0.25\linewidth}
      \includegraphics[width=\linewidth]{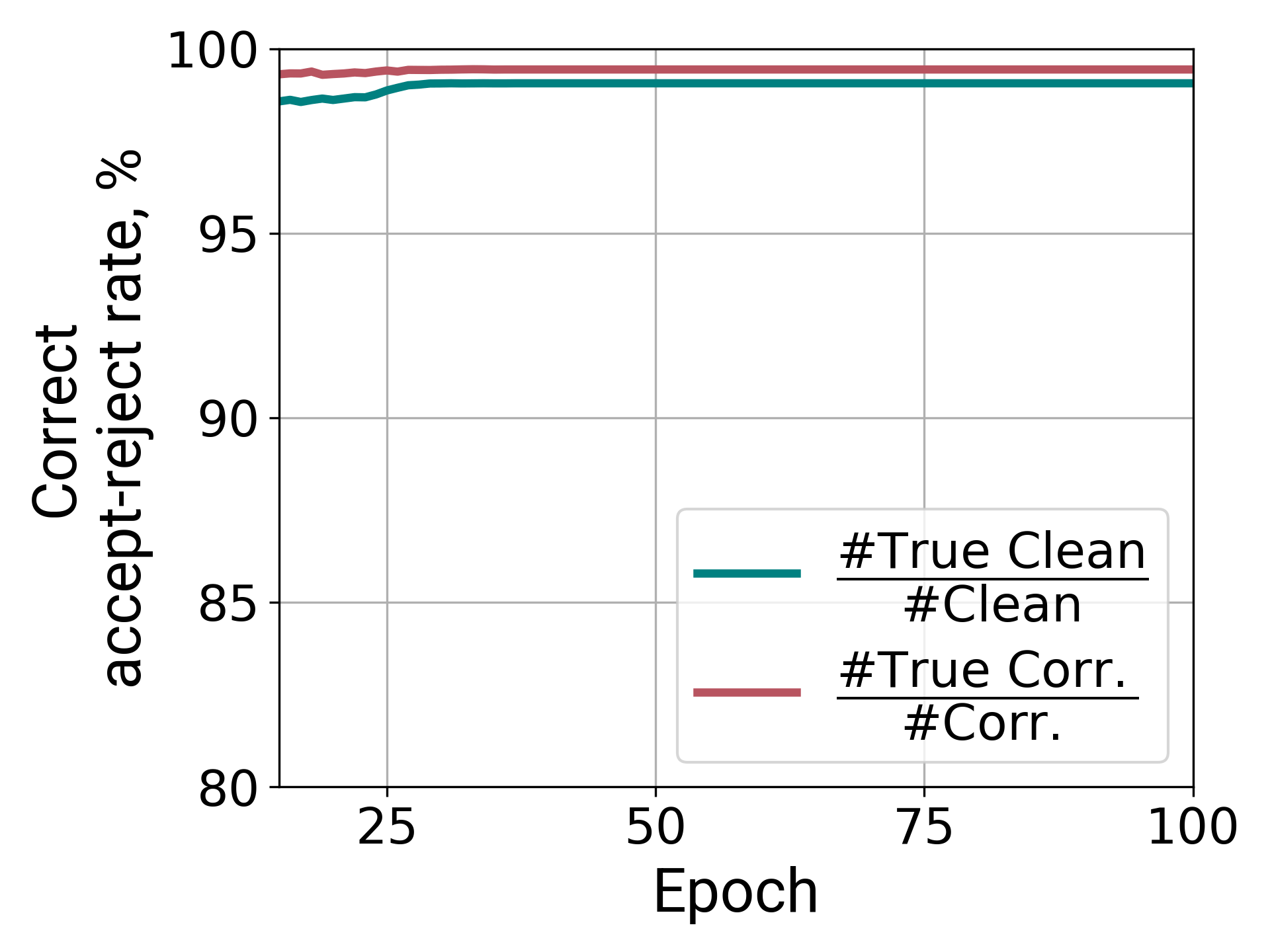}
      \caption*{Symmetric}
    \end{minipage}\hfill
    \begin{minipage}[t]{0.25\linewidth}
      \includegraphics[width=\linewidth]{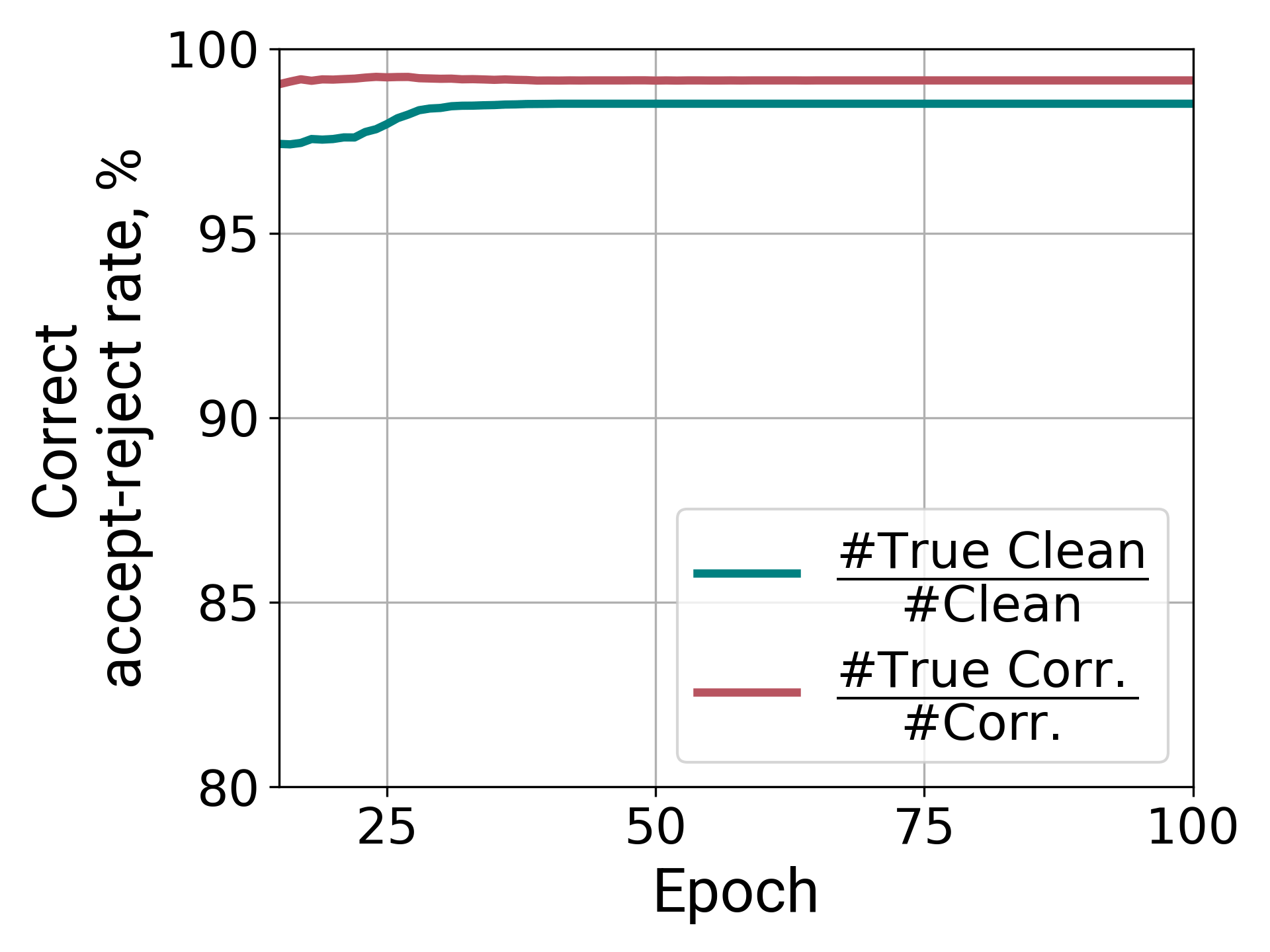}
    \caption*{Asymmetric}
    \end{minipage}\hfill
    \begin{minipage}[t]{0.25\linewidth}
      \includegraphics[width=\linewidth]{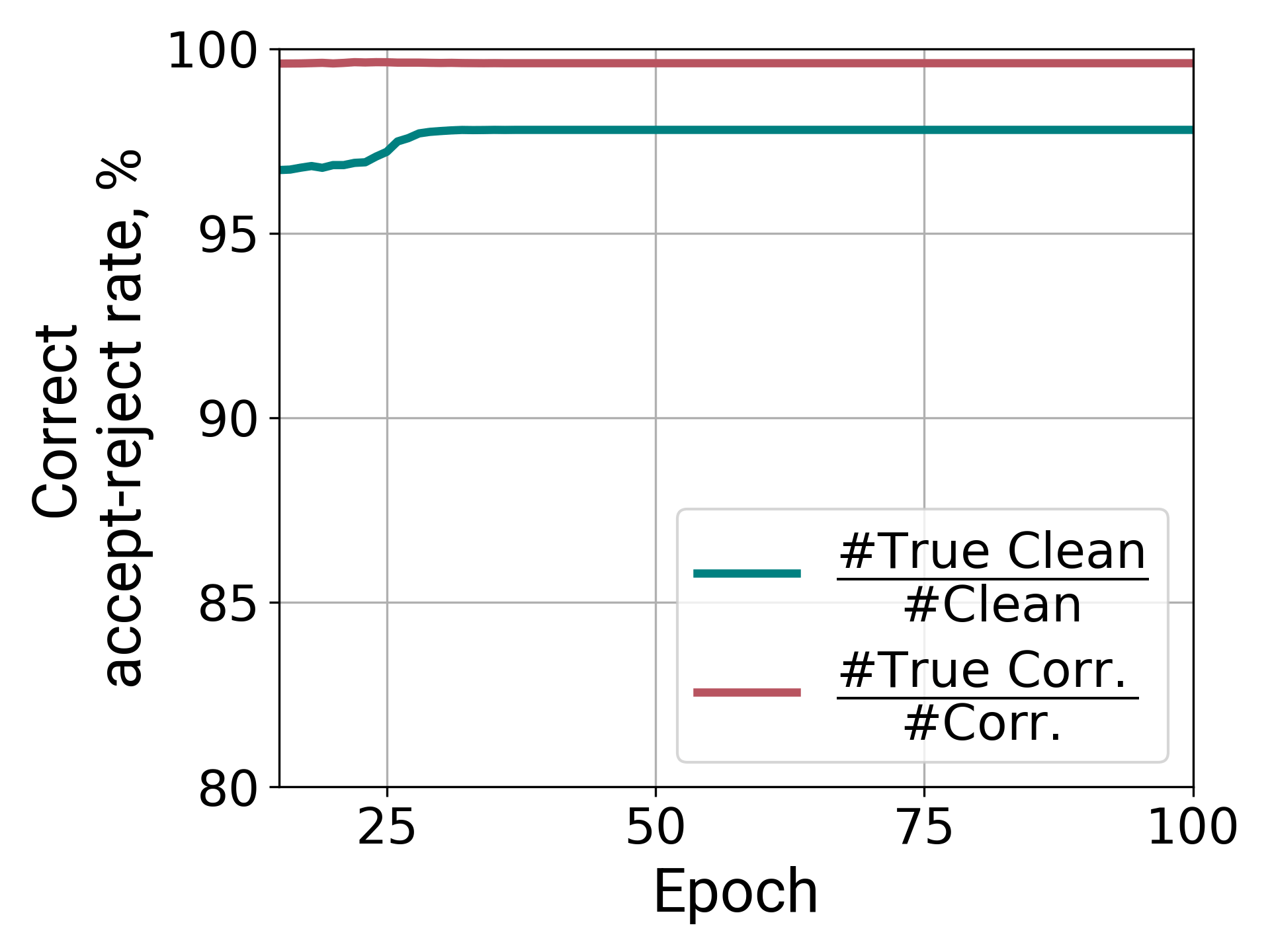}
      \caption*{Pairflip}
    \end{minipage}\hfill
    \begin{minipage}[t]{0.25\linewidth}
      \includegraphics[width=\linewidth]{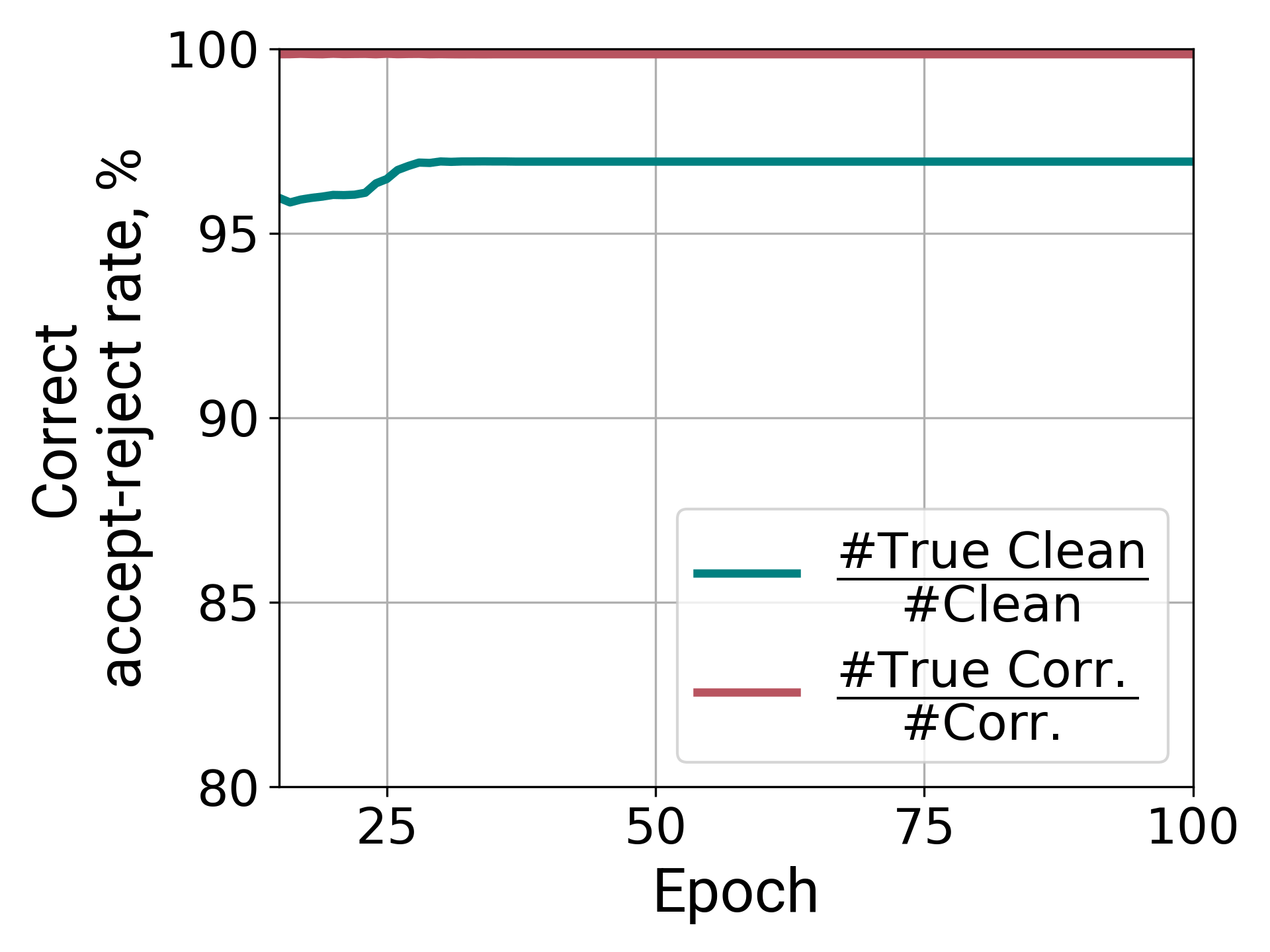}
      \caption*{Instance}
    \end{minipage}\hfill
\caption{
MNIST, 45\% noise rate. \emph{Top}: test accuracy (mean $\pm$ st. dev. over 5 runs). \emph{Bottom}: type I and type II errors (mean over 5 runs).
}
    \label{fig:performance-2}
\end{figure*}

\begin{figure*}[h]
    \centering
    \begin{minipage}[t]{0.25\linewidth}
      \includegraphics[width=\linewidth]{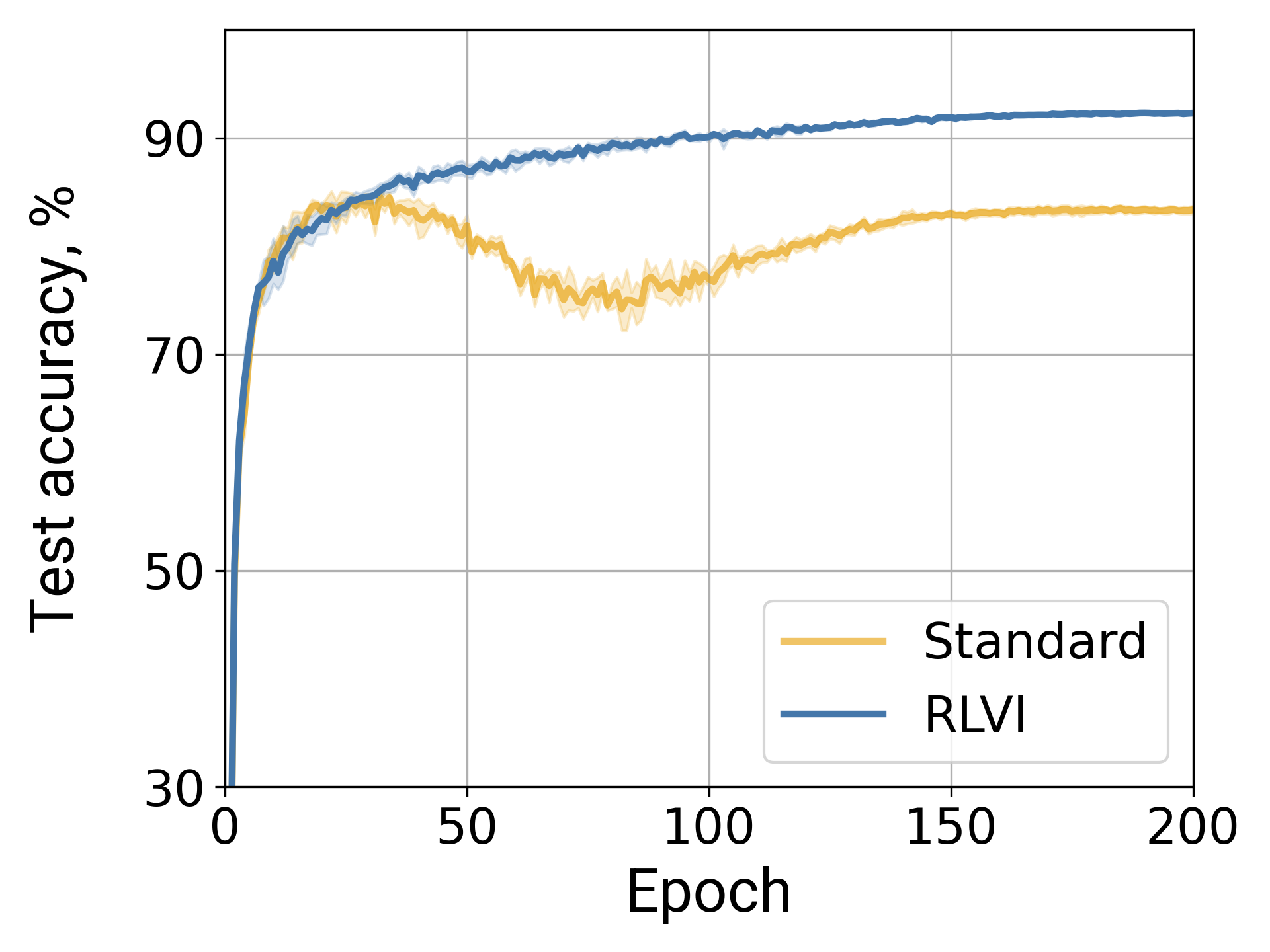}
    \end{minipage}\hfill
    \begin{minipage}[t]{0.25\linewidth}
      \includegraphics[width=\linewidth]{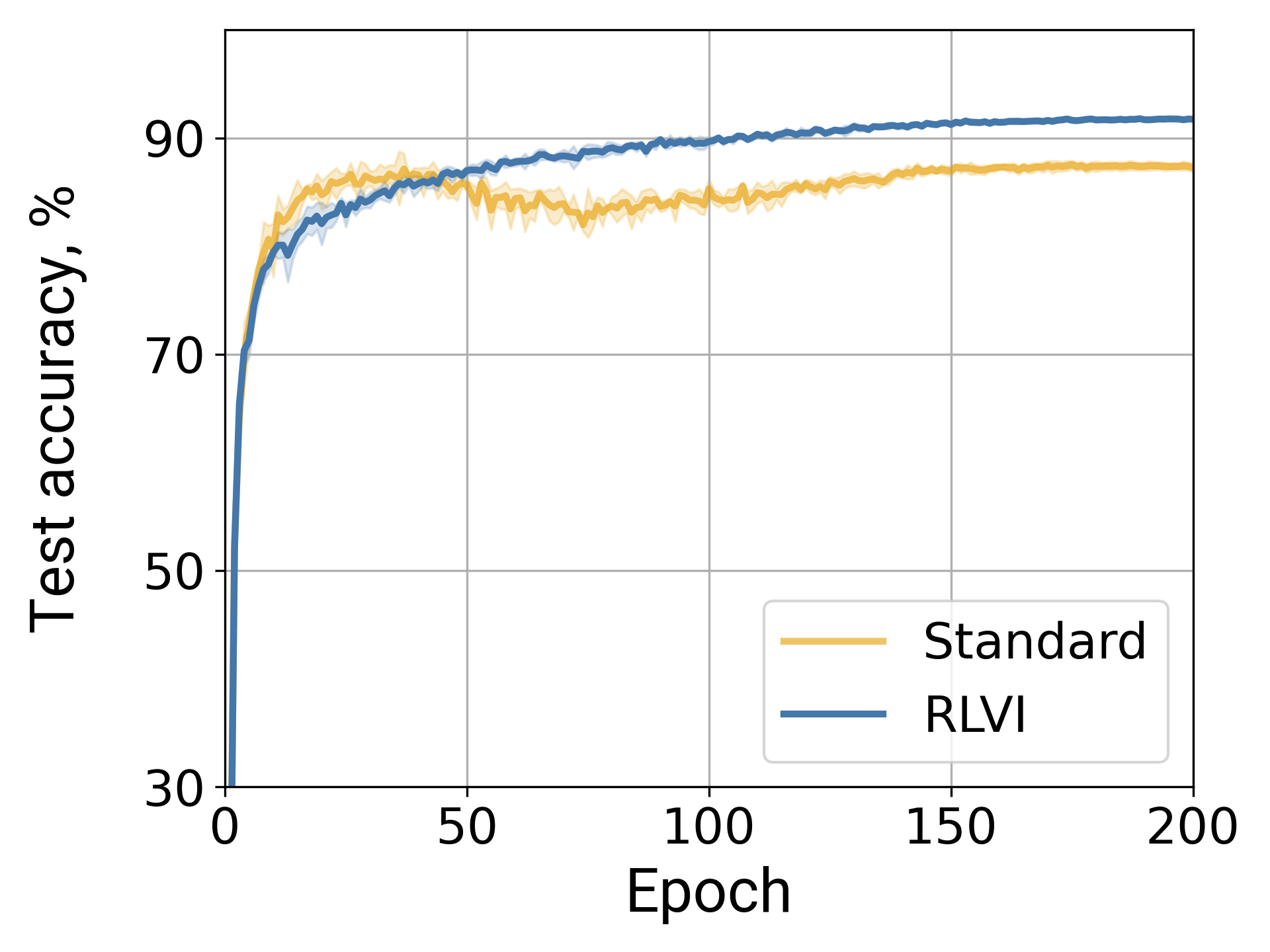}
    \end{minipage}\hfill
    \begin{minipage}[t]{0.25\linewidth}
      \includegraphics[width=\linewidth]{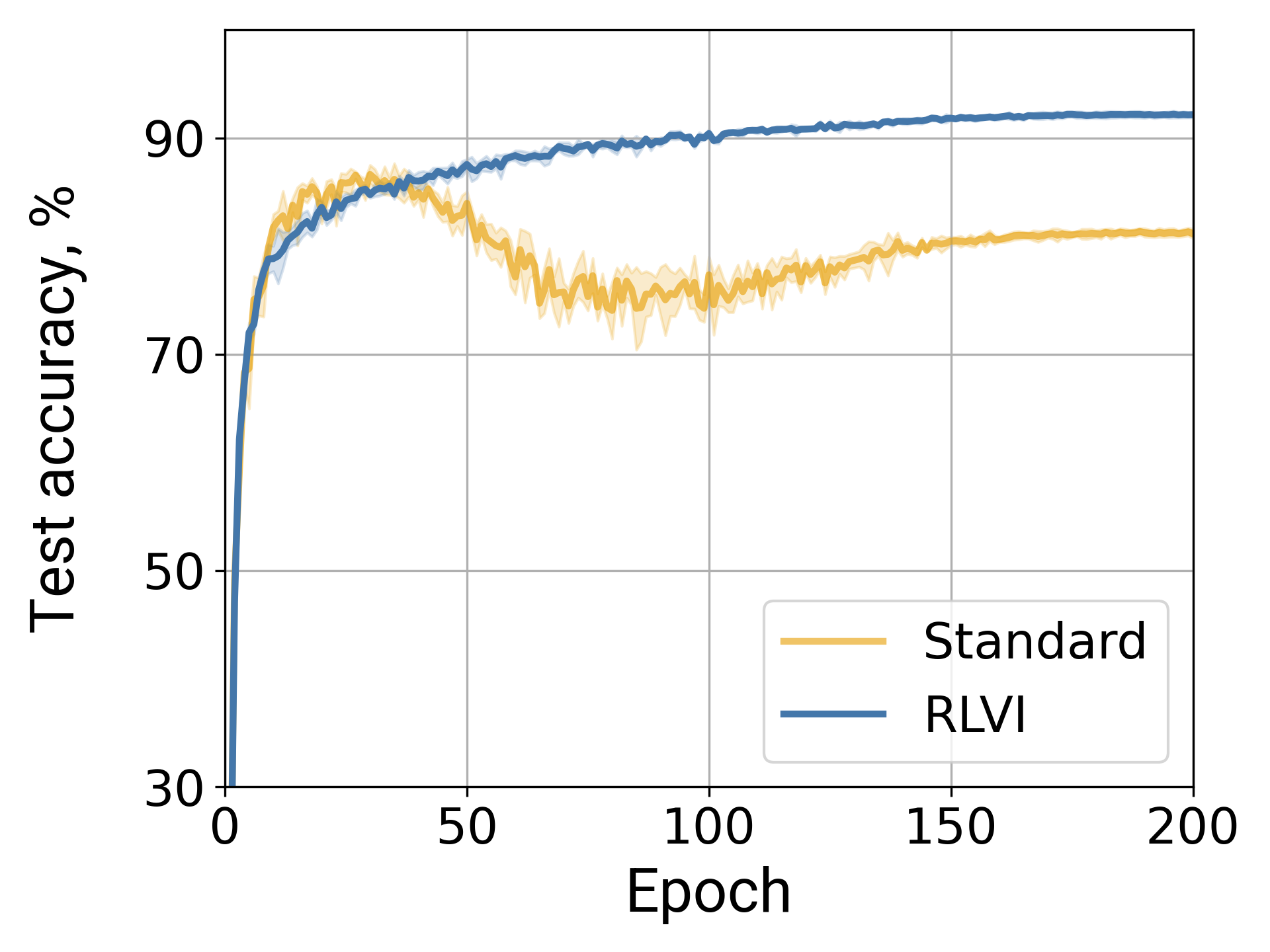}
    \end{minipage}\hfill
    \begin{minipage}[t]{0.25\linewidth}
      \includegraphics[width=\linewidth]{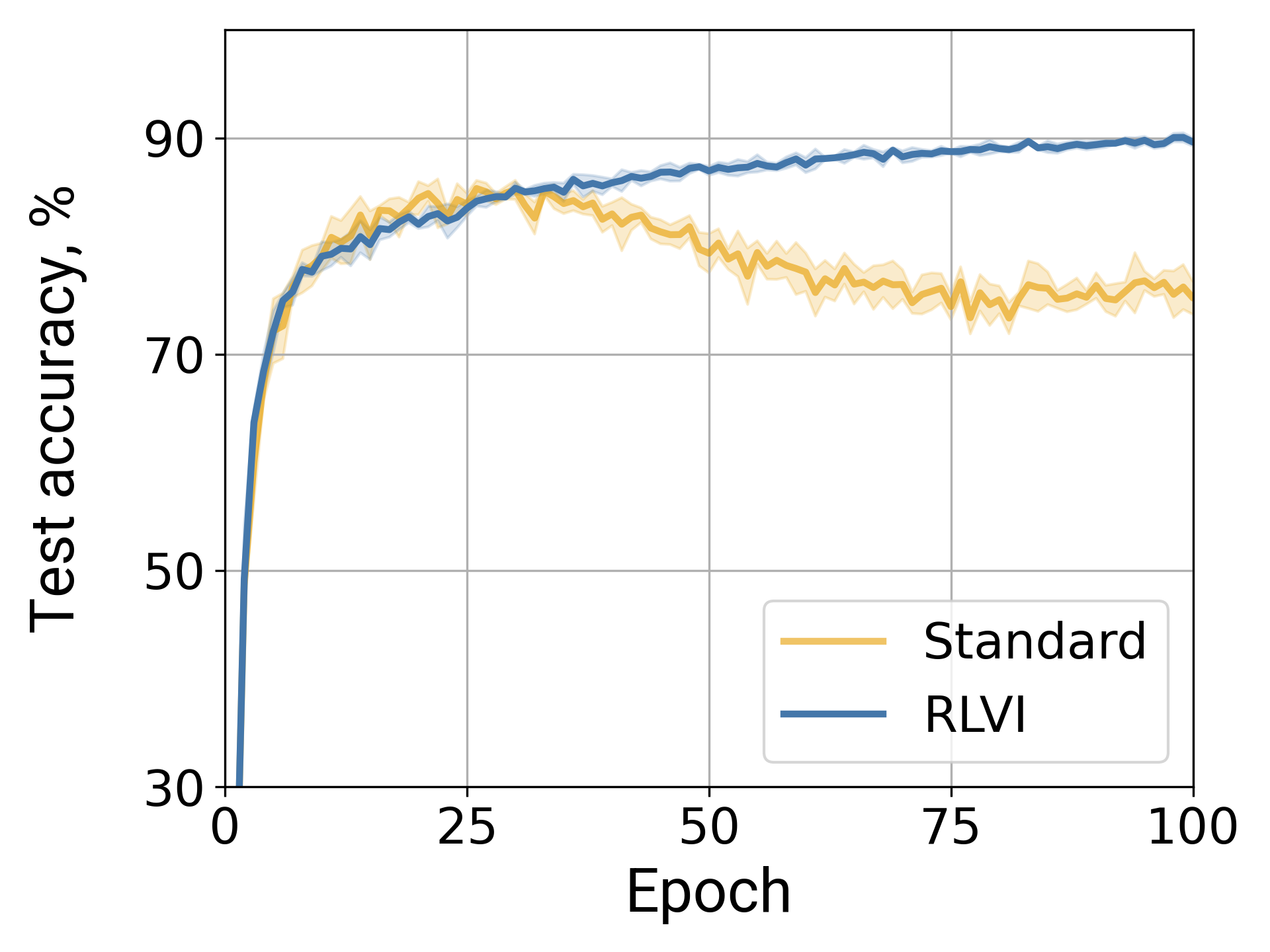}
    \end{minipage}\hfill
    \vfill
    \begin{minipage}[t]{0.25\linewidth}
      \includegraphics[width=\linewidth]{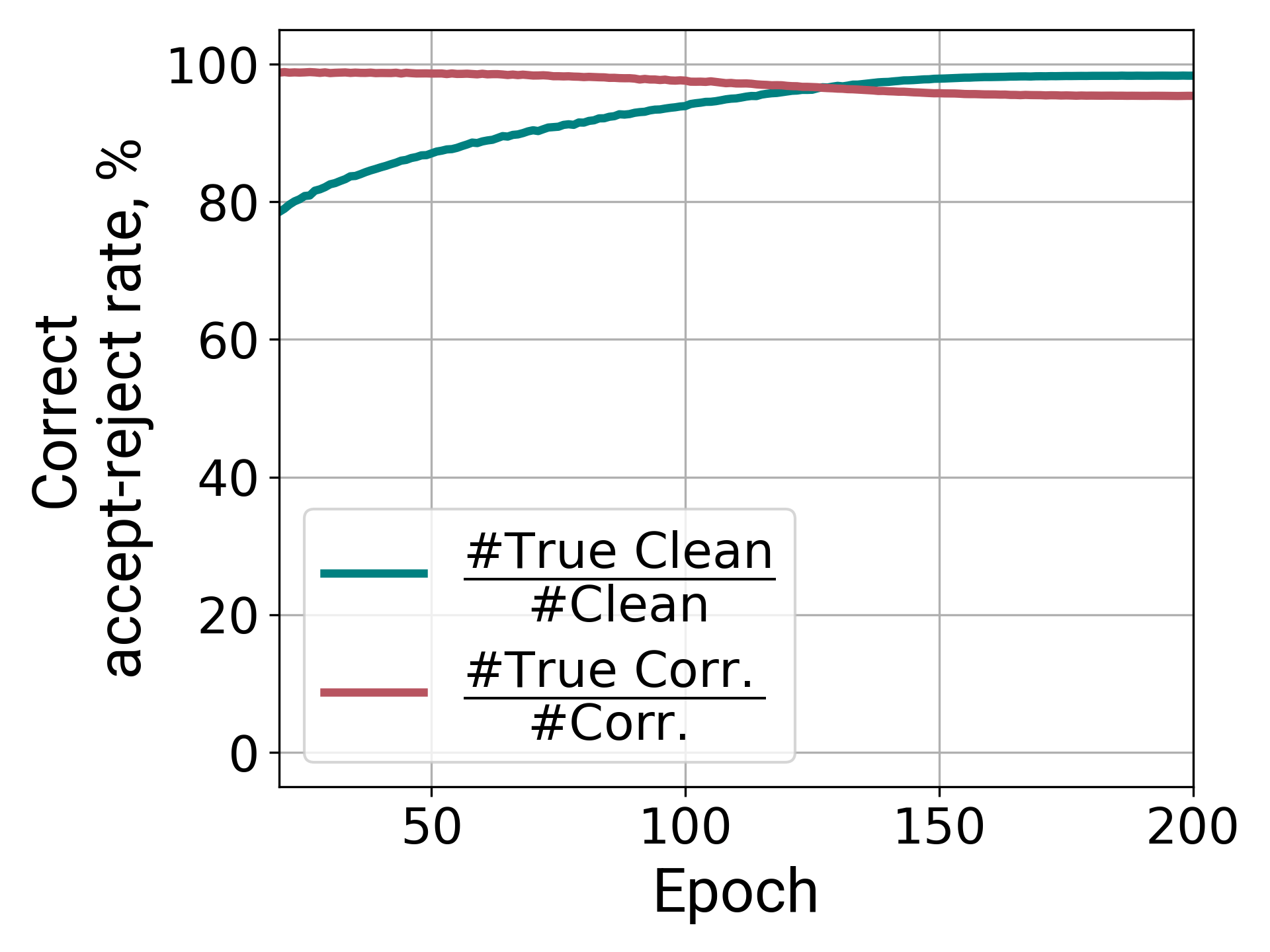}
      \caption*{Symmetric}
    \end{minipage}\hfill
    \begin{minipage}[t]{0.25\linewidth}
      \includegraphics[width=\linewidth]{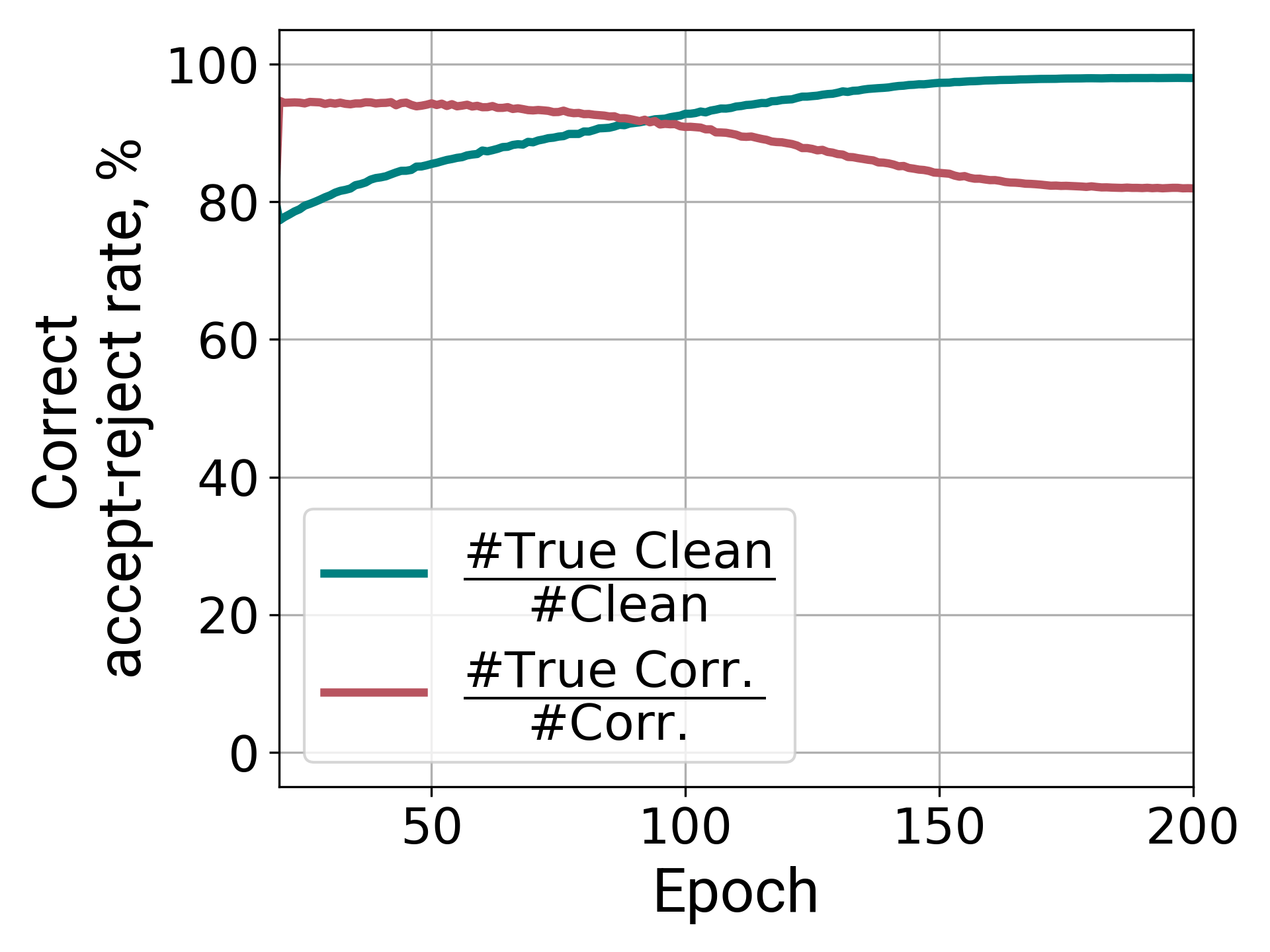}
      \caption*{Asymmetric}
    \end{minipage}\hfill
    \begin{minipage}[t]{0.25\linewidth}
      \includegraphics[width=\linewidth]{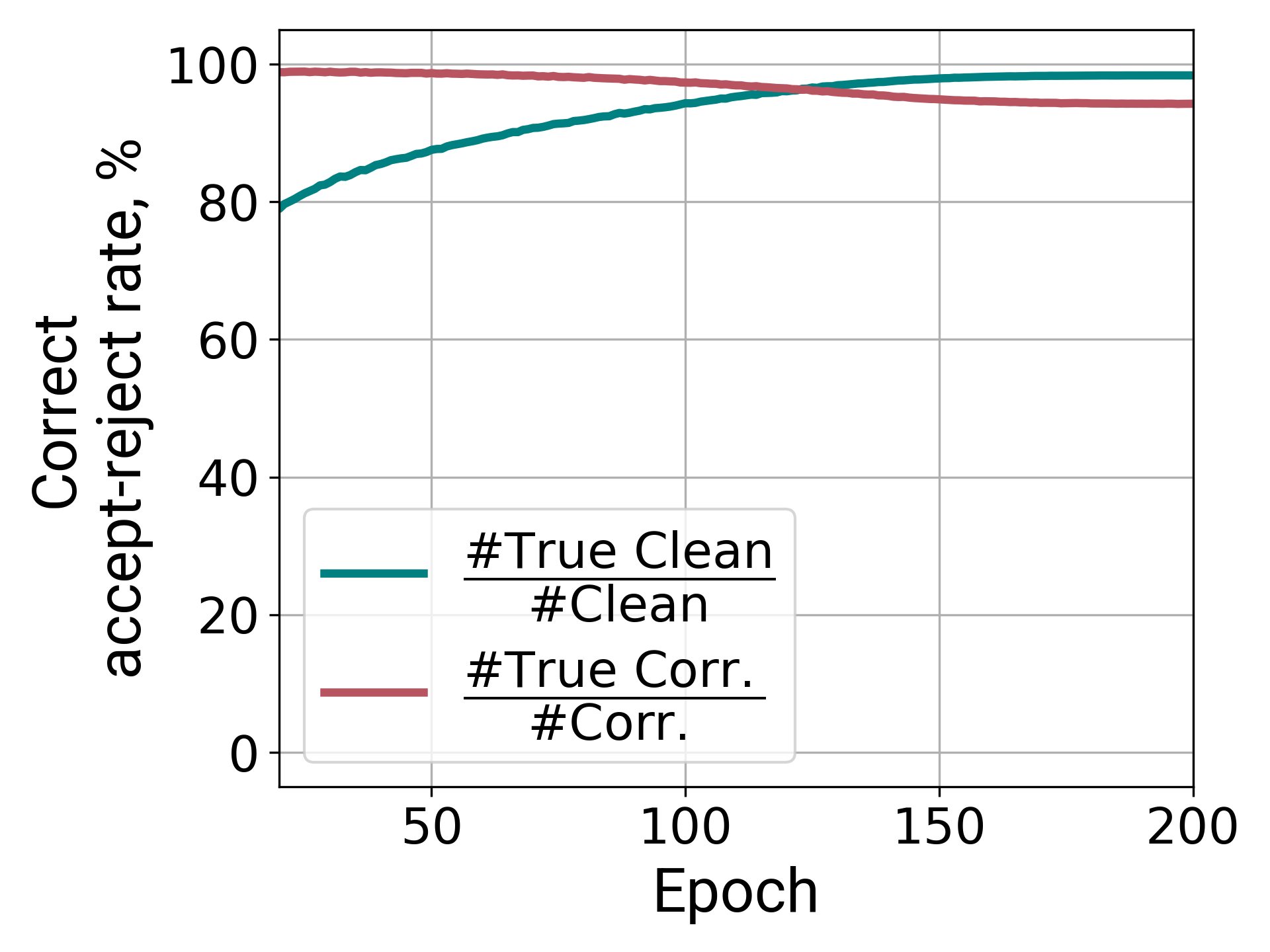}
      \caption*{Pairflip}
    \end{minipage}\hfill
    \begin{minipage}[t]{0.25\linewidth}
      \includegraphics[width=\linewidth]{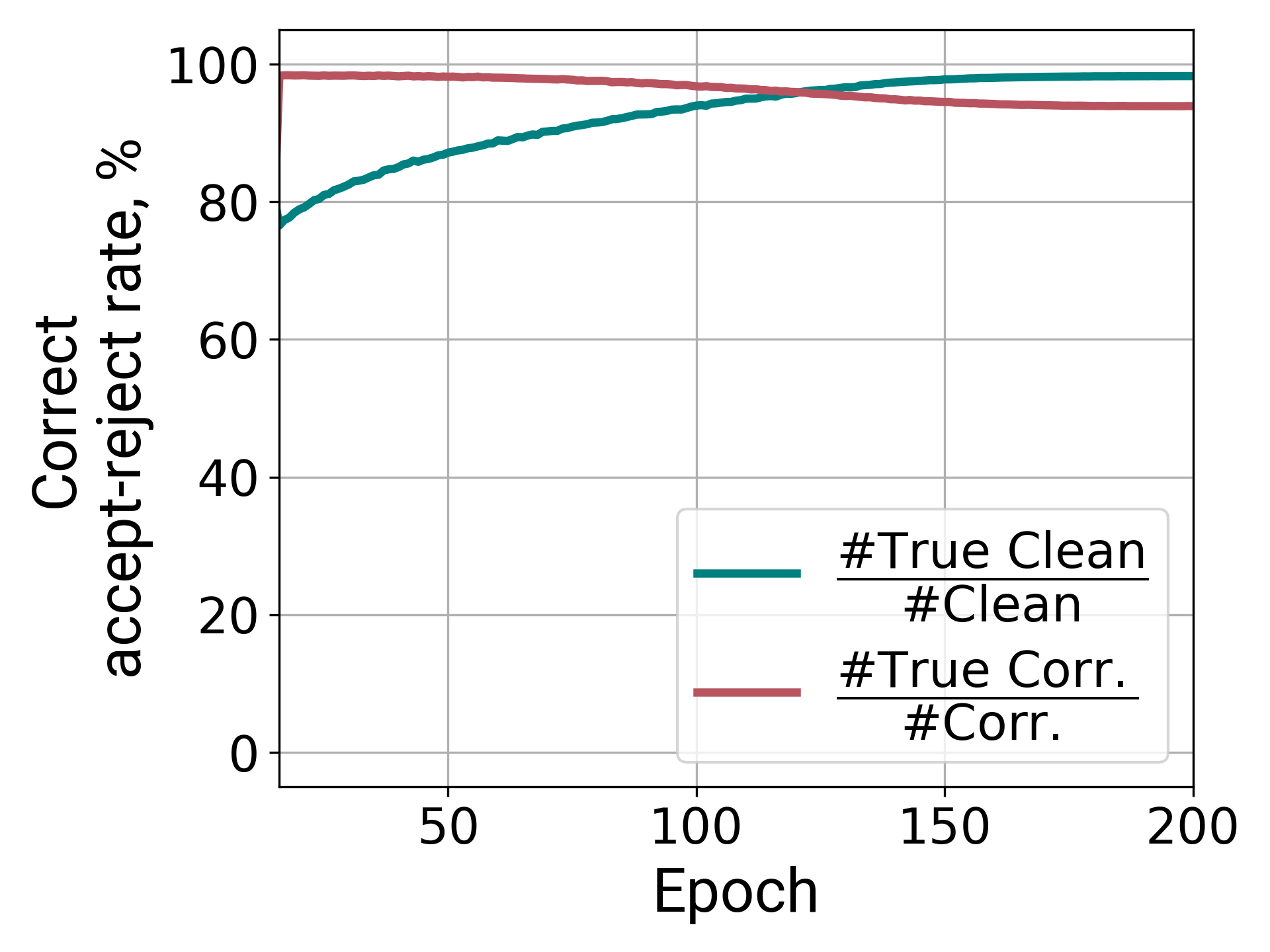}
      \caption*{Instance}
    \end{minipage}\hfill
\caption{
CIFAR10, 20\% noise rate. \emph{Top}: test accuracy (mean $\pm$ st. dev. over 5 runs). \emph{Bottom}: type I and type II errors (mean over 5 runs).
}
\label{fig:performance-3}
\end{figure*}

\begin{figure*}[h]
    \centering
    \begin{minipage}[t]{0.25\linewidth}
      \includegraphics[width=\linewidth]{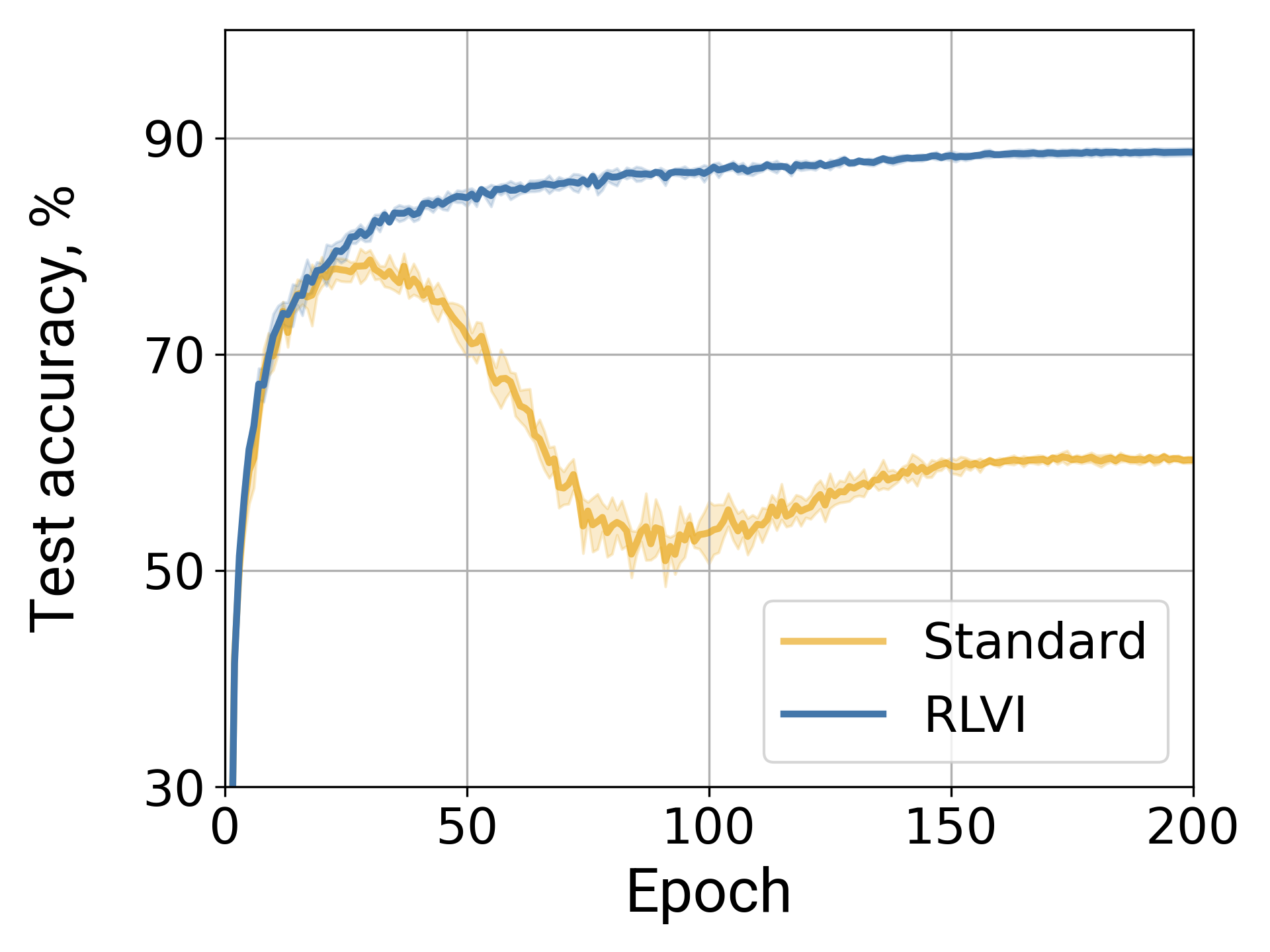}
    \end{minipage}\hfill
    \begin{minipage}[t]{0.25\linewidth}
      \includegraphics[width=\linewidth]{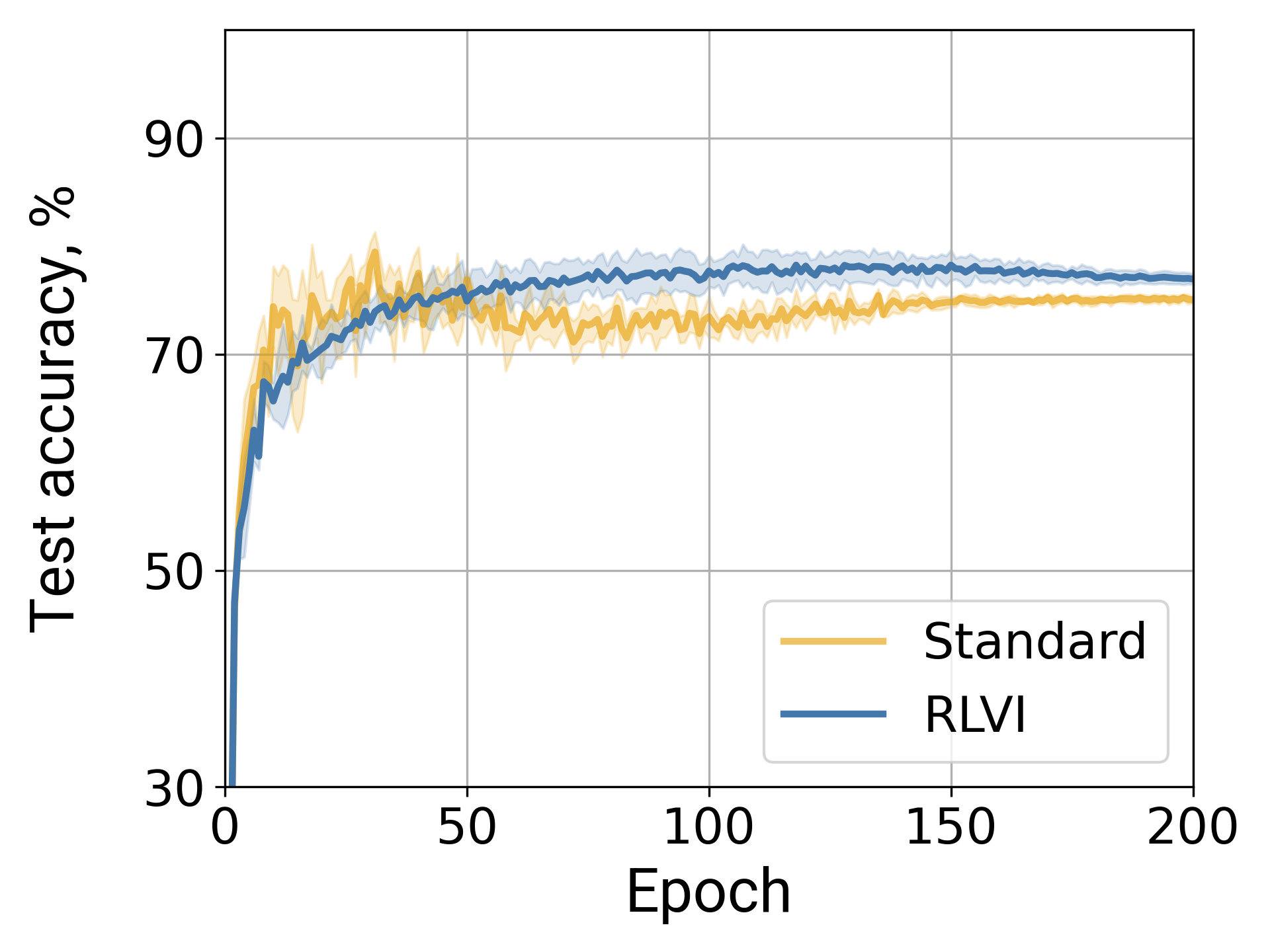}
    \end{minipage}\hfill
    \begin{minipage}[t]{0.25\linewidth}
      \includegraphics[width=\linewidth]{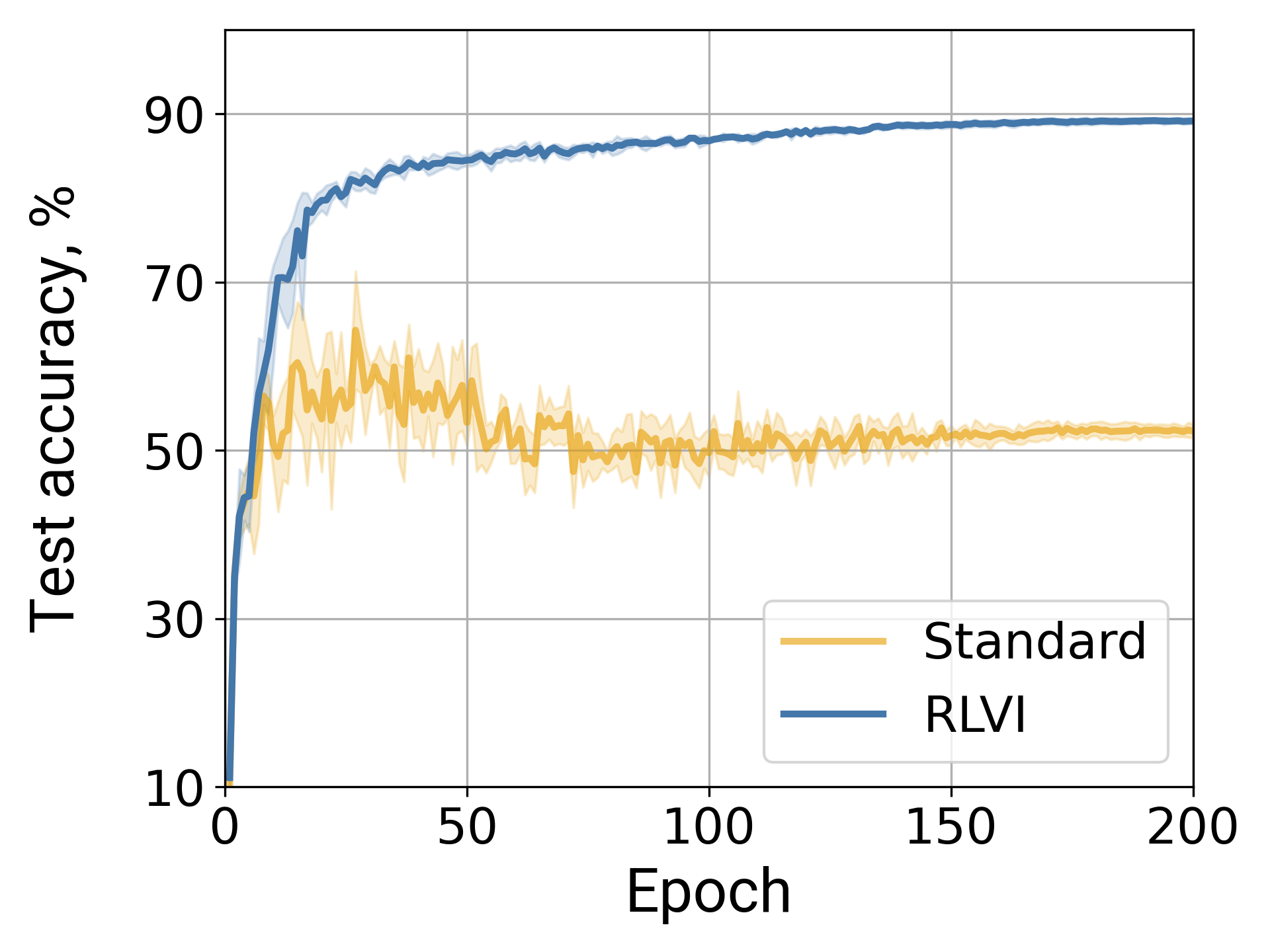}
    \end{minipage}\hfill
    \begin{minipage}[t]{0.25\linewidth}
      \includegraphics[width=\linewidth]{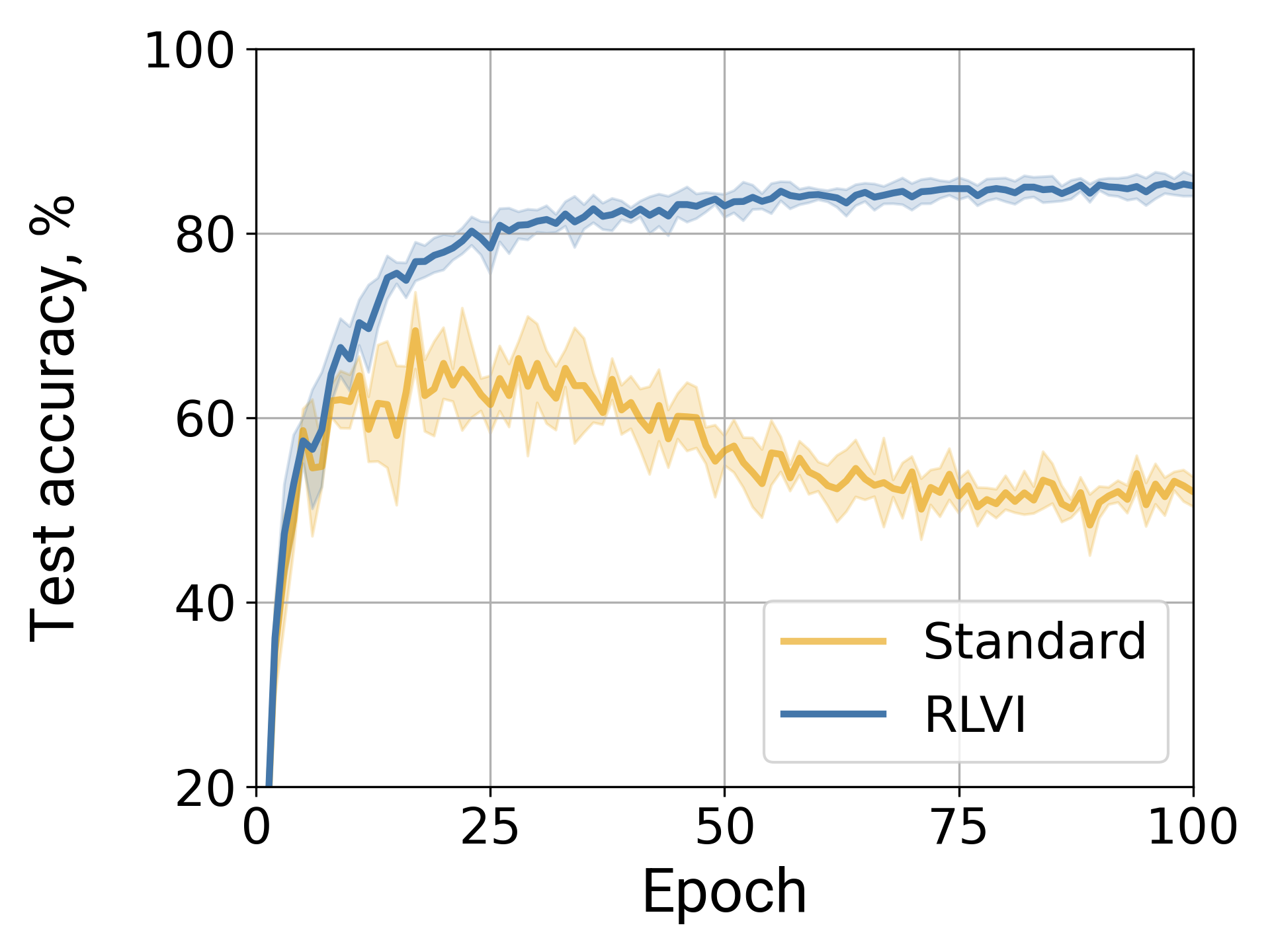}
    \end{minipage}\hfill
    \vfill
    \begin{minipage}[t]{0.25\linewidth}
      \includegraphics[width=\linewidth]{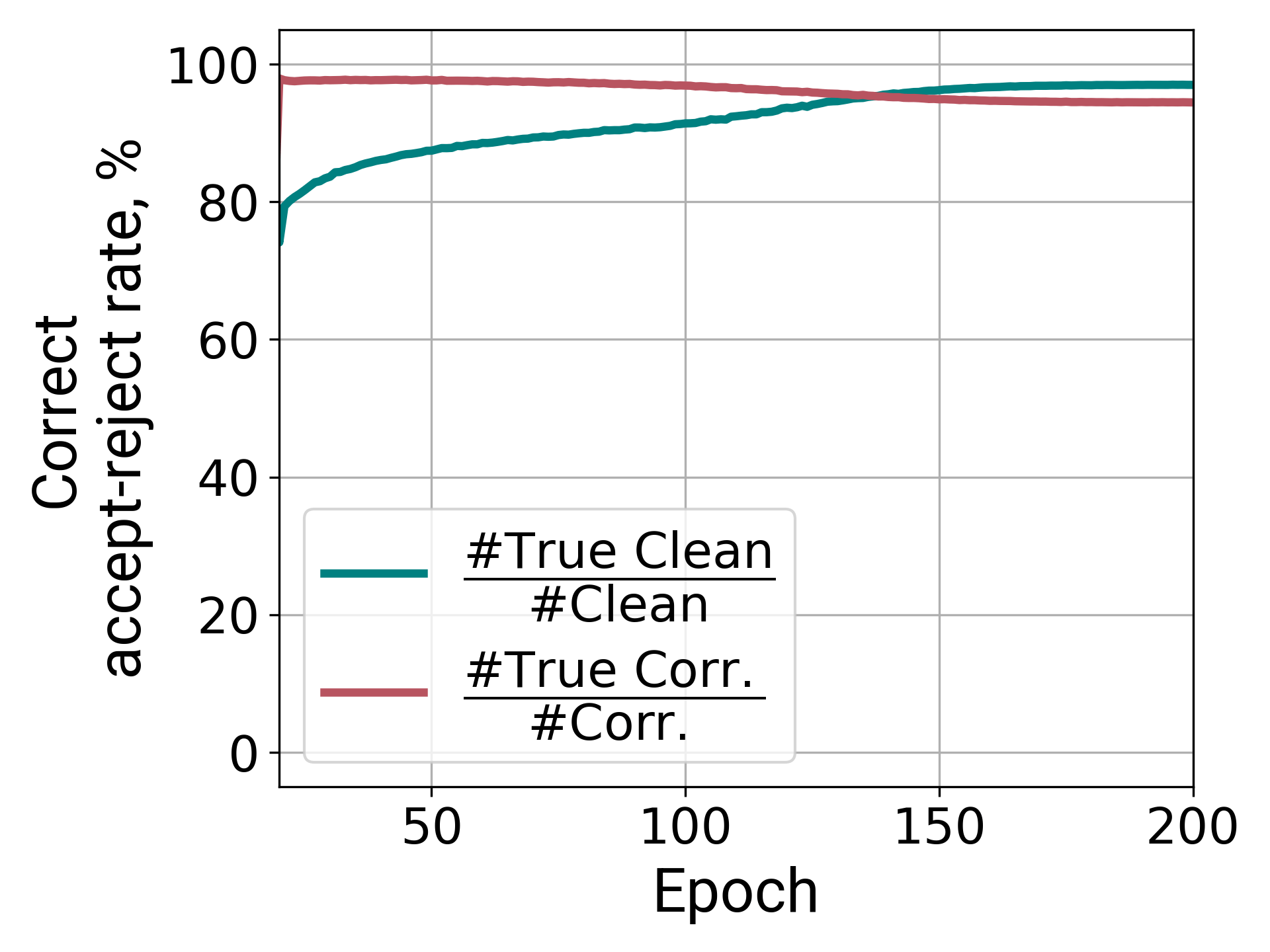}
      \caption*{Symmetric}
    \end{minipage}\hfill
    \begin{minipage}[t]{0.25\linewidth}
      \includegraphics[width=\linewidth]{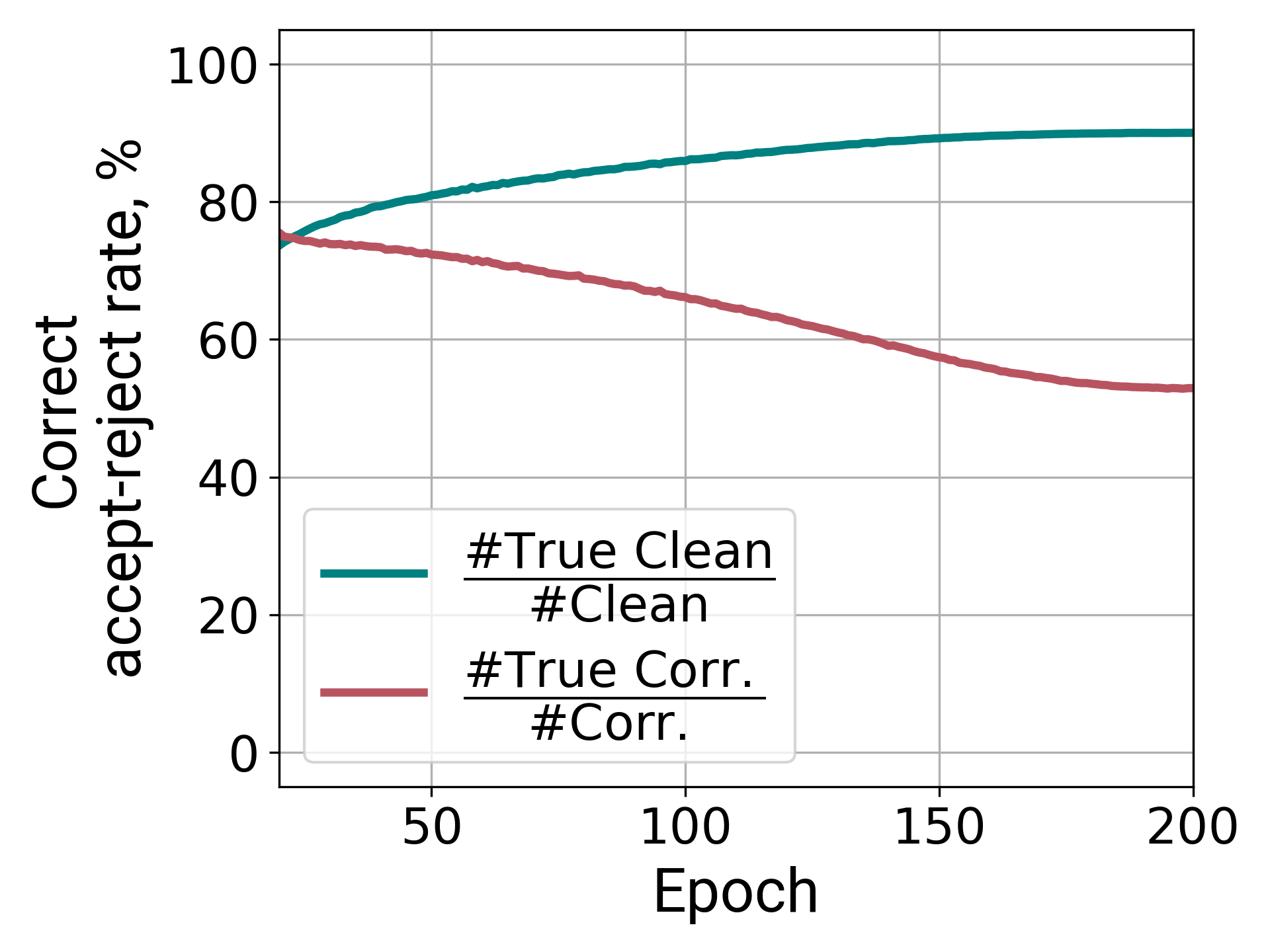}
    \caption*{Asymmetric}
    \end{minipage}\hfill
    \begin{minipage}[t]{0.25\linewidth}
      \includegraphics[width=\linewidth]{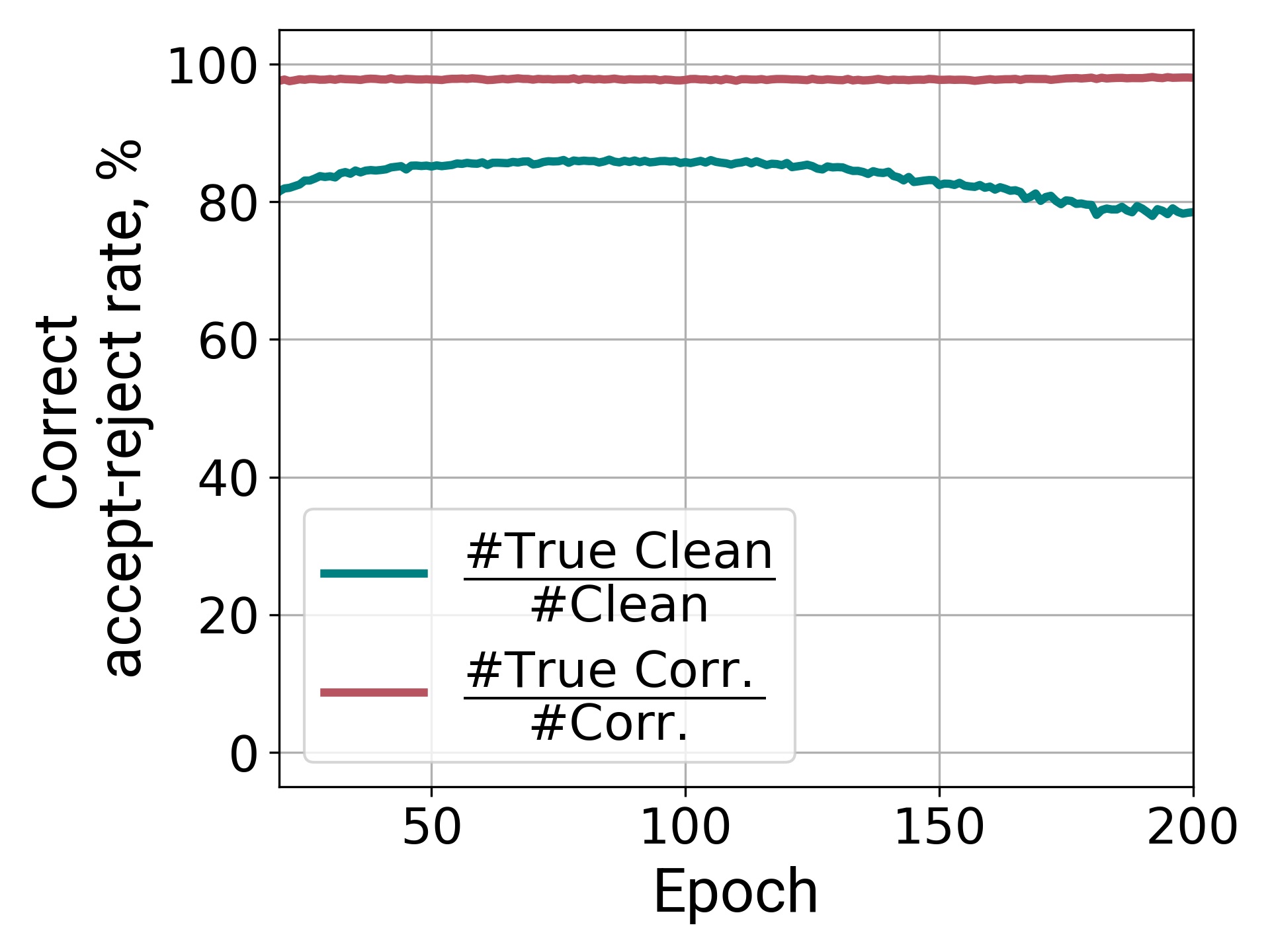}
      \caption*{Pairflip}
    \end{minipage}\hfill
    \begin{minipage}[t]{0.25\linewidth}
      \includegraphics[width=\linewidth]{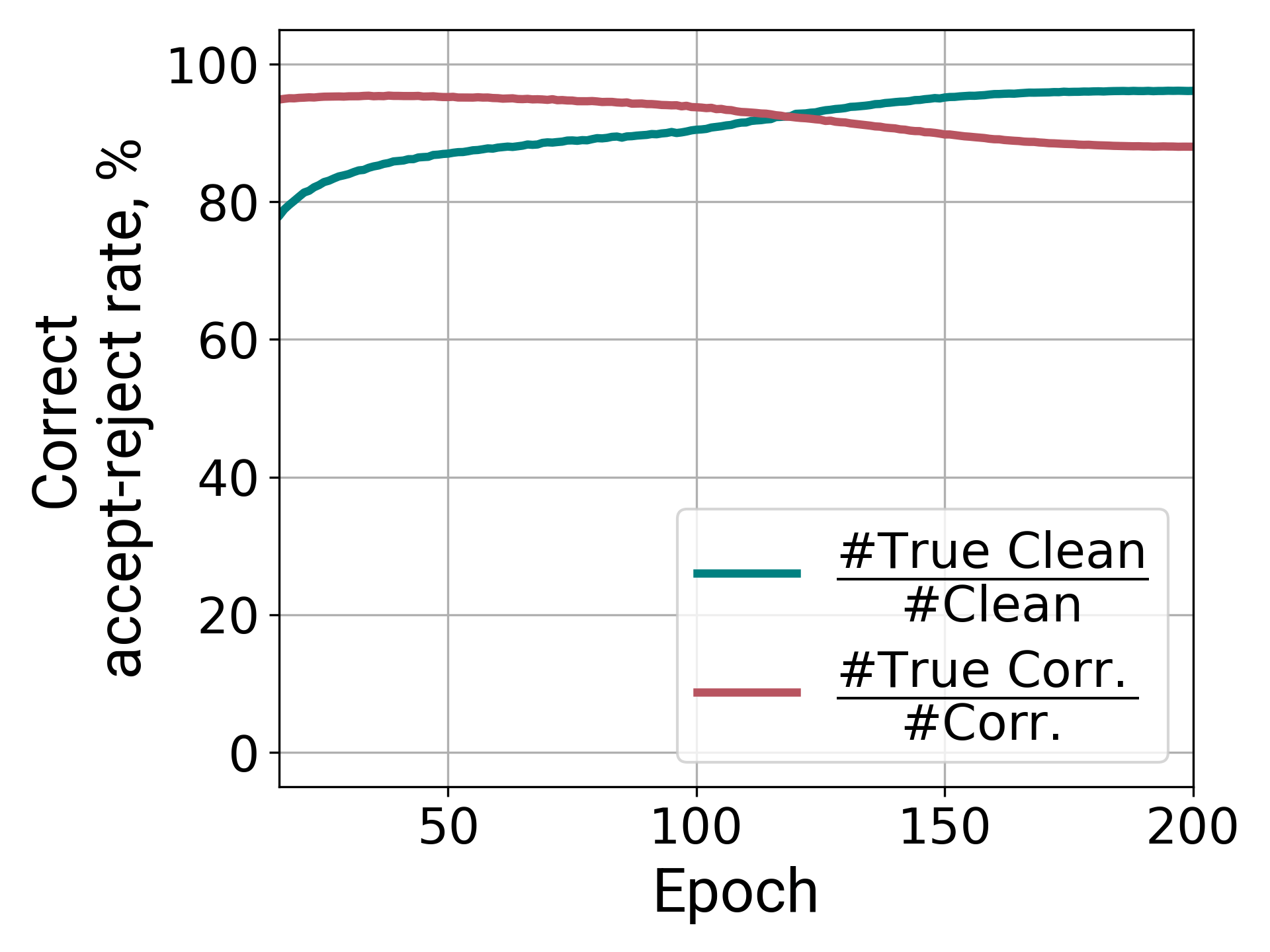}
      \caption*{Instance}
    \end{minipage}\hfill
\caption{
CIFAR10, 45\% noise rate. \emph{Top}: test accuracy (mean $\pm$ st. dev. over 5 runs). \emph{Bottom}: type I and type II errors (mean over 5 runs).
}
    \label{fig:performance-4}
\end{figure*}

\begin{figure*}[h]
    \centering
    \begin{minipage}[t]{0.25\linewidth}
      \includegraphics[width=\linewidth]{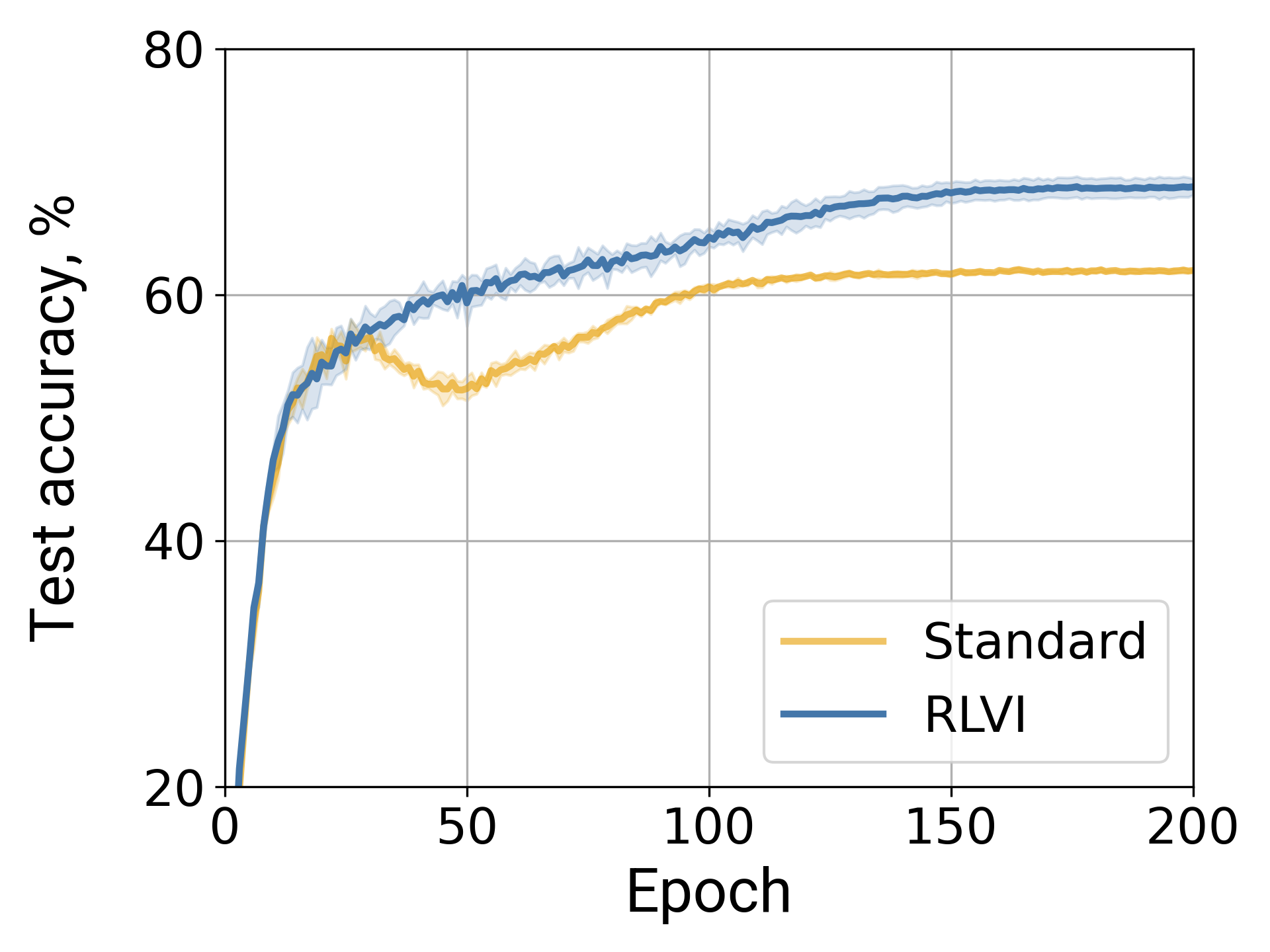}
    \end{minipage}\hfill
    \begin{minipage}[t]{0.25\linewidth}
      \includegraphics[width=\linewidth]{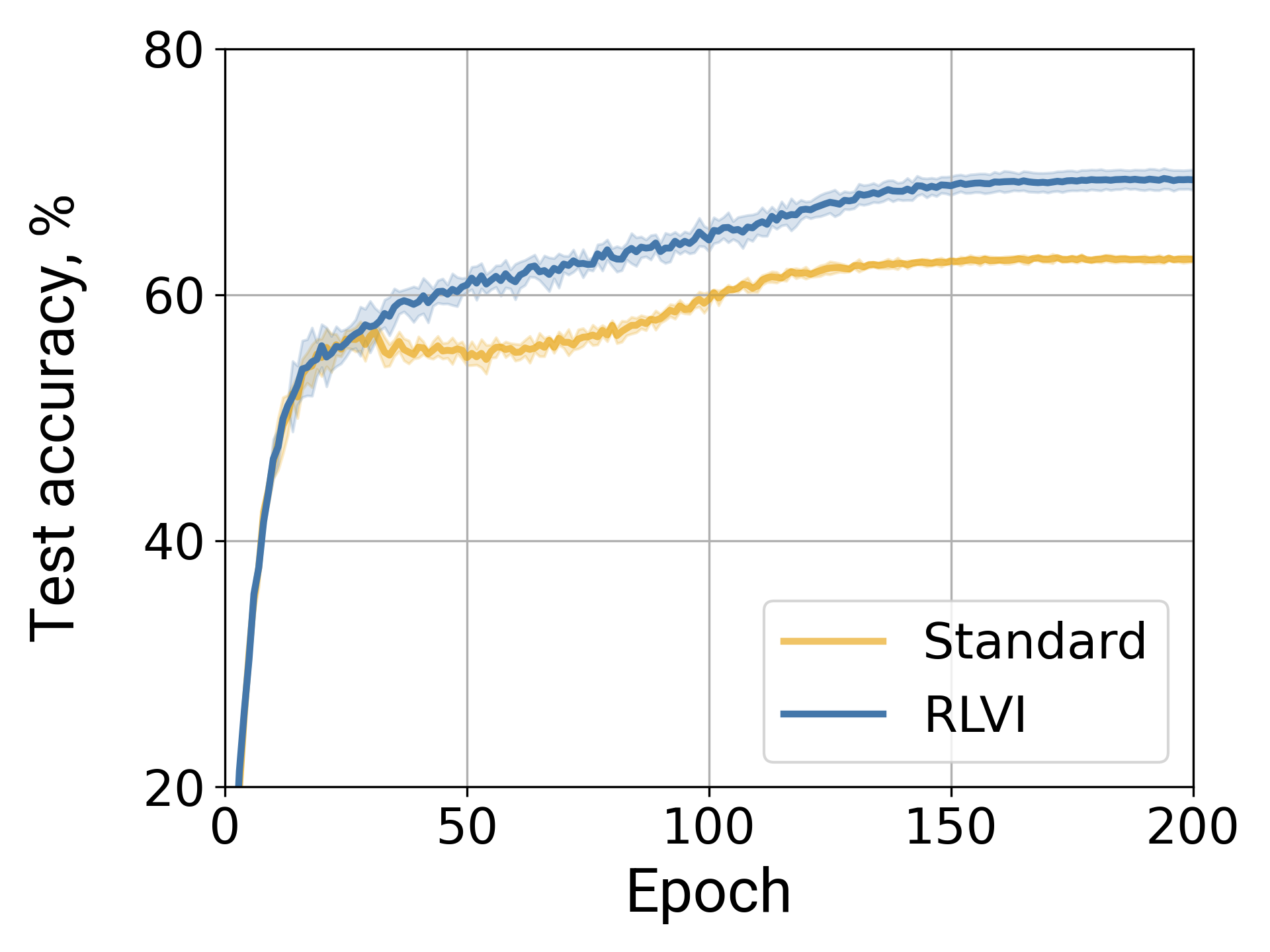}
    \end{minipage}\hfill
    \begin{minipage}[t]{0.25\linewidth}
      \includegraphics[width=\linewidth]{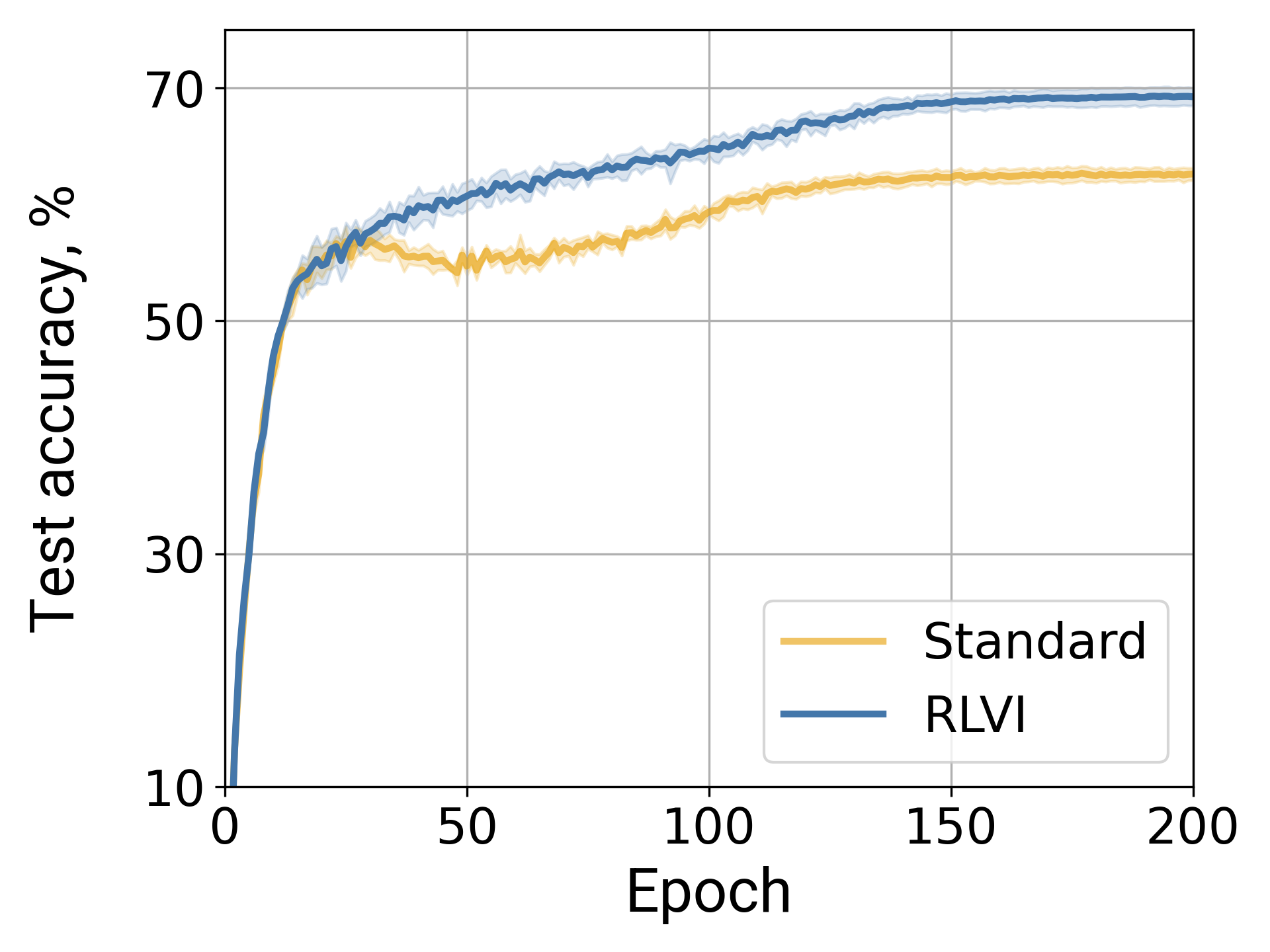}
    \end{minipage}\hfill
    \begin{minipage}[t]{0.25\linewidth}
      \includegraphics[width=\linewidth]{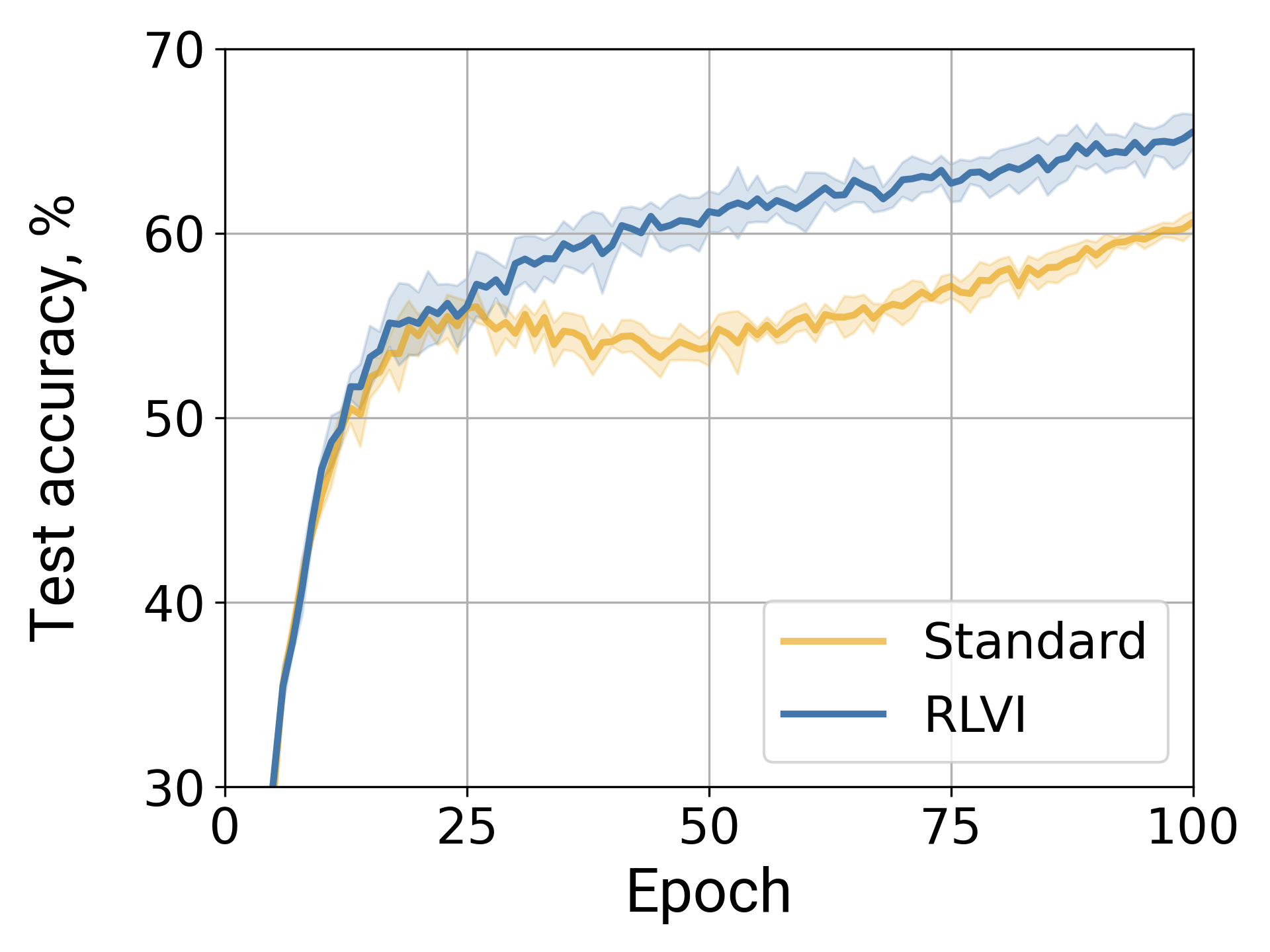}
    \end{minipage}\hfill
    \vfill
    \begin{minipage}[t]{0.25\linewidth}
      \includegraphics[width=\linewidth]{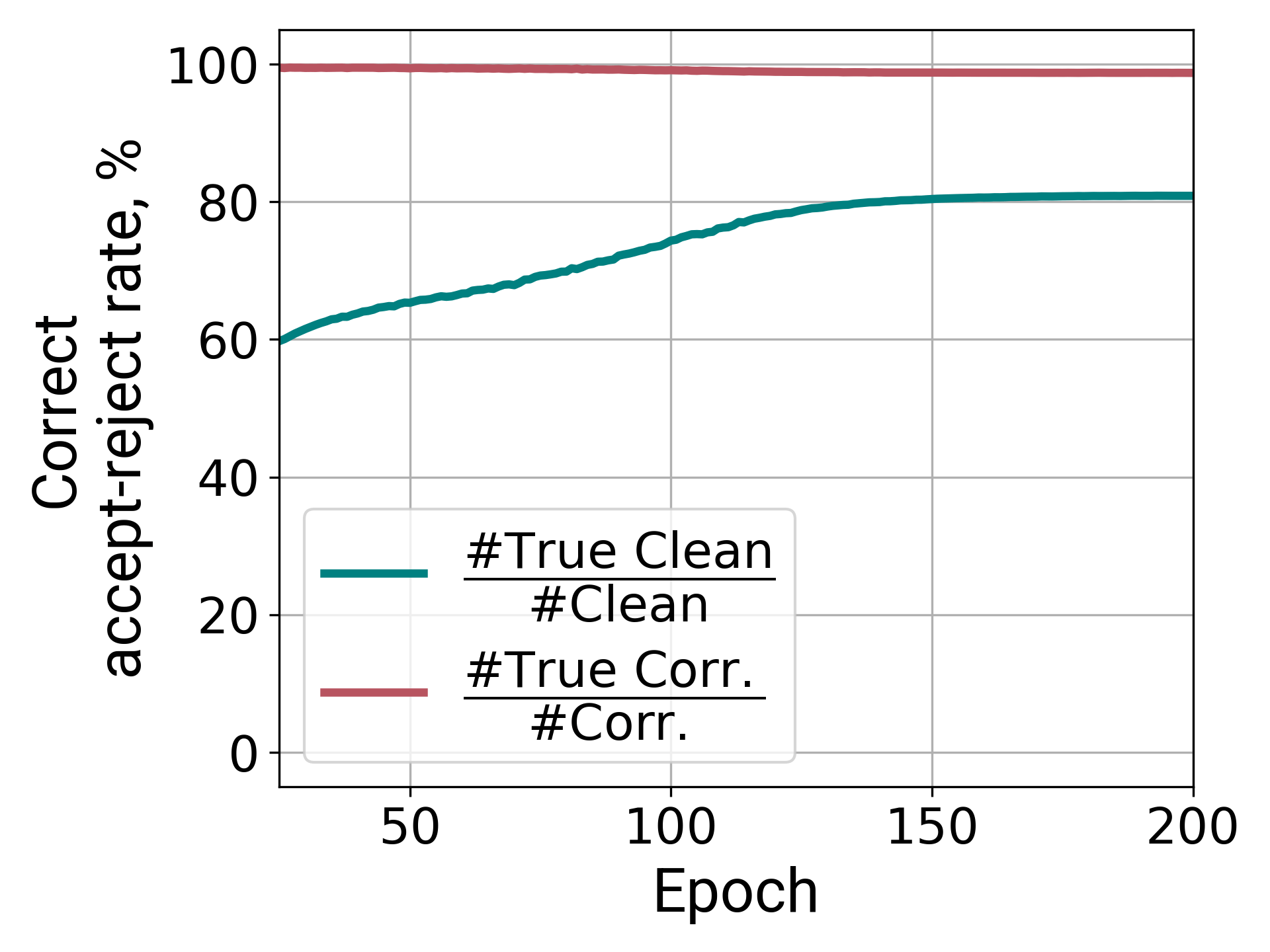}
      \caption*{Symmetric}
    \end{minipage}\hfill
    \begin{minipage}[t]{0.25\linewidth}
      \includegraphics[width=\linewidth]{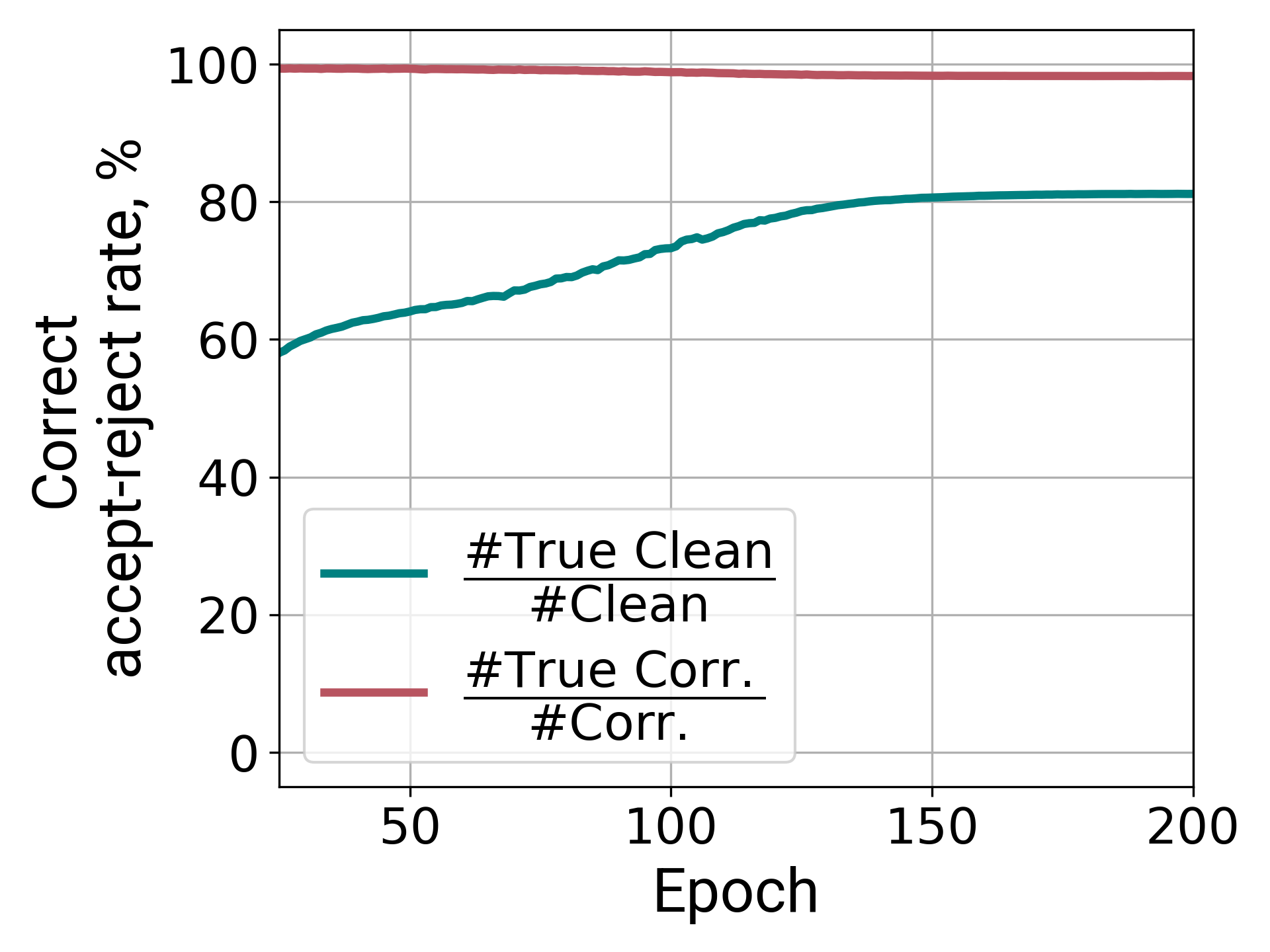}
      \caption*{Asymmetric}
    \end{minipage}\hfill
    \begin{minipage}[t]{0.25\linewidth}
      \includegraphics[width=\linewidth]{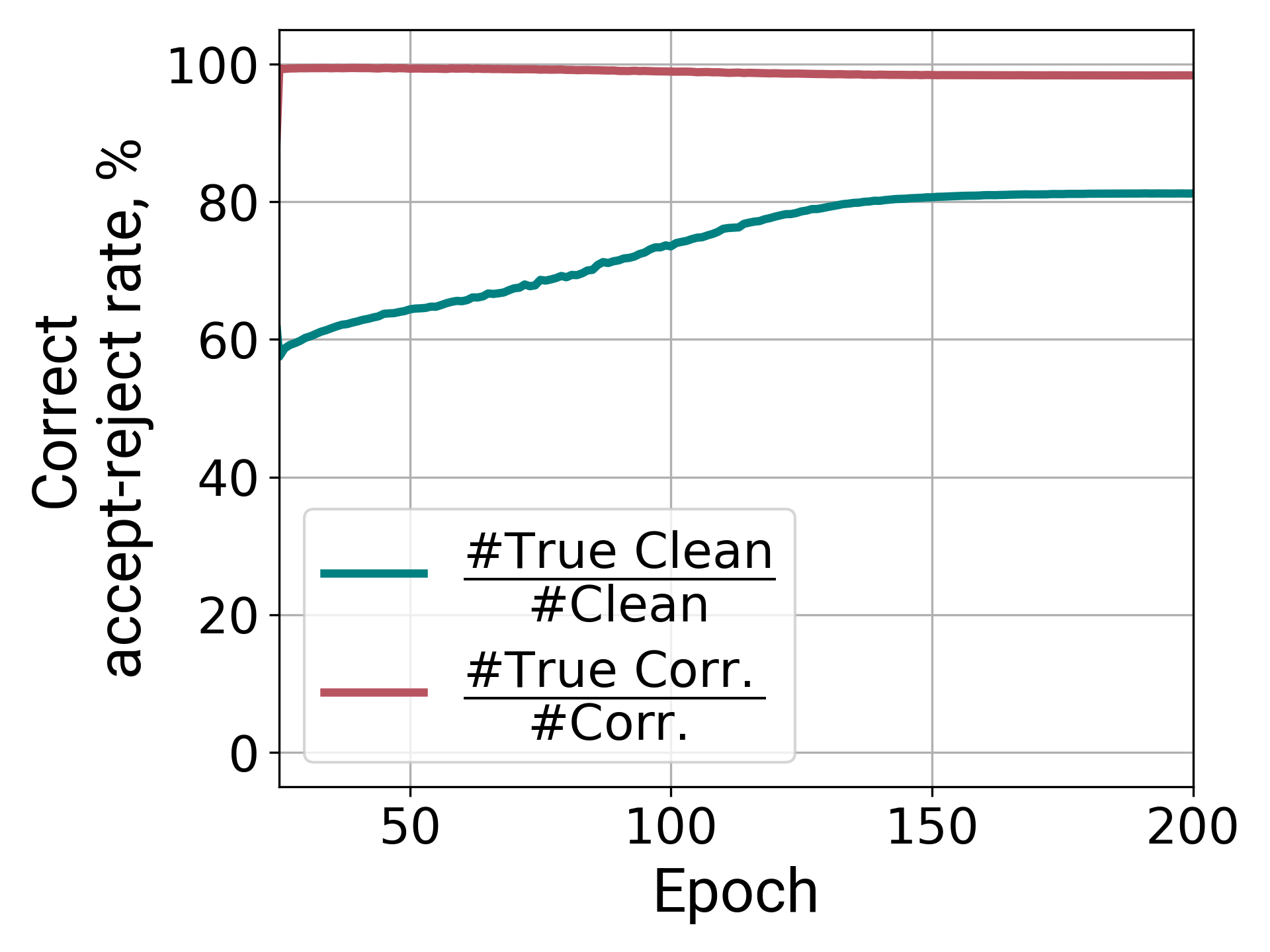}
      \caption*{Pairflip}
    \end{minipage}\hfill
    \begin{minipage}[t]{0.25\linewidth}
      \includegraphics[width=\linewidth]{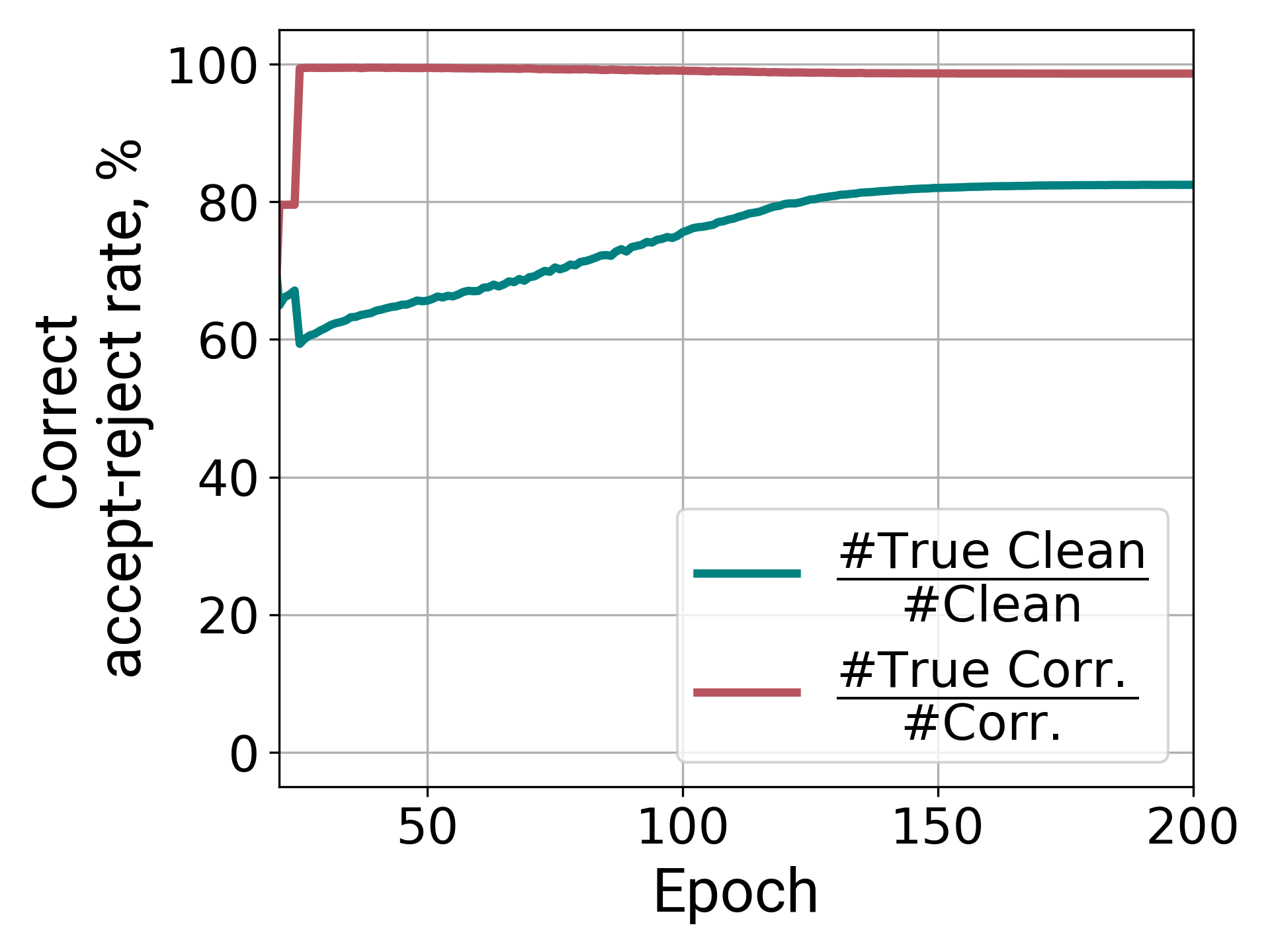}
      \caption*{Instance}
    \end{minipage}\hfill
\caption{
CIFAR100, 20\% noise rate. \emph{Top}: test accuracy (mean $\pm$ st. dev. over 5 runs). \emph{Bottom}: type I and type II errors (mean over 5 runs).
}
\label{fig:performance-5}
\end{figure*}

\begin{figure*}[h]
    \centering
    \begin{minipage}[t]{0.25\linewidth}
      \includegraphics[width=\linewidth]{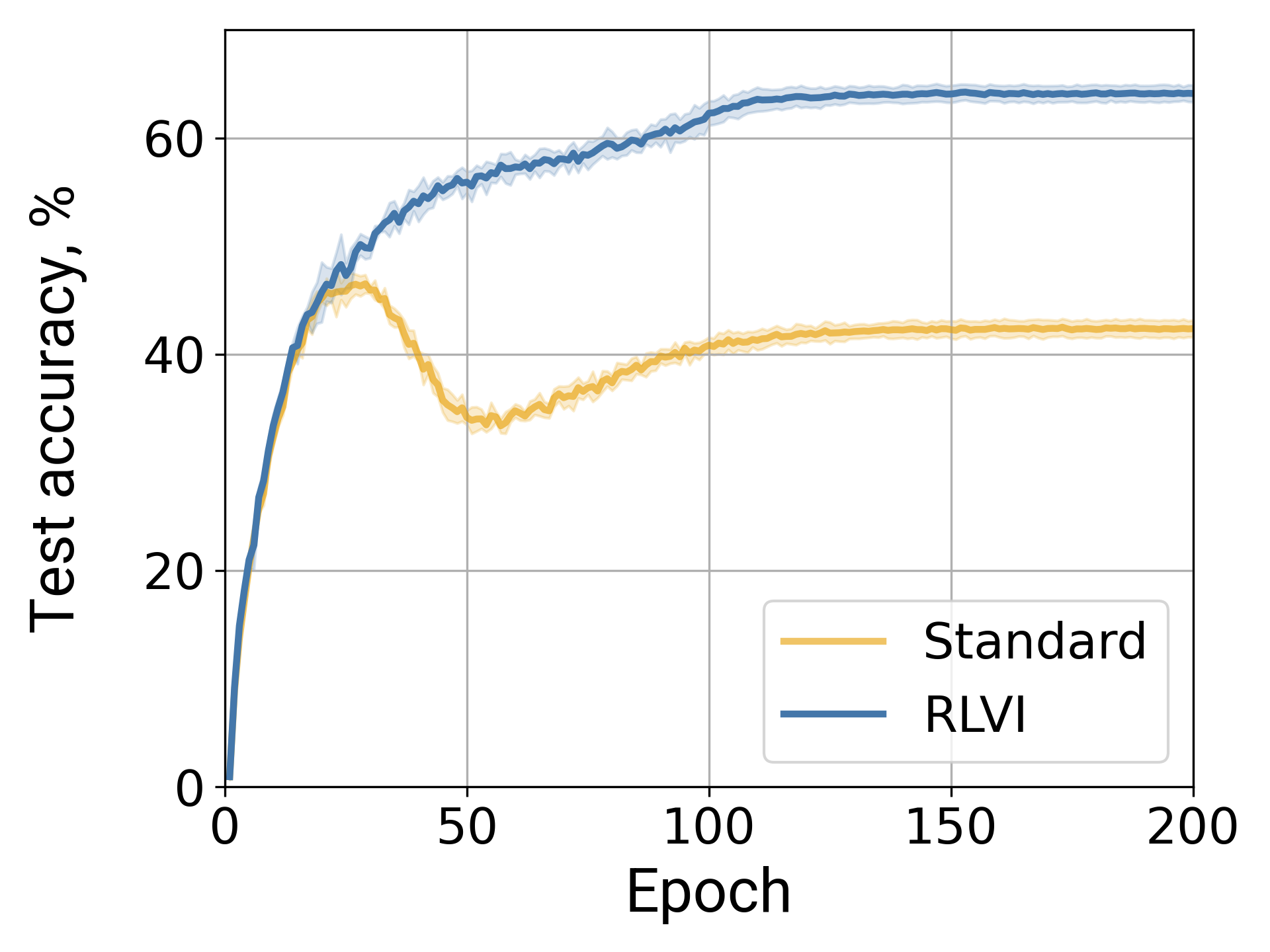}
    \end{minipage}\hfill
    \begin{minipage}[t]{0.25\linewidth}
      \includegraphics[width=\linewidth]{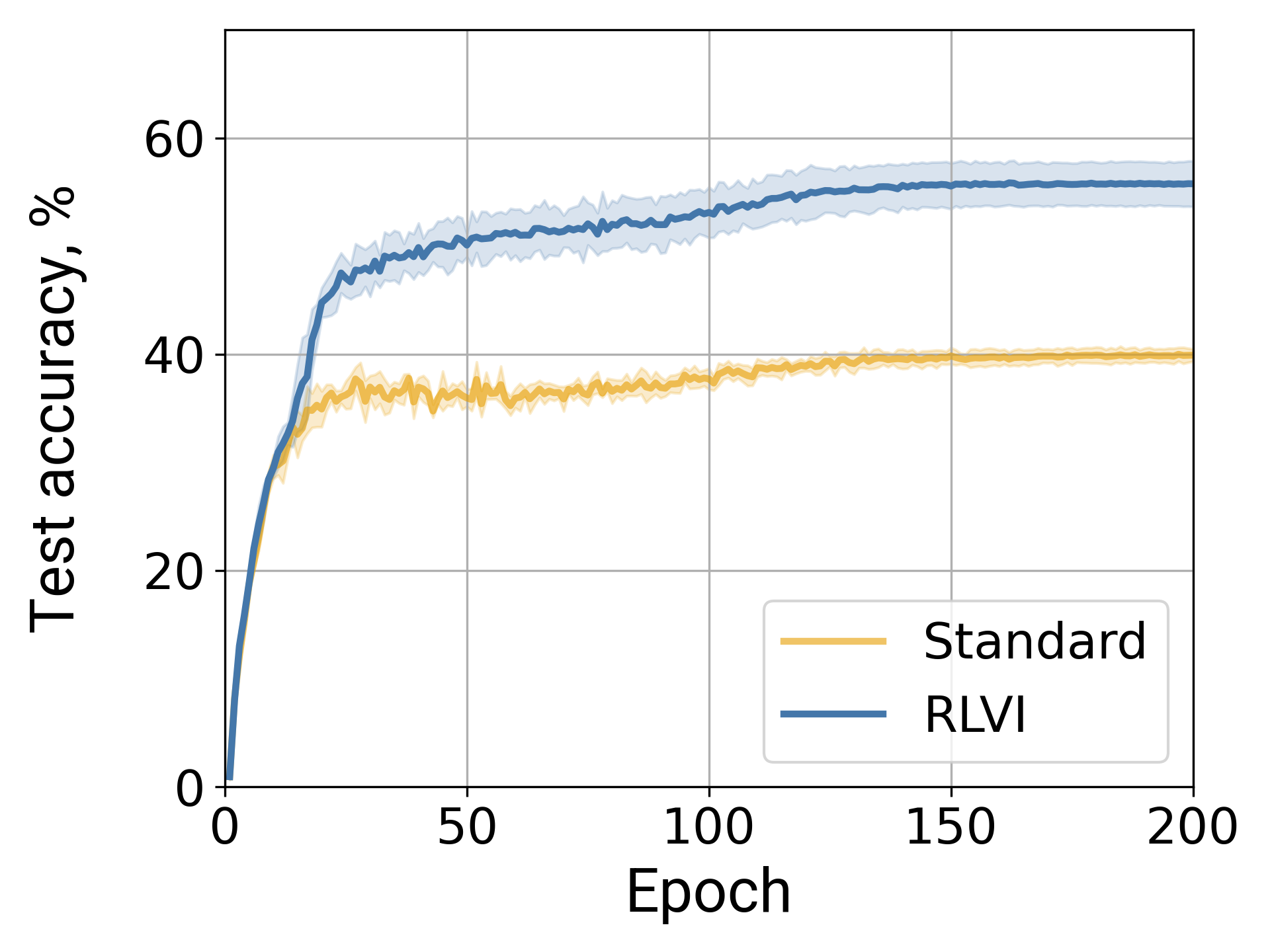}
    \end{minipage}\hfill
    \begin{minipage}[t]{0.25\linewidth}
      \includegraphics[width=\linewidth]{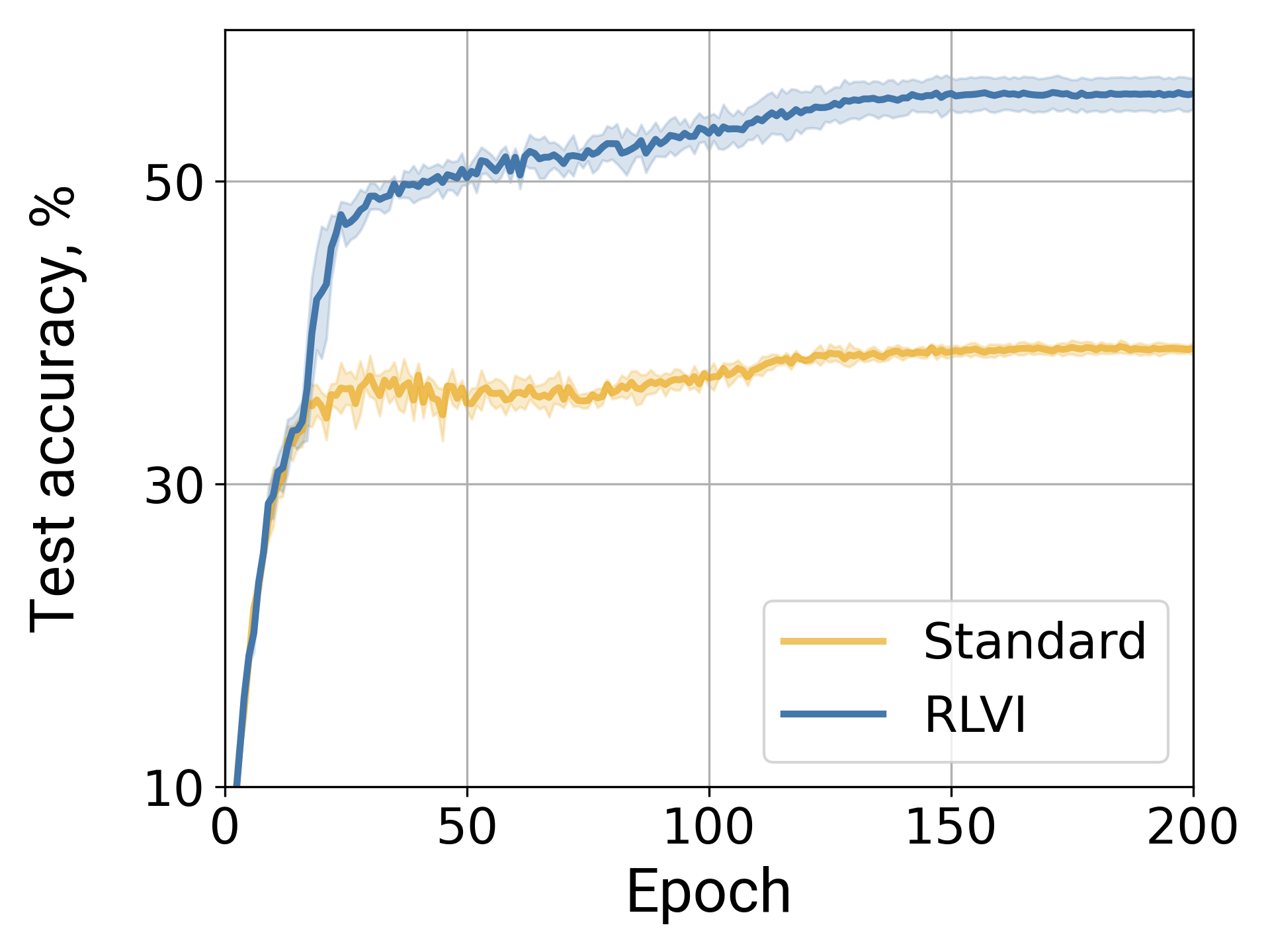}
    \end{minipage}\hfill
    \begin{minipage}[t]{0.25\linewidth}
      \includegraphics[width=\linewidth]{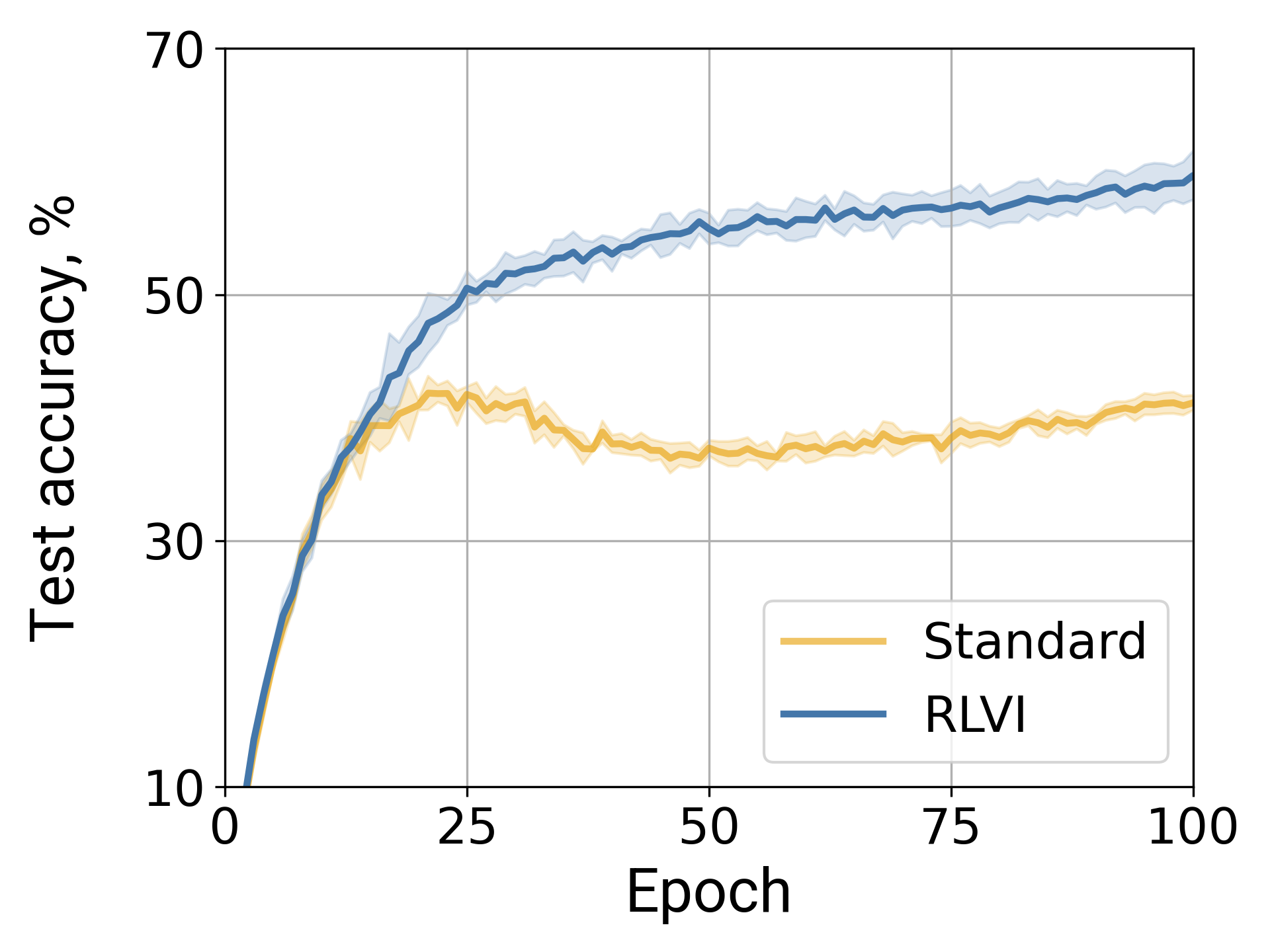}
    \end{minipage}\hfill
    \vfill
    \begin{minipage}[t]{0.25\linewidth}
      \includegraphics[width=\linewidth]{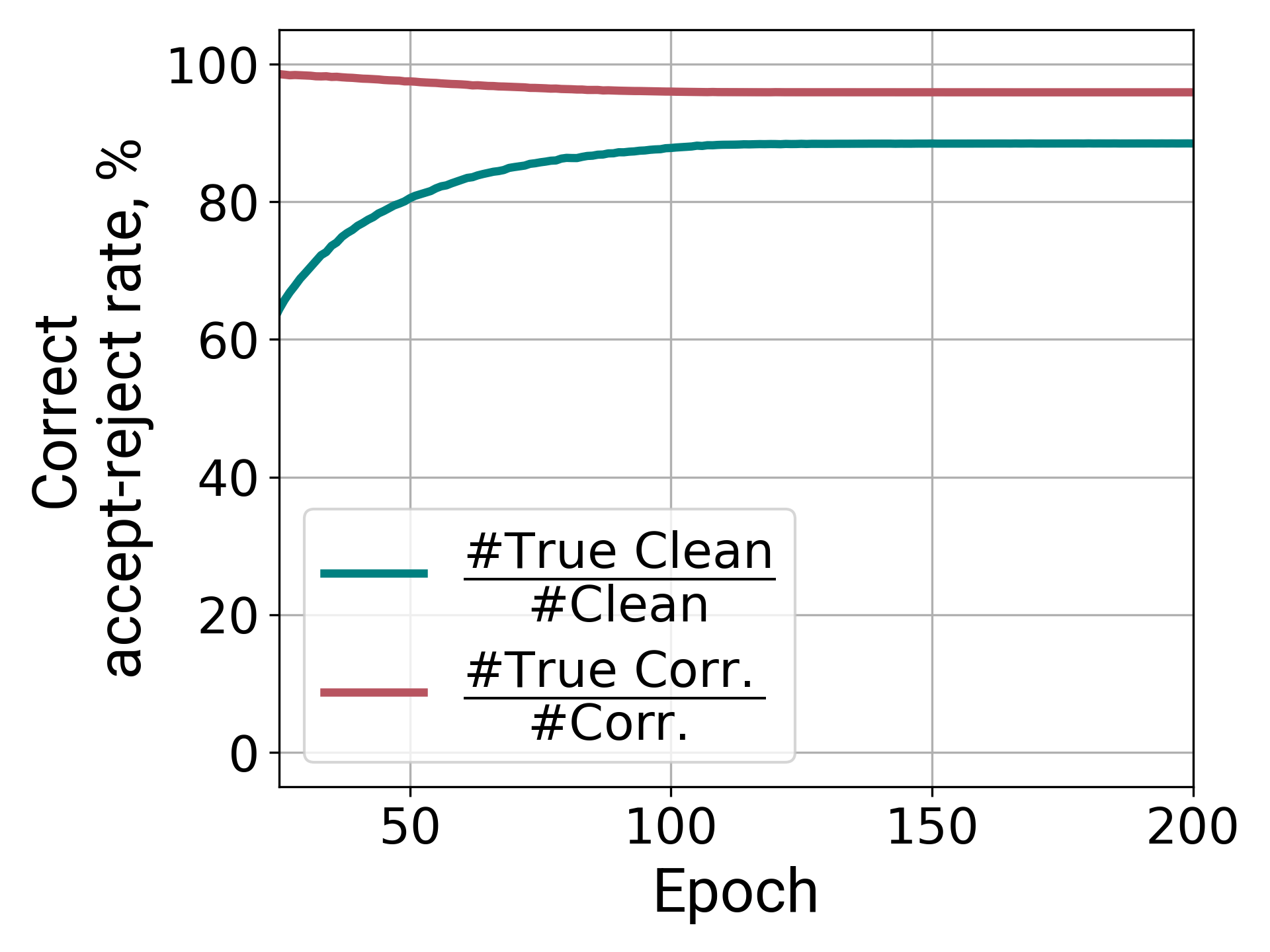}
      \caption*{Symmetric}
    \end{minipage}\hfill
    \begin{minipage}[t]{0.25\linewidth}
      \includegraphics[width=\linewidth]{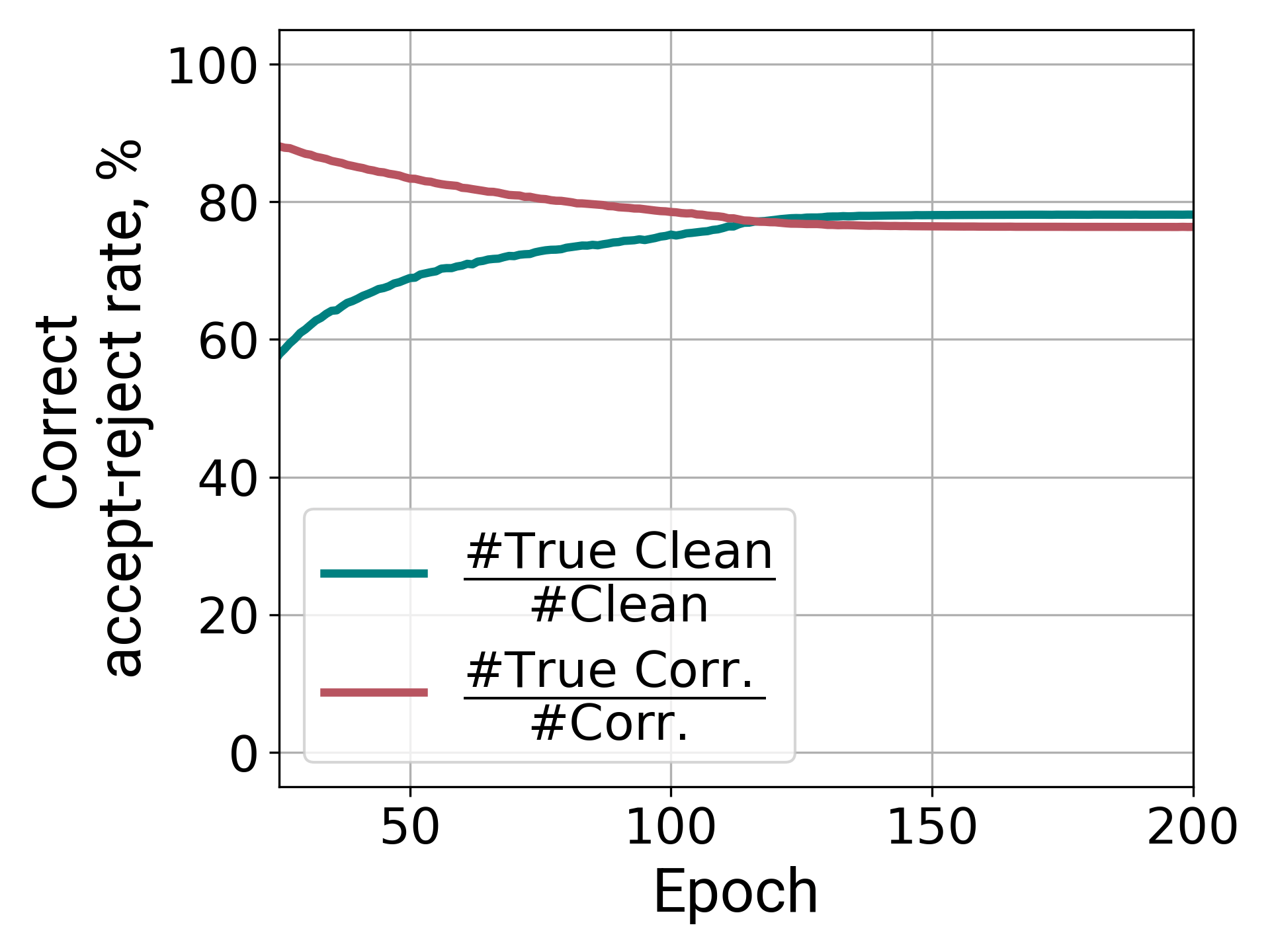}
    \caption*{Asymmetric}
    \end{minipage}\hfill
    \begin{minipage}[t]{0.25\linewidth}
      \includegraphics[width=\linewidth]{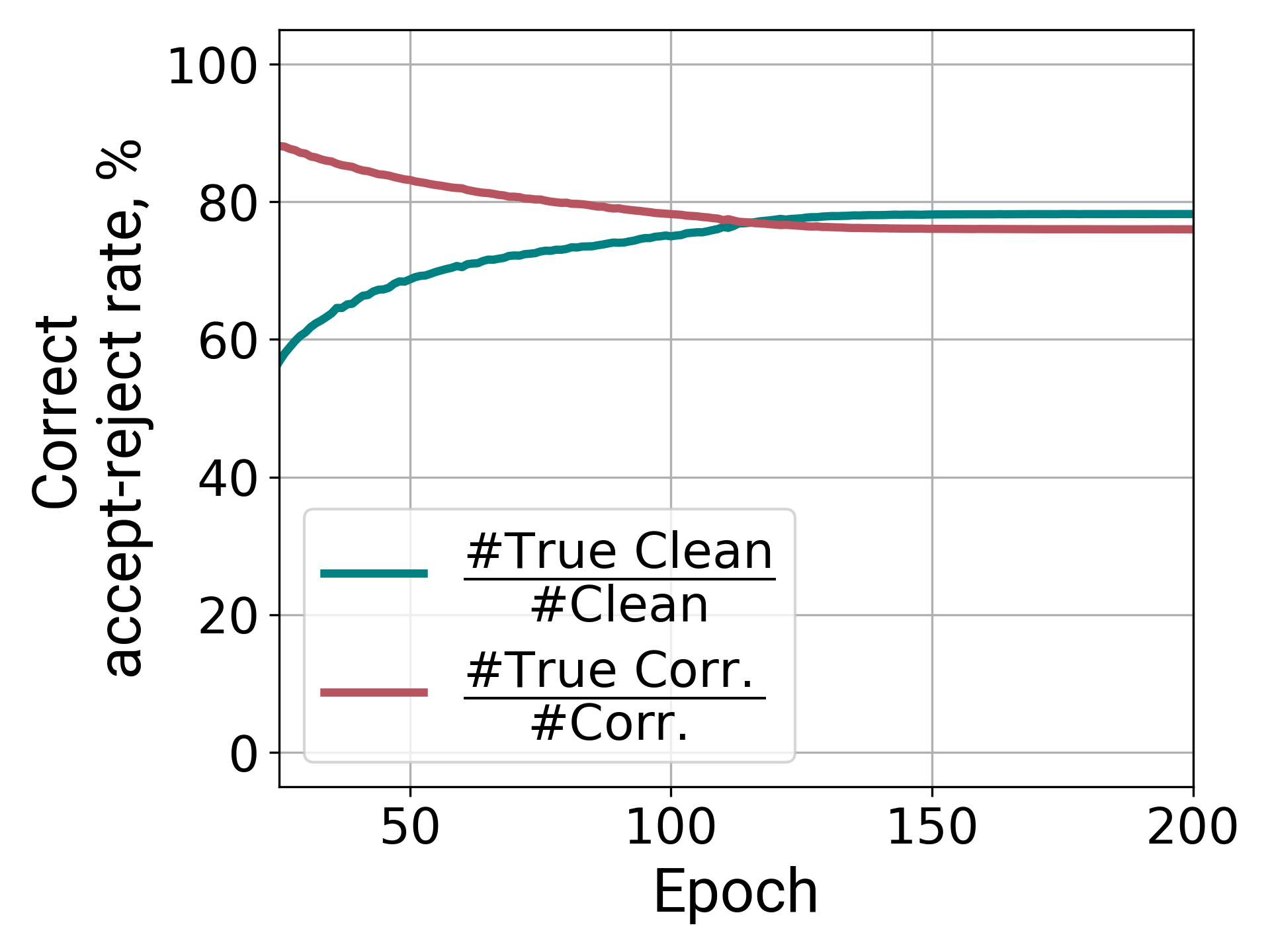}
      \caption*{Pairflip}
    \end{minipage}\hfill
    \begin{minipage}[t]{0.25\linewidth}
      \includegraphics[width=\linewidth]{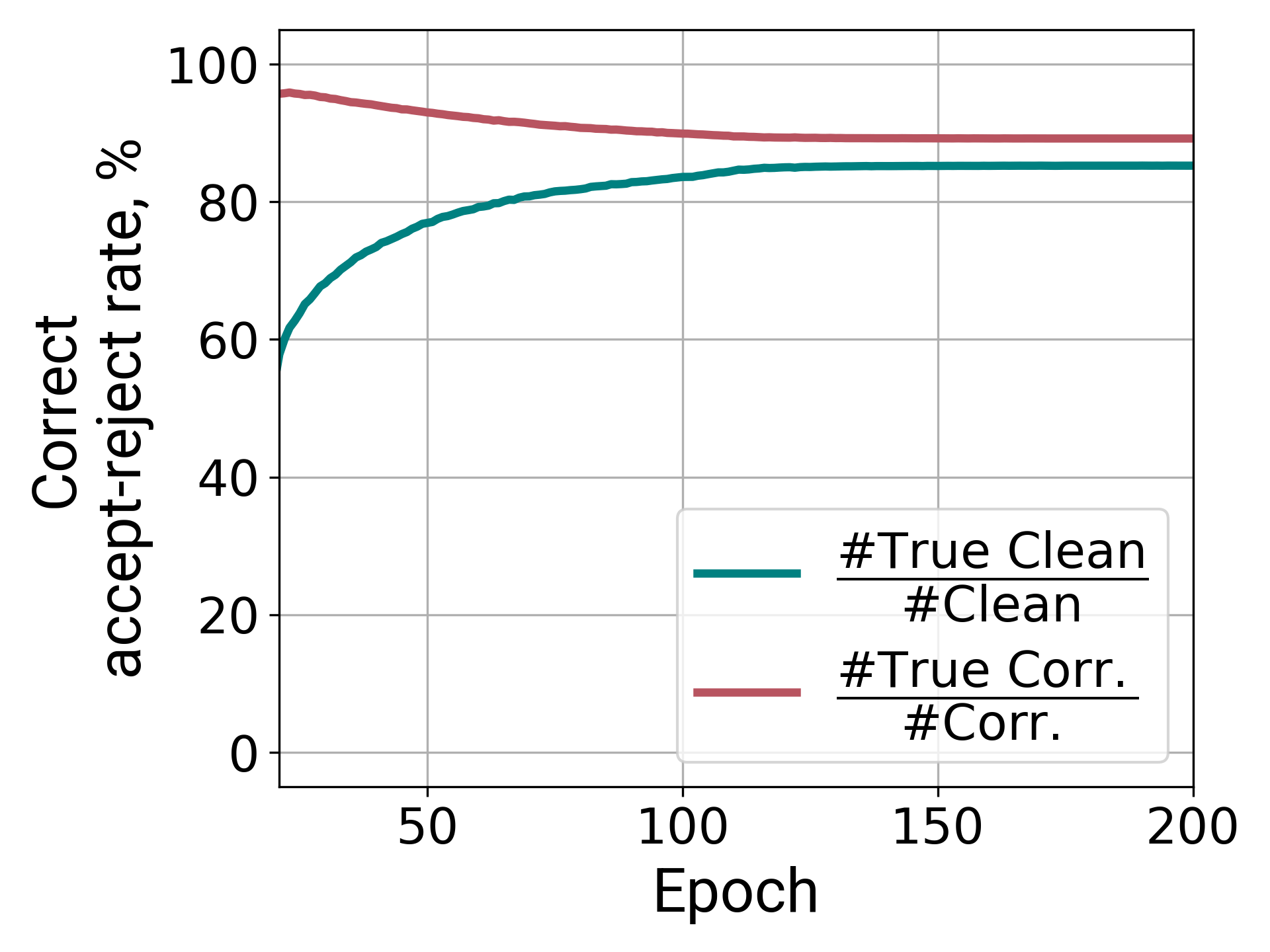}
      \caption*{Instance}
    \end{minipage}\hfill
\caption{
CIFAR100, 45\% noise rate. \emph{Top}: test accuracy (mean $\pm$ st. dev. over 5 runs). \emph{Bottom}: type I and type II errors (mean over 5 runs).
}
    \label{fig:performance-6}
\end{figure*}

\cref{fig:performance-1,fig:performance-2,fig:performance-3,fig:performance-4,fig:performance-5,fig:performance-6} are the plots, similar to \cref{fig:regularization}, showing the results of image classification after training on the data corrupted with three types of synthetic noise at corruption level $\varepsilon$ equal to 0.2 and 0.45. We also provide the ratio of correctly identified corrupted and non-corrupted images based on the introduced criterion. These figures show that, although type II error is not always below 5\%, overfitting is reduced in all settings, and RLVI achieves a higher accuracy than the standard method in all problem instances.

\subsection*{C3. Real corruption}
\label{real-noise}
\begin{figure*}[h]
\includegraphics[width=0.8\linewidth]{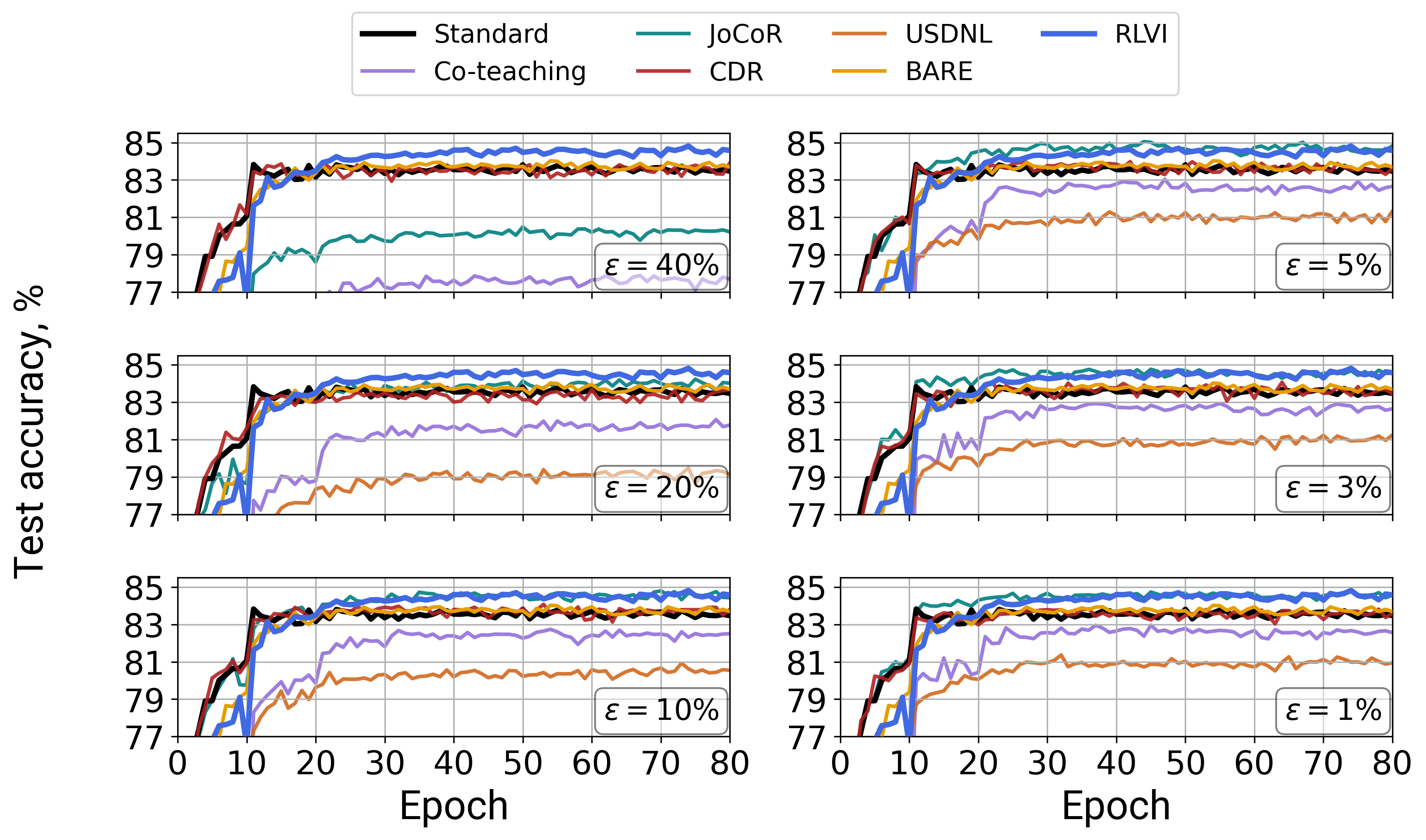}
  \centering
  \caption{Test accuracy (\%) for Food101, using different estimates of corruption level $\varepsilon$ for Co-teaching, JoCoR, CDR, and USDNL.}
  \label{fig:food-results}
\end{figure*}

In case of the Food101 dataset, a part of the training images is mislabeled. The ratio of such images, $\varepsilon$, is unknown, and thus we run the robust learning algorithms that depend on this hyperparameter: Co-teaching, JoCoR, CDR, and USDNL, using different estimates of $\varepsilon$ (40\%, 20\%, 10\%, 5\%, 3\%, and 1\%). \cref{fig:food-results} presents the accuracy of different methods on the manually cleaned testing set from Food101. As the estimate of $\varepsilon$ used in training decreases, the accuracy of alternative methods improves and becomes optimal with $\varepsilon$ around 3\%. In this experiment, RLVI achieved the best results when no regularization was used. This can be attributed to the weaker overfitting in the case of the Food101 dataset, as compared to examples with synthetic noise: test accuracy for RLVI without regularization in \cref{fig:food-results} gradually increases, in contrast to the corresponding curve in \cref{fig:regularization}.

\section*{D. Unbounded likelihood}
As we note in \cref{subsec:standard}, considering unbounded likelihood function in $\ell_{\state}(\observation)$ can lead to a degenerate solution $\state$ when using RLVI. In general, the problem of likelihood maximization is ill-posed, and one example when such pathological estimate arises is, e.g, the Gaussian Mixture Model \cite{bishop2006pattern}. In the following, we address covariance estimation from corrupted data. The corresponding negative log-likelihood function, which is used to estimate the mean $\bm{\mu}$ and the covariance matrix $\bm{\Sigma}$, is
\begin{align}
    \ell_{\state}(\bm{z}) = \dfrac{1}{2} \left[(\bm{z} - \bm{\mu})^\top \bm{\Sigma}^{-1} (\bm{z} - \bm{\mu}) + \ln{\det{\bm{\Sigma}}} + d\ln{2\pi}\right],
\label{eq:cov-loss}
\end{align}
where $d$ is the dimension of the problem and $\state = (\bm{\mu}, \bm{\Sigma})$. The first and second terms of \eqref{eq:cov-loss} become unbounded as $\bm{\Sigma}$ approaches rank deficiency. We note that the covariance estimate for RLVI is of the form: 
\begin{align}
    &\bm{\Sigma}_{\text{RLVI}} = \dfrac{1}{\weight^\top\one} \sum_{i=1}^{n} \pi_i (\bm{z}_i - \bm{\mu}_{\text{RLVI}}) (\bm{z}_i - \bm{\mu}_{\text{RLVI}})^{\top},
\end{align}
where $\bm{\mu}_{\text{RLVI}} = \sum_{i=1}^{n} \pi_i \bm{z}_i \,/\, \weight^\top\one$.
Therefore it is possible to find a set of weights $\pi_i$ such that $\bm{\Sigma}_\text{RLVI}$ becomes rank-deficient in a manner that minimizes the loss. This is indeed confirmed in our experiments.

To avoid such a singular solution, we impose additional regularization for $\weight$ and ensure that enough samples are used for learning: $\sum_{i=1}^{n}\pi_i = n(1 - \varepsilon) \ge n_0$. That is, the total number of non-corrupted samples should be at least $n_0$. The corresponding constrained optimization problem for $\weight$ is
\begin{align}
    \begin{cases}
        &\min\limits_{\weight\in (0; 1)^n} \mathcal{L}(\state, \weight), \\
        &\weight^\top\one \geq n_0.
    \end{cases}
    \label{eq:constrained}
\end{align}

Its Lagrangian $L(\weight, \lambda)$ is thus the sum of objective $\mathcal{L}(\state, \weight)$, and a constraint weighted with Lagrange multiplier $\lambda$:
\begin{align}
    L(\weight, \lambda) = \mathcal{L}(\state, \weight) + \lambda \left( n_0 - \weight^\top\one \right).
\end{align}
In this analysis we consider the optimization step with respect to $\weight$, and thus omit $\state$ from the variables of the Lagrangian above. Consequently, the corresponding Karush–Kuhn–Tucker conditions give
\begin{enumerate}
    \item $\nabla_{\weight} L(\weight, \lambda) = \nabla_{\weight} \mathcal{L}(\state, \weight) - \lambda \cdot \one = 0 \implies \pi_i = \left(1 + \dfrac{\naw}{\aw} e^{\lossi - \lambda}\right)^{-1}$,
    \item $\lambda \geq 0$,
    \item $\lambda \cdot \left( n_0 - \weight^\top\one \right) = 0$.
\end{enumerate}

Note that if the constraint is not active, we fall back to the default RLVI formulation in which $\weight$ are computed with \eqref{eq:fixed-point}. In contrast, when the constraint is active, a change to the solution $\weight$ is required.

a. Inactive constraint: $\weight^\top\one > n_0 \implies \lambda = 0$, and
\begin{align}
    \pi_i &= \left(1 + \dfrac{\naw}{\aw} e^{\lossi} \right)^{-1}.
    \label{inactive-1}
\end{align}
b. Active constraint: $\weight^\top\one = n_0 \implies \lambda > 0$. In this case, $\weight^\top\one = n_0$, and for $\weight$ and $\lambda$, we obtain the following equations:
\begin{align}
    \begin{cases}
        \pi_i = \left(1 + \dfrac{n - n_0}{n_0} e^{\lossi - \lambda} \right)^{-1}, \\
        \lambda > 0, \\
        \sum\limits_{i=1}^n \pi_i = n_0.
    \end{cases}
\label{eq:active}
\end{align}
Hence, in the case of a sparse solution (most $\pi_i$ are around zero), \eqref{eq:active} allows us to find the dual variable $\lambda_{\star}$ and, after a substitution, obtain the corrected $\weight$.

\begin{figure}[H]
    \includegraphics[width=0.4\linewidth]{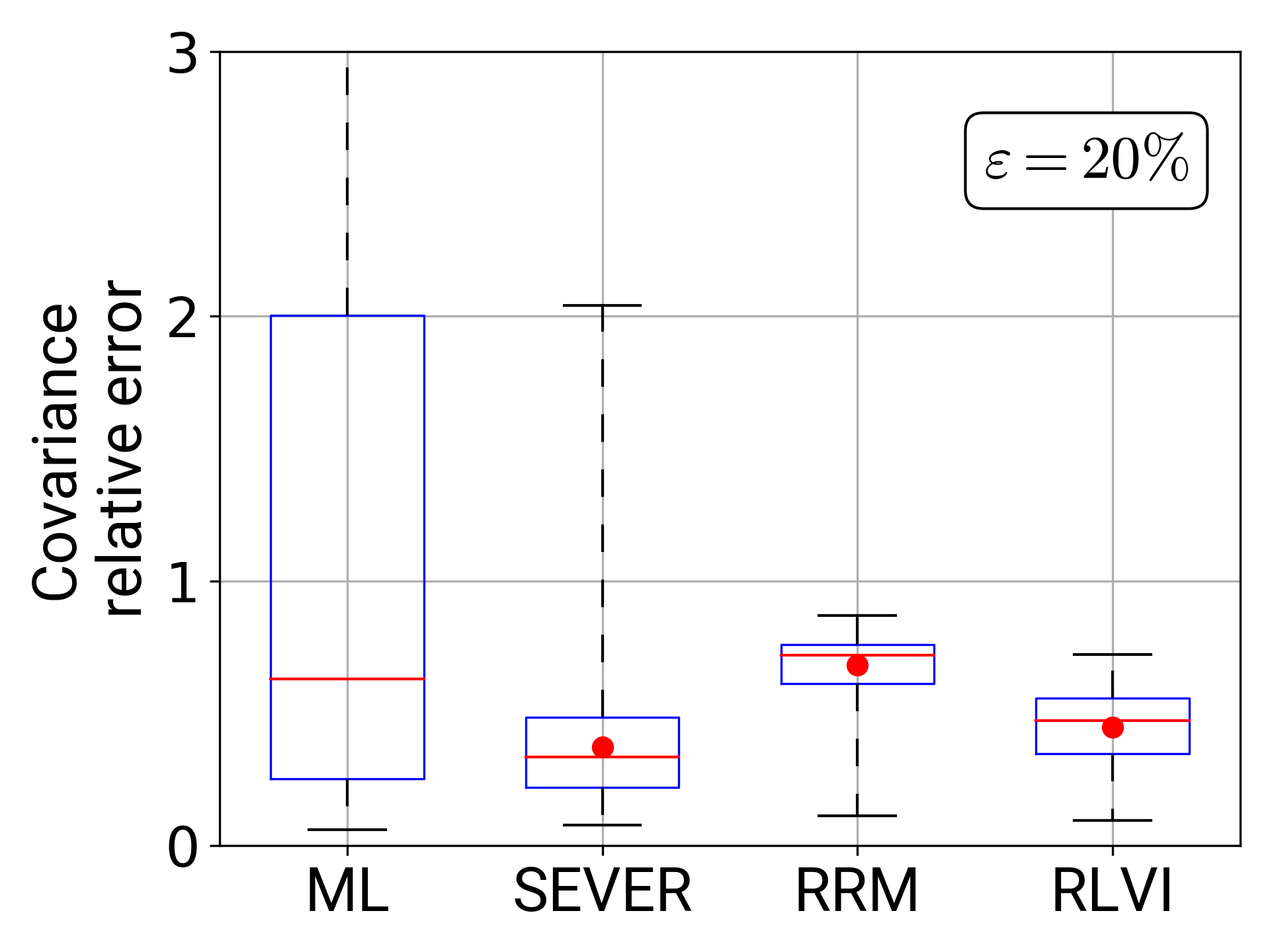}
    \centering
    \caption{Covariance estimation. Boxplots of relative errors after 100 Monte Carlo runs. Constrained RLVI \eqref{eq:constrained} is not subject to the singular covariance issue that arises due to the unbounded likelihood.}
    \label{fig:cov-est}
\end{figure}

To test this regularization in practice, we reproduce the covariance estimation problem from \cite{osama2020robust}: $n = 50$ samples are generated synthetically, of which $\varepsilon = 20\%$ are sampled from the corrupted distribution. The dimension of samples is $d = 2$; 100 Monte Carlo tests are performed, similar to \cref{subsec:standard}. Note that alternative 
methods, \textsc{SEVER} and \textsc{RRM}, use the following upper-bound: $\varepsilon \leq \widetilde{\varepsilon} = 30\%$. Hence, for a fair comparison, we set $n_0 = n (1 - \widetilde{\varepsilon}) = 35$. \Cref{fig:cov-est} shows relative errors obtained with standard Maximum Likelihood (ML), \textsc{SEVER}, \textsc{RRM}, and the constrained variant of RLVI according to \eqref{eq:constrained}. These results demonstrate that in the case of the unbounded likelihood one might still need to employ the hyperparameter $\widetilde{\varepsilon}$. But, in this case as well, RLVI can attain a low relative error with a tight confidence interval.

\end{document}